\documentclass[10pt,twocolumn,letterpaper]{article}

\usepackage[pagenumbers]{cvpr} 

\usepackage{graphicx}
\usepackage{amsmath}
\usepackage{amssymb}
\usepackage{booktabs}

\usepackage{wrapfig}
\usepackage{float} 

\usepackage[ruled,vlined]{algorithm2e}   
\usepackage{amssymb}
\usepackage{color}
\usepackage{multirow}
\usepackage{comment}

\usepackage[pagebackref,breaklinks,colorlinks]{hyperref}

\usepackage[capitalize]{cleveref}
\crefname{section}{Sec.}{Secs.}
\Crefname{section}{Section}{Sections}
\Crefname{table}{Table}{Tables}
\crefname{table}{Tab.}{Tabs.}

\def\cvprPaperID{2897} 
\def\confName{CVPR}
\def\confYear{2022}

\begin{document}
	
	\title{FedDC: Federated Learning with Non-IID Data \\
	via Local Drift Decoupling and Correction}
	
\author{Liang Gao$^1$ \quad
	Huazhu Fu$^2$ \quad
	Li Li$^{3,}$\footnotemark[4]  
	\quad
	Yingwen Chen$^{1,}$\footnotemark[4]
	\quad
	Ming Xu$^1$
	\quad
    Cheng-Zhong Xu$^3$
    \\
   \small{$^1$National University of Defense Technology, China. \quad  $^2$IHPC, ASTAR, Singapore. \quad $^3$University of Macau, IOTSC, China.} 
}

	\maketitle

\renewcommand{\thefootnote}{\fnsymbol{footnote}}
\footnotetext[4]{Li Li (LLiLi@um.edu.mo) and Yingwen Chen (ywch@nudt.edu.cn) are corresponding authors.} 

	\begin{abstract}

	Federated learning (FL) allows multiple clients to collectively train a high-performance global model without sharing their private data. However, the key challenge in federated learning is that the clients have significant statistical heterogeneity among their local data distributions, which would cause inconsistent optimized local models on the client-side. 
	To address this fundamental dilemma, we propose a novel federated learning algorithm with local drift decoupling and correction (FedDC). Our FedDC only introduces lightweight modifications in the local training phase, in which each client utilizes an auxiliary local drift variable to track the gap between the local model parameter and the global model parameters. The key idea of FedDC is to utilize this learned local drift variable to bridge the gap, i.e., conducting consistency in parameter-level. 
	The experiment results and analysis demonstrate that FedDC yields expediting convergence and better performance on various image classification tasks, robust in partial participation settings, non-iid data, and heterogeneous clients.

\end{abstract}

	\section{Introduction}
\label{Sec_Intro} 

Federated learning (FL) is an emerging distributed machine learning paradigm that leverages decentralized data from multiple clients to jointly train a shared global model under the coordination of a central server, without sharing the individuals' raw data~\cite{MOTHUKURI2021619,pmlr-v54-mcmahan17a,DBLP:journals/corr/abs-1901-08277,Kairouz2019,Feng2021}. This makes FL surpass traditional parallel optimization to avoid systemic privacy risk~\cite{2017arXiv170602677G,2015arXiv151100175I,wang2018giant,li2014communication,Li2021}. 
FedAvg~\cite{pmlr-v54-mcmahan17a} is a widely used FL aggregation algorithm, in which each client executes multiple stochastic gradient descent (SGD) steps in each communication round to minimize the local empirical risk. After that, a central server updates the parameters of the global model with the updates returned by the clients.
However, recent researches~\cite{li2020federated,li2019convergence,karimireddy2021scaffold} demonstrate that FedAvg could not converge well  with heterogeneous data (non-iid). The data distribution of clients in FL can be highly differential because clients independently collect the local data with their own preferences and sampling space. 
The non-iid distributed data leads to inconsistency in clients' local objective functions and optimization directions. 
The studies in~\cite{khaled2020tighter,karimireddy2021scaffold} prove that the data heterogeneity introduces drift in clients' local updates, which slows down the convergence speed. The parameter drift between an FL model and a centralized learning model comes from two parts: the residual parameter drift in the last round, and the gradient drift in the current round~\cite{zhao2018federated}. 
Due to the difference in data distribution, there is a fundamental contradiction between minimizing local empirical loss and reducing global empirical loss. Therefore, in a highly heterogeneous environment, FedAvg lacks a convergence guarantee, which only obtains compromised convergence speed and model performance.

To address this client drift, some methods have been proposed to reduce the variance of local updates~\cite{karimireddy2021scaffold,li2020federated}. For example, FedProx~\cite{li2020federated} adds a proximal term to force reduction of model differences between local and the global model. However, the proximal term hinders the global model from moving towards the global stationary point.
Scaffold~\cite{karimireddy2021scaffold} corrects client-drift with a control gradient variate. 
However, it only approximately reduces the gradient drift in each round but it is not able to eliminate it. 
The residual deviation would be accumulatively amplified during training according to the research of \cite{zhao2018federated}, which is the primary factor that slows down the convergence speed and causes lower performance. 
In fact, most of the previous FL methods force the local models to be consistent to the global model. They finally get a model that neglects the inconsistency between local objectives and global objectives. They have a certain effect by reducing gradient drift, but the gradually enlarged parameter deviation persists.

We admit the fact that the local optimal points of clients are fundamentally inconsistent with the global optimal point in the heterogeneous FL setup. The local stationary points of clients can be arbitrarily different from the global stationary point. Based on this observation, we propose a new \textit{federated learning algorithm with local drift decoupling and correction (FedDC)}, to handle the inconsistent objectives with auxiliary drift variables to track the local parameter drift between the local models and the global model. 
Our FedDC dynamically updates the local objective function of each client, which contains (1) a constraint penalty term that indicates the relationship among the global parameter, drift variables and the local parameters, and (2) a gradient correction term to reduce the gradient drift in each training round. 
We decouple the local models and the global model in the training process by introducing the drift variables, which reduces the impact of the local drift to the global objective and makes it converge quickly and reach better performance.
%
We execute experiments on several public datasets, including MNIST, fashion MNIST, CIFAR10, CIFAR100, EMNIST-L, tiny ImageNet and a synthetic dataset. The results demonstrate that our FedDC achieves the best performance and significantly faster converge speed compared with the competing FL methods (e.g., FedAvg~\cite{pmlr-v54-mcmahan17a}, FedProx~\cite{li2020federated}, Scaffold~\cite{karimireddy2021scaffold} and FedDyn~\cite{acar2021federated}) in both iid and non-iid client settings\footnote{The code is available at \url{https://github.com/gaoliang13/FedDC}}.
	\section{Related Work}

Recently, FL has become a hot research topic~\cite{2016arXiv161002527K,Kairouz2019,Dayan2021}. As a pioneering work, FedAVG \cite{pmlr-v54-mcmahan17a} conducts weighted parameter averaging in order to update parameters from multiple clients. The works in~\cite{khaled2020tighter,2019arXiv190411325K} show that FedAvg reaches asymptotic convergence for homogeneous clients. However, Woodworth \textit{et al.}~\cite{woodworth2018graph} demonstrate that the bound of FedAVG convergence can be totally different for heterogeneous clients. The studies in~\cite{li2019convergence,karimireddy2021scaffold} claim that the client drift in clients' updates caused by non-iid data is the main culprit that damages convergence rates in heterogeneous settings.
Prior works have shown that non-iid data would introduce challenges in FL such as gradient divergence, optimization direction biases, and unguaranteed convergence. Some works try to reduce the variance of clients' updates to speed up convergence. 
Minimizing the empirical risk function using a uniform global model over different clients which contains non-iid distributed data makes it difficult to converge to a splendid global model. 
FedProx~\cite{li2020federated} surmounts statistical heterogeneity and strengthens stability by adding a proximal regularization on the local model against the global model. The proximal term keeps the updated local parameter close to the global model, in this way it reduces potential gradient divergence. However, it violated the fact that the optimal points of local empirical objectives are different from the global optimal point, leading to a low performance. 
\textit{The major limitation of these methods is that they ignore the differences in client models, leading to sub-optimal performance and slowly converge speed in non-iid data distributions.}

In order to further analyze the correlation between client drift and data heterogeneity, some works conduct personalized local objectives with statistical variables.
Scaffold \cite{karimireddy2021scaffold} customizes gradients for each client to fix the client drifts between local models and the global model. Similarly, FedDyn \cite{acar2021federated} proposes a dynamic regularizer for each device to align the global and device solutions and save transmission costs. 
Another type of work tries to optimize the parameter aggregation step on the central server to get a better global model. \cite{zhang2021personalized} dynamic calculates the optimally weighted combination of clients' local model by figuring out how much a client can benefit from the global model. Reddi \textit{et al.} \cite{DBLPjournalscorrabs-2003-00295} propose federated adaptive optimization based on the interplay between client heterogeneity and communication efficiency to prevent unfavourable convergence behaviour. Yang \textit{et al.} \cite{yang2021achieving} achieve linear speedup with non-iid data with two-sided learning rates in local update and global update. These methods are compatible with our method, which could be easily integrated into our method.     
These improved methods achieve better speedup in convergence and enjoy better performance than FedAvg.
%
%
However, the theory of Zhao \textit{et al.} \cite{zhao2018federated} indicates that the parameter deviation would be accumulated and cause a sub-optimal solution. \textit{In this paper, we propose the FedDC, which decouples the local and global models by tracking and bridging the local drift.}


%

\section{Local Drift in Federated Learning}

In FL, we assume that there are $N$ clients in a federation, and suppose $D_i$ is client $i$'s private local dataset. The goal is to get a global model $w^*$ training over the global dataset $D=\bigcup_{i\in{[N]}} D_i$ that solves the objective: 
\vskip -10pt
\begin{equation}\label{fedavg_loss} 
w^*=\arg\min_{w}L(w)= \sum_{i=1}^N \frac{|D_i|}{|D|}L_i(w),
\end{equation}
where $w$ is the parameter of the global model, $L(w)$ is the empirical loss on the global dataset $D$, $|D_i|$ is the number of samples on $D_i$, $|D|$ is the number of samples on $D$, $L_i(w)=\mathbb{E}_{(x,y)\in D_i}l(w;(x,y))$ is the local empirical loss on client $i$'s local dataset $|D_i|$.
In order to avoid privacy leaking, any client can not share its raw data with others. FedAvg is proposed to coordinate multiple clients to cooperatively train the global model with a central server while preserving data privacy~\cite{pmlr-v54-mcmahan17a}.
Specifically, in FedAvg, for each training round, all clients optimize their local models on the local datasets, then the server takes the expectation of the local model parameters to update the global model as follows:
\vskip -10pt
\begin{equation}\label{fedavg_global}
w= \sum_{i=1}^N \frac{|D_i|}{|D|}\theta_i,
\end{equation}
where $w$ is the global model parameter, $\theta_i$ is client $i$'s local model parameters. 
Then, the updated global model parameter is broadcast to clients and utilized as the start point of local models in the next round.

\begin{figure}[!t]
	\centering 
	\includegraphics[width=0.6\linewidth]{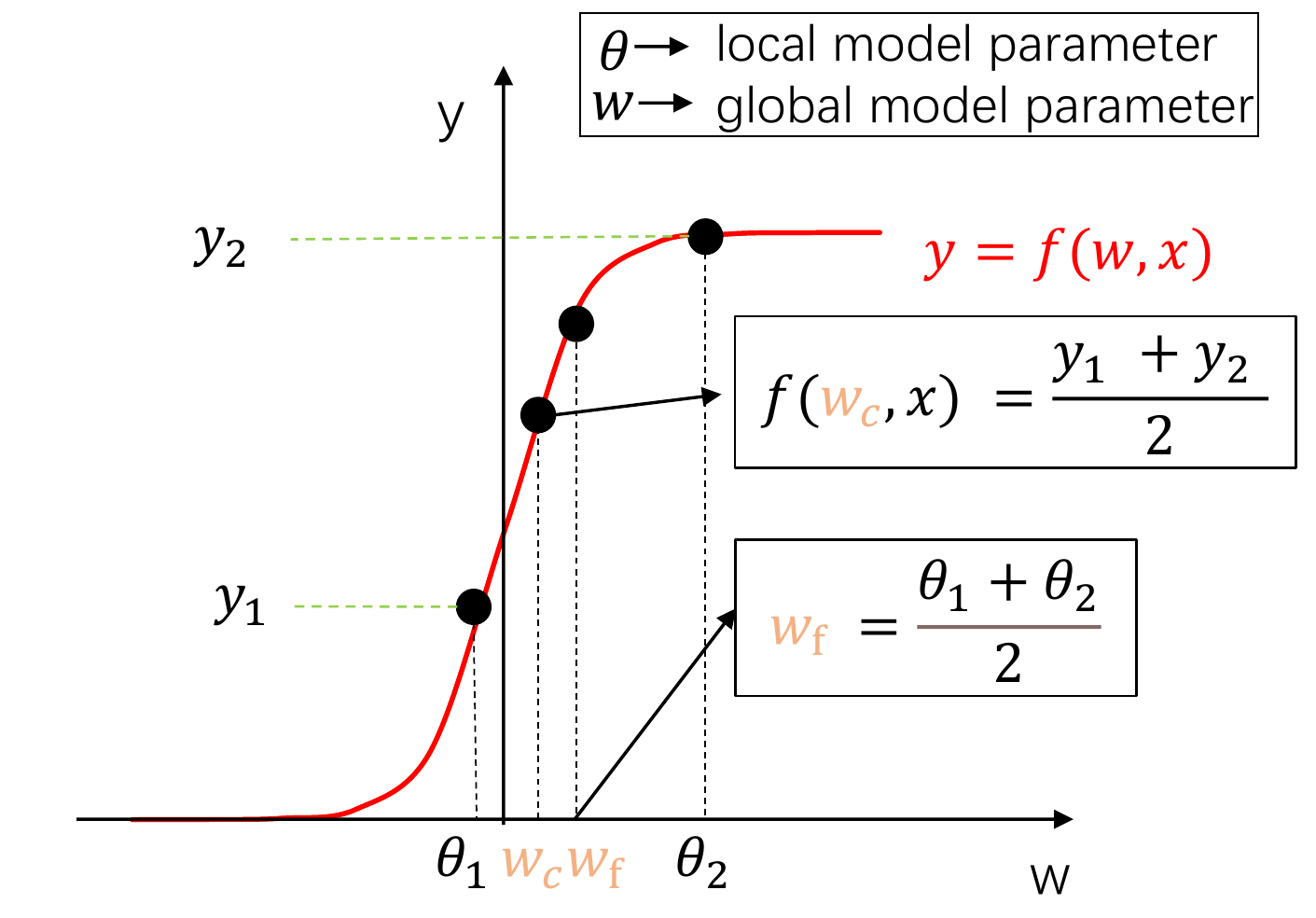}
	\vskip -5pt
	\caption{Illustration of the local drift in FedAvg with a $Sigmod$ activation function $f$. $w_c$ is the parameter of the model trained with centralized data (ideal model), $w_f$ is the parameter of the model generated by FedAvg. $\theta_1$ and $\theta_2$ are the parameters of local models of client $1$ and client $2$, respectively.}
	\label{fig_tsne}
	\vskip -10pt
\end{figure}

There is a drift between each client's local model trained on the local dataset and the global model trained on the global dataset directly \cite{NEURIPS2020_564127c0, li2020federated}. If the drift is ignored, the server would get an skewed global model.
%
%
%
FL faces the challenge of heterogeneous data.
With the highly skewed non-IID data in FL, the performance of FedAvg is significantly reduced \cite{NEURIPS2020_564127c0, li2020federated}, which indicates that the FedAvg method that ignores local drift leads to the deviation of the global model.
In Figure \ref{fig_tsne}, we show a simple example that client's local drift would result in a biased global model in FedAvg. 
We suppose that there is a non-linear transformation function $f$ (e.g., $Sigmoid$ function in the activation layer) in the model. Suppose $\theta_1$ and $\theta_2$ are local parameters of client $1$ and client $2$, $w_c$ is the ideal model parameter and $w_f$ is the model parameter generated through FedAvg. The local drifts (denoted as $h$) of client $1$ and client $2$ are $(h_1=w_c-\theta_1)$ and $(h_2=w_c-\theta_2)$, respectively. $x$ is a data point, the corresponding outputs on client $1$ is $y_1= f(\theta_1, x)$ and $y_2= f(\theta_2, x)$ on client $2$. Then the model parameter generated by FedAvg can be represented as $w_f= \frac{\theta_1+\theta_2}{2}$. The centralized model is an ideal model that would get the ideal output, that is, $f(w_c, x)=\frac{y_1+y_2}{2}$. Thus, the parameter of centralized model is $w_c=f^{-1}(\frac{y_1+y_2}{2})/x$, where $f^{-1}$ is the inverse function of $f$. Since $f$ is a non-linear function, we have $w_f\neq w_c$ and $f(w_f, x) \neq \frac{y_1+y_2}{2}$.
That indicates the global model in FedAvg is skewed, which is likely to converge slowly and with poor accuracy.
Therefore, \textit{we can learn the local drift between the global model and the local model, and bridge the local drift before uploading the local model parameters to the server.} This is in line with the intuition of FL.


\section{Proposed Method}
Based on the above observation, we propose a novel federated learning algorithm with local drift decoupling and correction (FedDC), which aims to improve the robustness and speed of model convergence by learning the model drift and bridging the drift on the client-side.
Our FedDC introduces lightweight modifications in the training phase to decouple the global model from clients' local models using the local drift.
Specifically, in the local training phase, each client learns a local drift variable that represents the gap between its local model and the global model.
Then, the local drift variable is used to correct the local model parameters before the parameter aggregating phase. In this way, FedDC decreases the distance between the local model parameters and the global model parameters, which also decreases the negative influence of the skewed local model on the global model.

\subsection{Objectives in FedDC}
First, we define a local drift variable $h_i$ for each client. In an ideal condition, the local drift variable should satisfy the restriction: $h_i=w-\theta_i$, where $\theta_i$ is the parameter of client $i$'local model, and $w$ is the parameter of the global model.
In the whole training process, we need to keep this restriction to prevent the local drift variable from getting out of our control. Therefore, for client $i$, we further convert this restriction as a penalized term as:
\begin{equation} 
\setlength{\abovedisplayskip}{3pt}
\setlength{\belowdisplayskip}{3pt}
R_i(\theta_i,h_i,w) =||h_i+\theta_i-w||^2, \forall i \in [N].
\label{eq-penalized}
\end{equation}
Each client utilizes this penalized term with its empirical loss term on the corresponding dataset to train the model parameters and the local drift variables. In this way, we transform an equation-constrained optimization problem into an unconstrained optimization problem. 

In FedDC, the objective function of each client contains three components: the local empirical loss term, the penalized term, and a gradient correction term. 
Specifically, for client $i$ ($\forall i \in [N]$), the local objective of $\theta_i$ is to minimize the following objective function:
\begin{equation}\label{local_loss}
\setlength{\abovedisplayskip}{3pt}
\setlength{\belowdisplayskip}{3pt}
F(\theta_i; h_i,D_i,w)=L_i(\theta_i)+ \frac{\alpha}{2}R_i(\theta_i;w_i,w) + G_i(\theta_i;g_i,g),
\end{equation}
where $L_i$ is the typical empirical loss,
$R_i$ is the penalized term in Eq.~\ref{eq-penalized}, $\alpha$ is a hyper-parameter that controls the weight of $R_i$, 
and $G_i$ is the gradient correction term that controls the gradient stochastic optimization.
Inspired by Scaffold \cite{karimireddy2021scaffold}, we set the gradient correction term as $G_i(\theta_i;g_i,g)=\frac{1}{\eta K}\langle\theta_i,g_i-g\rangle,$ where $\eta$ is the learning rate, $K$ is the amount of training iterations in one round. $g_i$ is the local update value of $i$-th client's local parameters in last round, $g$ is the average update value of all clients' local parameters in last round. In $t$-th round, we have $g_i=\theta_i^t-\theta_i^{t-1}$ and $g=\mathbb{E}_{i\in[N]}g_i$, where $\theta_i^t$ and $\theta_i^{t-1}$ are client $i$'s local model parameters in $t$-th round and $(t-1)$-th round, respectively. The role of term $G_i$ is to reduce the variance of local gradients. 

\textbf{Updating the local model parameters.}
At the beginning of each round, the server first sends the global parameters of the previous round to all clients. 
Each client $i$ ($\forall i \in [N]$) loads the global model parameter to the local model (set $\theta_i=w$) and then updates the local model by minimizing the objective function in Eq.~\ref{local_loss}. We assume each training round contains $K$ local training iterations, in $k$-th local training iteration of $t$-th round, the local model parameter is updated as follows:
\begin{equation}\label{eq_update_theta}
\setlength{\abovedisplayskip}{3pt}
\setlength{\belowdisplayskip}{3pt}
\theta_i^{t,k+1}=\theta_i^{t,k}-\eta \frac{\partial F(\theta_i^{t,k}; h_i^{t},D_i,w^{t})}{\partial \theta_i^{t,k}},
\end{equation}
where $\eta$ is the learning rate.
The Eq.~\ref{eq_update_theta} is executed $K$ times in each round.

\textbf{Updating the local drift variables.}
Then, we introduce the updating method of the local drift variable $h_i$. We use the superscripted $^+$ symbol to indicate the updated parameters at $K$-th local iteration. In FedDC, the local drift variables track the gaps between local models and the global model. In a training round, we suppose the global model parameter $w$ is updated to $w^+$ while the local model parameter is fixed.
Then we can update the local drift variable using $h_i^+ = h_i+(w_i^+-w)$.
However, it is impossible to update the global model directly due to the unavailable global data.

Another way to optimize $h_i$ is minimizing the objective loss using the partial derivative of $h_i$ in Eq.~\ref{local_loss} with $\theta_i$ and $w$ fixed on the client-side. However, that costs $K$ training iterations of back-propagation.
In order to reduce the calculation, assuming that we have first updated the local model parameters from $\theta_i$ to $\theta_i^+$ which is a must-do step. Then we consider the following two points: 1) at the beginning of each round, the local model parameters is assigned with the global model parameter: $\theta_i=w$. 2) for client $i$, the local model parameter $\theta_i^+$ is an estimation of the updated global model parameter $w_i^+$. Thus, instead of $h_i^+ = h_i+(w_i^+-w)$, we can approximately update the local drift variable using:
\begin{equation}\label{eq_update_h}
\setlength{\abovedisplayskip}{3pt}
\setlength{\belowdisplayskip}{3pt}
h_i^+ = h_i+(w_i^+-w_i) \approx h_i+(\theta_i^+-\theta_i),
\end{equation}
where $\theta_i$ in Eq.~\ref{eq_update_h} is the shorthand of $\theta_i^{t,0}$ and $\theta_i^+$ the shorthand of $\theta_i^{t,K}$ in $t$-th round. 
In this way, we reuse the updates of local model parameter to update the local drift and avoid performing the back-propagation process for $h_i$.

\textbf{Updating the global model parameters.}
To update the global model parameters, before the model aggregation phase each client corrects its local model parameters using the local drift variables: ($\theta_i^++h_i^+$). Then each client uploads the corrected local parameters to the server. Similar to FedAvg, the server performs a weighted average of the corrected local parameters to obtain the global model parameters:
\begin{equation}\label{FedDC_aggragate}
\setlength{\abovedisplayskip}{3pt}
\setlength{\belowdisplayskip}{3pt}
w^+= \sum_{i=1}^N \frac{|D_i|}{|D|}(\theta_i^++h_i^+),
\end{equation}
where $|D_i|$ is the sample amount on client $i$, $w^+$ is the updated global model.

\begin{figure}[!t]
	\centering  
	\includegraphics[width=1\linewidth]{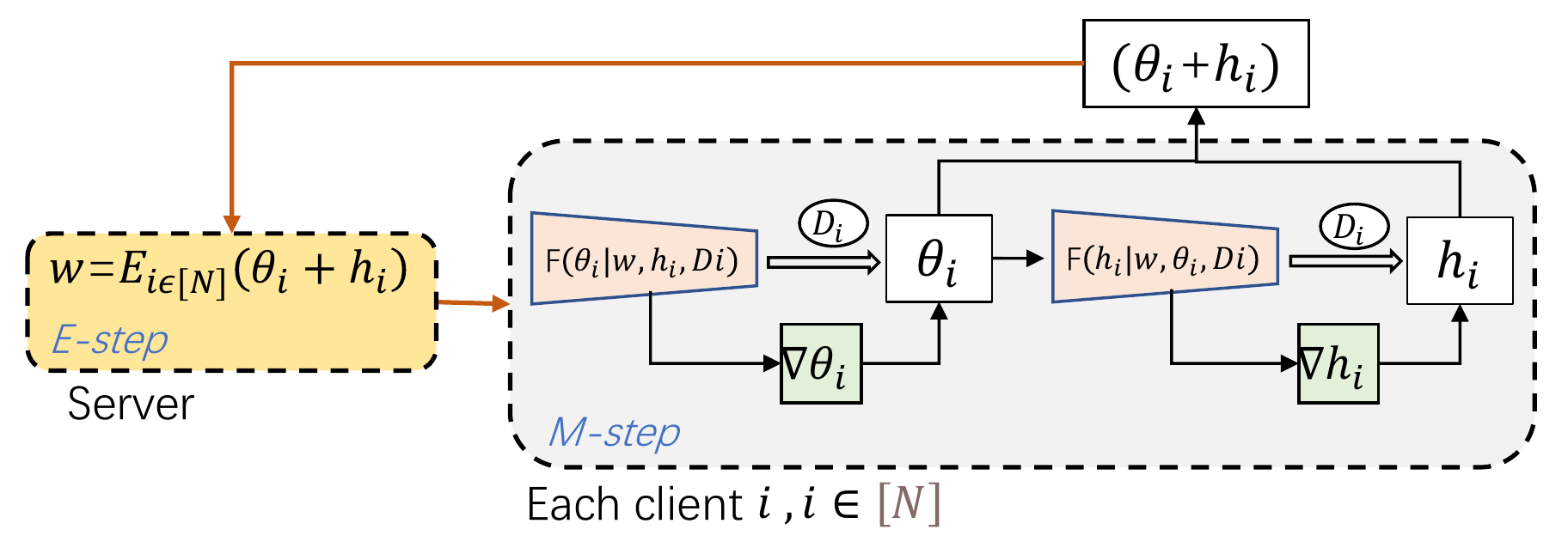}
	\vskip -5pt
	\caption{The training procedure of FedDC using Expectation-Maximum (EM) algorithm. In each round, the local parameters and the global parameters are iteratively updated on the client-side (M-step) and the server-side (E-step) respectively.}
	\label{fig_schedual} \vskip -15pt
\end{figure}

\subsection{Training Process}
We summarize the training procedure of FedDC with the Expectation-Maximum (EM) algorithm. 
The EM algorithm is used to solve the parameter optimization problem in the case there is missing information.  
In FedDC, the traditional machine learning method of directly optimizing parameters is not applicable as there are three types of variables. Moreover, the local parameters and the global parameters are updated on different devices. We can iteratively fix two variables while optimizing the other one at a time. In this way, we seek the extreme value of one variable one step, and finally, approach the extreme value of these variables step by step. The training process of FedDC is shown in Figure \ref{fig_schedual}.
In each round, we execute the Maximization step (M-step) on the client-side to optimize the local model parameter $\theta_i$ and the local drift variables $h_i$. Then we execute the Expectation step (E-step) on the server-side to update the global model parameter $w$. 

\subsection{Convergence of FedDC}
We proved the convergence of FedDC in non-convex case. For non-convex and $\beta$-Lipschitz smooth local empirical loss function $L_i,\forall i\in[N]$, there exists a $\beta_d>0$, where $\bar\alpha=\alpha-\beta_d>0$ and $\nabla^2 L_i\geq -\beta_dI$. We assume the local empirical loss $L_i$ is non-convex and $B$-dissimilarity, in which $B(\theta^t)\leq B$. The global empirical loss of FedDC decreases as follows: 
\vskip -15pt
\begin{equation}{\label{lammap}} \mathbb{E}_{C_t} L(w^t)\leq L(w^{t-1})-2p||\nabla L(w^{t-1})||^2, 
\end{equation}
where $p=(\frac{\gamma}{\alpha}-\frac{B(1+\gamma)\sqrt{2}}{\bar\alpha \sqrt{N}}-\frac{\beta B(1+\gamma)}{\alpha\bar\alpha}-\frac{\beta(1+\gamma)^2B^2}{2\bar\alpha^2}-\frac{\beta B^2(1+\gamma)^2(2\sqrt{2C}+2)}{\bar\alpha^2 N})>0,$
and $C_t$ is the active client set in round $t$ which contains $C$ clients.
The more details of the convergence guarantee are provided in Appendix~\textcolor{red}{B}.

\subsection{Discussion}
Our FedDC appears has similar goal with the previous methods like SCAFFOLD, FedProx and FedDyn as they all try to reduce the gap between the local model parameters and the global model parameters caused by non-iid data, but there are fundamental differences. 
The general approach of the previous methods (e.g. SCAFFOLD, FedProx and FedDyn) is to limit the local optimization direction to reduce the parameter gap between the local models and the global model, that is, restricting $\theta_i$ to be close to $w$ (that is $\min ||\theta_i-w||$). 
However, restricting the optimization direction of the local model hinder it in fitting the local dataset distributions, because the local distribution and global distribution can be inconsistent. 
In FedDC, we think learning the parameter gap is better than limiting it.
FedDC utilizes the local drift variable to learn the parameter gap between the local model and the global model. And then the local drift variable is used to bridge the gap, where we learn the local drift $h_i$ to achieve the goal $\min ||\theta_i+h_i-w||$.
In other words, FedDC does not hinder the local models from learning local features and minimizing the local empirical risks. We attribute the advantages of FedDC to that it learns the local drift and well bridges the parameter gap without hindering the local training process.

	\section{Experiments}
\label{Sec_exp}

In this section, we evaluate the effectiveness of FedDC and compare FedDC with several advanced methods in various datasets and settings. Specifically, the evaluation is mainly conducted from two perspectives: 1) convergence speed and 2) model accuracy. 
Due to the space limitation, more detailed experiment results and the ablation study are given in Appendix~\textcolor{red}{A}.

\subsection{Dataset and Baselines}
We explore on six benchmark datasets: MNIST \cite{726791}, fashion MNIST \cite{xiao2017fashion}, CIFAR10, CIFAR100 \cite{Krizhevsky2009LearningML}, EMNIST-L \cite{7966217}, Tiny ImageNet \cite{pouransari2014tiny} and the Synthetic \cite{li2020federated} datasets. For all of them, we adopt the same training/testing splits as previous works \cite{acar2021federated,pmlr-v54-mcmahan17a,li2020federated}. 
In the iid setting, training samples are randomly selected and equally assigned to clients. 
All the clients have the same amount of training data, and each client's data points are evenly distributed in all categories. 
In the non-iid data settings, the label ratios follow the Dirichlet distribution \cite{yurochkin2019bayesian}. We set two non-iid data settings, and they are denoted as D1 and D2 in which the Dirichlet parameters are 0.6 and 0.3 respectively. 
Besides, we produce unbalanced data by samplings samples with a lognormal distribution, in which we set the variance as $0.3$. 
For the Synthetic dataset, following the setting in \cite{acar2021federated}, we generate three types of data settings, including homogeneity setting which is denoted as "Synthetic(0,0)", objective heterogeneity setting which is denoted as "Synthetic(1,0)", data heterogeneity setting which denoted as "Synthetic(0,1)".
More detailed settings are given in Appendix~\textcolor{red}{A}.

We verify the experimental results based on four network architectures in order to emphasize the versatility of the proposed method. We use a multi-class logistic classification model for the Synthetic dataset. For the MNIST digit classification task, the same fully-connected network (FCN) is adopted as~\cite{pmlr-v54-mcmahan17a}. A convolutional neural network (CNN) is adopted to classify the samples on CIFAR10 and CIFAR100, as used in~\cite{pmlr-v54-mcmahan17a}. On Tiny ImageNet, a pre-trained ResNet18 \cite{he2016deep} is adopted to show the efficiency of FedDC on the pre-trained models.


We compare FedDC with several advanced methods, including FedAvg \cite{pmlr-v54-mcmahan17a}, FedProx \cite{li2020federated}, Scaffold \cite{karimireddy2021scaffold} and FedDyn \cite{acar2021federated}. 
FedProx uses the proximal term to reduce the gradient variance. Scaffold attempts to correct the local updates with a gradient correction term, and FedDyn aligns the client models using a dynamic regularizer.
Different from FedDC, these methods all emphasize the consistency of client models and the global model and ignore the local drift in the parameter aggregation phase. 


\subsection{Hyper-parameter Settings}
We apply the typical FL architecture, where multiple clients get their local updates in each communication round through training models with their local datasets, and a central server aggregates client updates to update the global model. We utilize the SGD algorithm as the local optimizer for all methods. In addition, in order to maintain consistency, for all methods on the true world datasets, we set batch size as $50$ in the local training phase, the local training epochs as $5$ in each round, the initial learning rate as $0.1$, and the decay rate as $0.998$. All the above settings follow the previous work \cite{acar2021federated}. We set the hyper-parameter $\alpha=0.01$ of FedDC on CIFAR10, CIFAR100 and Tiny ImageNet, $\alpha=0.1$ of FedDC on MNIST, fashion MNIST and EMNIST-L. In the Synthetic dataset, we set the number of clients as $20$ and the local batch size as $10$, $\alpha=0.005$ for FedDC. As for specific hyper-parameters of the baselines, we keep the same settings as their referred papers. We set FedDyn's hyper-parameter $\alpha=0.01$ and FedProx's hyper-parameter $\mu=10^{-4}$. If there are parameter settings different from the above described, it will be specifically explained in the Appendix. We also explore the effect of different values of $\alpha$ in FedDC (See Appendix~\textcolor{red}{A}).

\begin{table*}[!t]
	\footnotesize
	\centering \vskip -10pt
	\caption{
		The communication rounds in different methods to achieve the same target accuracy. The left half is the result of full participation, and the right is the result of partial participation, each of which includes one iid setting and two non-iid settings where the 0.6-Dirichlet non-iid setting is denoted as "D1", and the 0.3-Dirichlet non-iid setting is denoted as "D2".
		In addition, we denote the communication round of each method to achieve the target accuracy as "$R\#$", the corresponding convergence speedup relative to FedAvg as "$S\uparrow$". We use $\textgreater{}$ sign to represent the method that could not achieve the target accuracy within the communication constraint.}
		\vskip -5pt
	\label{tab_communication_saving_fashion}
		\begin{tabular}{|ccccccccccccc|}
			\hline
			\multicolumn{1}{|c|}{\multirow{3}{*}{Model}} & \multicolumn{6}{c|}{Full Participation} & \multicolumn{6}{c|}{Partial Participation (15\%)} \\ \cline{2-13} 
			\multicolumn{1}{|c|}{} & \multicolumn{2}{c|}{D1} & \multicolumn{2}{c|}{D2} & \multicolumn{2}{c|}{iid} & \multicolumn{2}{c|}{D1} & \multicolumn{2}{c|}{D2} & \multicolumn{2}{c|}{iid} \\ \cline{2-13} 
			\multicolumn{1}{|c|}{} & \multicolumn{1}{c|}{$R\#$} & \multicolumn{1}{c|}{$S\uparrow$} & \multicolumn{1}{c|}{$R\#$} & \multicolumn{1}{c|}{$S\uparrow$} & \multicolumn{1}{c|}{$R\#$} & \multicolumn{1}{c|}{$S\uparrow$} & \multicolumn{1}{c|}{$R\#$} & \multicolumn{1}{c|}{$S\uparrow$} & \multicolumn{1}{c|}{$R\#$} & \multicolumn{1}{c|}{$S\uparrow$} & \multicolumn{1}{c|}{$R\#$} & $S\uparrow$ \\ \hline
			\multicolumn{13}{|c|}{MNIST, 100 client, Target accuracy $98\%$} \\ \hline
			\multicolumn{1}{|c|}{FedAvg} & 258 & - & 492 & - & 142 & \multicolumn{1}{c|}{-} & 361 & - & \textgreater{}600 & - & 158 & - \\
			\multicolumn{1}{|c|}{FedProx} & 263 & 0.98$\times$ & 480 & 1.03$\times$ & 136 & \multicolumn{1}{c|}{1.04$\times$} & 383 & 0.94$\times$ & 418 & 1.44$\times$ & 149 & 1.06$\times$ \\
			\multicolumn{1}{|c|}{Scaffold} & 58 & 4.45$\times$ & 58 & 8.48$\times$ & 53 & \multicolumn{1}{c|}{2.68$\times$} & 62 & 5.82$\times$ & 72 & 8.33$\times$ & 50 & 3.16$\times$ \\
			\multicolumn{1}{|c|}{FedDyn} & 46 & 5.61$\times$ & 51 & 9.65$\times$ & 27 & \multicolumn{1}{c|}{5.26$\times$} & 122 & 2.96$\times$ & 153 & 3.92$\times$ & 71 & 2.23$\times$ \\
			\multicolumn{1}{|c|}{FedDC} & 35 & 7.37$\times$ & 37 & 13.3$\times$ & 26 & \multicolumn{1}{c|}{5.46$\times$} & 60 & 6.02$\times$ & 62 & 9.68$\times$ & 46 & 3.43$\times$ \\ \hline
			\multicolumn{13}{|c|}{fashion MNIST, 100 client, Target accuracy $89\%$} \\ \hline
			\multicolumn{1}{|c|}{FedAvg} & \textgreater{}300 & - & 273 & - & 112 & \multicolumn{1}{c|}{-} & \textgreater{}300 & - & \textgreater{}300 & - & 144 & - \\
			\multicolumn{1}{|c|}{FedProx} & \textgreater{}300 & 1$\times$ & \textgreater{}300 & 0.91$\times$ & 130 & \multicolumn{1}{c|}{0.86$\times$} & \textgreater{}300 & 1$\times$ & \textgreater{}300 & 1$\times$ & 128 & 1.13$\times$ \\
			\multicolumn{1}{|c|}{Scaffold} & 117 & 2.56$\times$ & 169 & 1.61$\times$ & 85 & \multicolumn{1}{c|}{1.32$\times$} & 133 & 2.26$\times$ & \textgreater{}300 & 1$\times$ & 108 & 1.33$\times$ \\
			\multicolumn{1}{|c|}{FedDyn} & 150 & 2$\times$ & 211 & 1.29$\times$ & 38 & \multicolumn{1}{c|}{2.95$\times$} & \textgreater{}300 & 1$\times$ & 267 & 1.12$\times$ & 85 & 1.69$\times$ \\
			\multicolumn{1}{|c|}{FedDC} & 86 & 3.49$\times$ & 126 & 2.17$\times$ & 24 & \multicolumn{1}{c|}{4.67$\times$} & 87 & 3.49$\times$ & 252 & 1.19$\times$ & 63 & 2.29$\times$ \\ \hline
			\multicolumn{13}{|c|}{EMNIST-L, 100 client, Target accuracy $94\%$} \\ \hline
			\multicolumn{1}{|c|}{FedAvg} & 142 & - & 192 & - & 107 & \multicolumn{1}{c|}{-} & 153 & - & 245 & - & 108 & - \\
			\multicolumn{1}{|c|}{FedProx} & 135 & 1.05$\times$ & 198 & 0.97$\times$ & 92 & \multicolumn{1}{c|}{1.16$\times$} & 145 & 1.06$\times$ & 240 & 1.02$\times$ & 105 & 1.03$\times$ \\
			\multicolumn{1}{|c|}{Scaffold} & 43 & 3.30$\times$ & 52 & 3.69$\times$ & 30 & \multicolumn{1}{c|}{3.57$\times$} & 44 & 3.48$\times$ & 68 & 3.6$\times$ & 42 & 2.57$\times$ \\
			\multicolumn{1}{|c|}{FedDyn} & 30 & 4.73$\times$ & 52 & 3.69$\times$ & 27 & \multicolumn{1}{c|}{3.96$\times$} & 73 & 2.1$\times$ & 81 & 3.06$\times$ & 61 & 1.61$\times$ \\
			\multicolumn{1}{|c|}{FedDC} & 43 & 3.3$\times$ & 60 & 3.2$\times$ & 21 & \multicolumn{1}{c|}{5.1$\times$} & 48 & 3.19$\times$ & 74 & 3.31$\times$ & 47 & 2.3$\times$ \\ \hline
			\multicolumn{13}{|c|}{CIFAR10, 100 client, Target accuracy $80\%$} \\ \hline
			\multicolumn{1}{|c|}{FedAvg} & \textgreater{}1000 & - & \textgreater{}1000 & - & 286 & \multicolumn{1}{c|}{-} & 616 & - & \textgreater{}1000 & - & \textgreater{}1000 & - \\
			\multicolumn{1}{|c|}{FedProx} & 474 & 2.11$\times$ & \textgreater{}1000 & 1$\times$ & 277 & \multicolumn{1}{c|}{1.03$\times$} & 459 & 1.34$\times$ & \textgreater{}1000 & 1$\times$ & 307 & 3.28$\times$ \\
			\multicolumn{1}{|c|}{Scaffold} & 165 & 6.06$\times$ & 218 & 4.59$\times$ & 120 & \multicolumn{1}{c|}{2.38$\times$} & 200 & 3.08$\times$ & 263 & 3.80$\times$ & 126 & 7.93$\times$ \\
			\multicolumn{1}{|c|}{FedDyn} & 60 & 16.67$\times$ & 75 & 17.54$\times$ & 55 & \multicolumn{1}{c|}{5.2$\times$} & 193 & 3.19$\times$ & 195 & 5.12$\times$ & 145 & 6.9$\times$ \\
			\multicolumn{1}{|c|}{FedDC} & 53 & 18.86$\times$ & 70 & 14.28$\times$ & 43 & \multicolumn{1}{c|}{6.65$\times$} & 141 & 4.37$\times$ & 143 & 6.99$\times$ & 108 & 9.26$\times$ \\ \hline
			\multicolumn{13}{|c|}{CIFAR100, 100 client, Target accuracy $40\%$} \\ \hline
			\multicolumn{1}{|c|}{FedAvg} & 476 & - & 847 & - & \textgreater{}1000 & \multicolumn{1}{c|}{-} & 615 & - & 520 & - & 724 & - \\
			\multicolumn{1}{|c|}{FedProx} & 502 & 0.95$\times$ & 507 & 1.67$\times$ & 273 & \multicolumn{1}{c|}{3.66$\times$} & 980 & 0.63$\times$ & 503 & 1.03$\times$ & 650 & 1.11$\times$ \\
			\multicolumn{1}{|c|}{Scaffold} & 91 & 5.23$\times$ & 94 & 9.01$\times$ & 84 & \multicolumn{1}{c|}{11.9$\times$} & 106 & 5.8$\times$ & 114 & 3.56$\times$ & 113 & 6.41$\times$ \\
			\multicolumn{1}{|c|}{FedDyn} & 51 & 9.33$\times$ & 53 & 15.98$\times$ & 56 & \multicolumn{1}{c|}{17.85$\times$} & 149 & 4.42$\times$ & 148 & 3.51$\times$ & 143 & 5.06$\times$ \\
			\multicolumn{1}{|c|}{FedDC} & 39 & 12.2$\times$ & 41 & 20.65$\times$ & 37 & \multicolumn{1}{c|}{27.03$\times$} & 102 & 6.03$\times$ & 103 & 5.05$\times$ & 100 & 7.04$\times$ \\ \hline
		\end{tabular}
	\vskip -10pt
\end{table*}

\begin{table}[!t]
	\centering 
	\footnotesize
	\caption{The top-1 test accuracy on Tiny ImageNet with 20 clients training for 10 rounds on iid and non-iid settings.}
	\vskip -5pt
	\label{figure_tiny}
	\begin{tabular}{|l|l|l|l|}
		\hline
		Method & D1 & D2 & iid \\ \hline
		FedAvg & 43.86 & 42.62 & 44.30 \\
		FedProx & 43.55 & 42.25 & 44.11 \\
		Scaffold & 44.38 & 43.38 & 45.07 \\
		FedDyn & 45.37 & 44.71 & 45.61 \\
		FedDC & \textbf{46.44} & \textbf{46.60} & \textbf{47.91} \\ \hline
	\end{tabular}
	\vskip -15pt
\end{table}

\begin{table*}[!t]
	\footnotesize
	\centering \vskip -10pt
	\caption{The top-1 test accuracy (\%) on iid, non-iid and unbalanced data for full client participation and partial client participation ($15\%$) levels. There are three settings for the amount of clients: Setting 1 (100 clients), Setting 2 (500 clients) and Setting 3 (20 clients).
	}
	\vskip -5pt
	\label{tab-best-accuracy}
	\begin{tabular}{|l|lllll|lllll|}
		\hline
		Method & FedAvg & FedProx & Scaffold & FedDyn & FedDC & FedAvg & FedProx & Scaffold & FedDyn & FedDC \\\hline
		\multicolumn{1}{|l|}{\textbf{Setting 1}} & \multicolumn{5}{c|}{100 clients full participation} & \multicolumn{5}{c|}{100 clients partial participation} \\ \hline
		CIFAR10-iid & 82.16 & 81.85 & 84.61 & 85.26 & \textbf{86.18} & 81.67 & 82.16 & 84.68 & 84.50 & \textbf{85.71} \\
		CIFAR10-D1 & 80.42 & 80.70 & 84.13 & 85.26 & \textbf{85.64} & 81.05 & 81.32 & 83.57 & 84.10 & \textbf{84.77} \\
		CIFAR10-D2 & 79.14 & 78.89 & 82.96 & 84.14 & \textbf{84.32} & 79.77 & 79.84 & 82.53 & 82.30 & \textbf{84.58} \\
		CIFAR10-unbalance & 81.37 & 81.90 & 84.45 & 85.68 & \textbf{86.31} & 81.68 & 81.88 & 84.44 & 84.30 & \textbf{85.35} \\\hline
		CIFAR100-iid & 39.68 & 40.39 & 51.26 & 52.07 & \textbf{55.52} & 40.80 & 40.67 & 49.80 & 51.20 & \textbf{55.40} \\
		CIFAR100-D1 & 40.48 & 40.15 & 51.16 & 52.84 & \textbf{55.34} & 41.76 & 41.83 & 50.01 & 51.75 & \textbf{54.65} \\
		CIFAR100-D2 & 40.11 & 40.93 & 50.44 & 51.89 & \textbf{54.86} & 41.81 & 41.84 & 50.25 & 51.13 & \textbf{53.91} \\ 
		CIFAR100-unbalance & 40.03 & 39.93 & 51.30 & 52.81 & \textbf{55.69} & 40.90 & 41.05 & 50.57 & 51.01 & \textbf{55.27} \\\hline
		MNIST-iid & 98.12 & 98.12 & 98.32 & \textbf{98.51} & 98.45 & 98.15 & 98.11 & 98.45 & 98.38 & \textbf{98.47} \\
		MNIST-D1 & 98.09 & 98.05 & 98.39 & 98.44 & \textbf{98.48} & 98.13 & 98.12 & 98.45 & 98.30 & \textbf{98.49} \\
		MNIST-D2 & 97.98 & 97.96 & 98.45 & 98.46 & \textbf{98.51} & 98.00 & 98.04 & 98.37 & 98.30 & \textbf{98.40} \\
		MNIST-unbalance & 98.12 & 98.10 & 98.35 & \textbf{98.60} & 98.46 & 98.15 & 98.13 & 98.50 & 98.34 & \textbf{98.53} \\\hline
		\multicolumn{1}{|l|}{\textbf{Setting 2}} & \multicolumn{5}{c|}{500 clients full participation} & \multicolumn{5}{c|}{500 clients partial participation} \\ \hline
		CIFAR10-iid & 73.43 & 72.77 & 81.56 & 84.07 & \textbf{84.93} & 73.26 & 72.58 & 81.58 & 82.49 & \textbf{84.19} \\
		CIFAR100-iid & 26.03 & 28.22 & 45.62 & 50.22 & \textbf{54.25} & 27.36 & 26.50 & 30.45 & 44.11 & \textbf{50.61} \\ \hline
		\multicolumn{1}{|l|}{\textbf{Setting 3}} & \multicolumn{5}{c|}{20 clients full participation} & \multicolumn{5}{c|}{20 clients partial participation} \\ \hline
		Synthetic(0,0) & 98.65 & 98.65 & 98.65 & 99.25 & \textbf{99.35} & 98.75 & 98.70 & 98.65 & 99.32 & \textbf{99.57} \\
		Synthetic(1,0) & 97.83 & 97.82 & 97.90 & 98.65 & \textbf{98.83} & 97.70 & 97.67 & 97.90 & 98.82 & \textbf{99.23} \\
		Synthetic(0,1) & 97.75 & 97.75 & 97.90 & 99.10 & \textbf{99.30} & 98.52 & 98.50 & 98.58 & 99.30 & \textbf{99.62} \\ \hline
	\end{tabular}
	\vskip -8pt
\end{table*}

\subsection{Results and Analysis}

We run vast experiments to determine the superiority of FedDC on the convergence speed and the model performance. Besides, we also demonstrate the robustness and superiority of FedDC in different participation levels, different client scale and different data heterogeneity.  
All results are reported based on the global model. As the baselines and FedDC consume the same computational resource in each round, so that we report the number of communication rounds instead of the FLOPS. 
The goal of FedDC mainly includes two perspectives: (1) speeding the model convergence rate to reduce the communication cost, and (2) improving the model performance trained on different datasets. Our results highlight the benefit of FedDC compared to the existing FL optimization approaches. 



\begin{figure*}[!t]
	\footnotesize
	\centering \vskip -2pt
	\centering 
	\begin{subfigure}{0.32\linewidth}
		\includegraphics[width=1\linewidth]{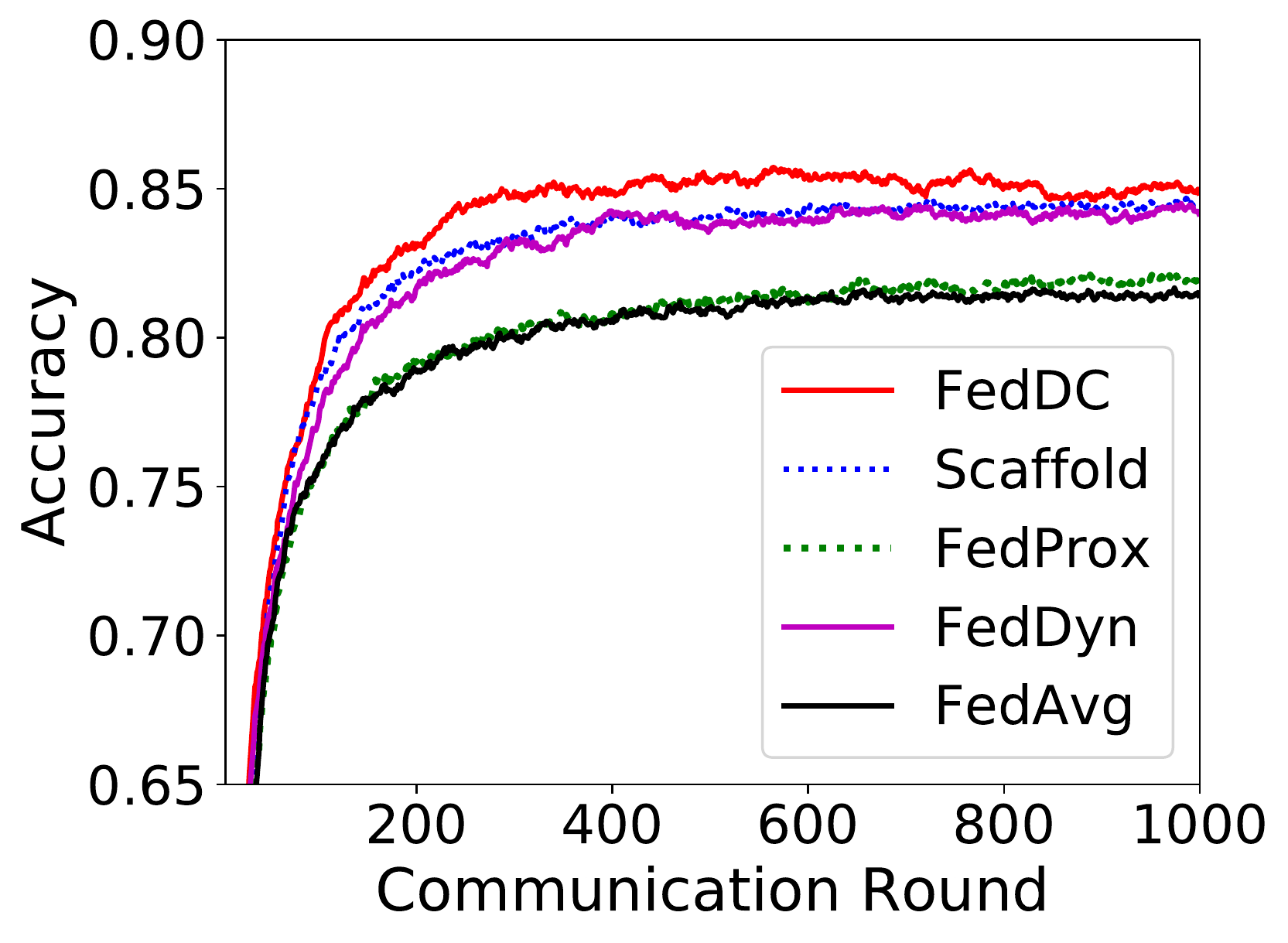}
		\caption{CIFAR10, iid, partial participation}\vskip -10pt
	\end{subfigure}
	\begin{subfigure}{0.32\linewidth}
		\includegraphics[width=1\linewidth]{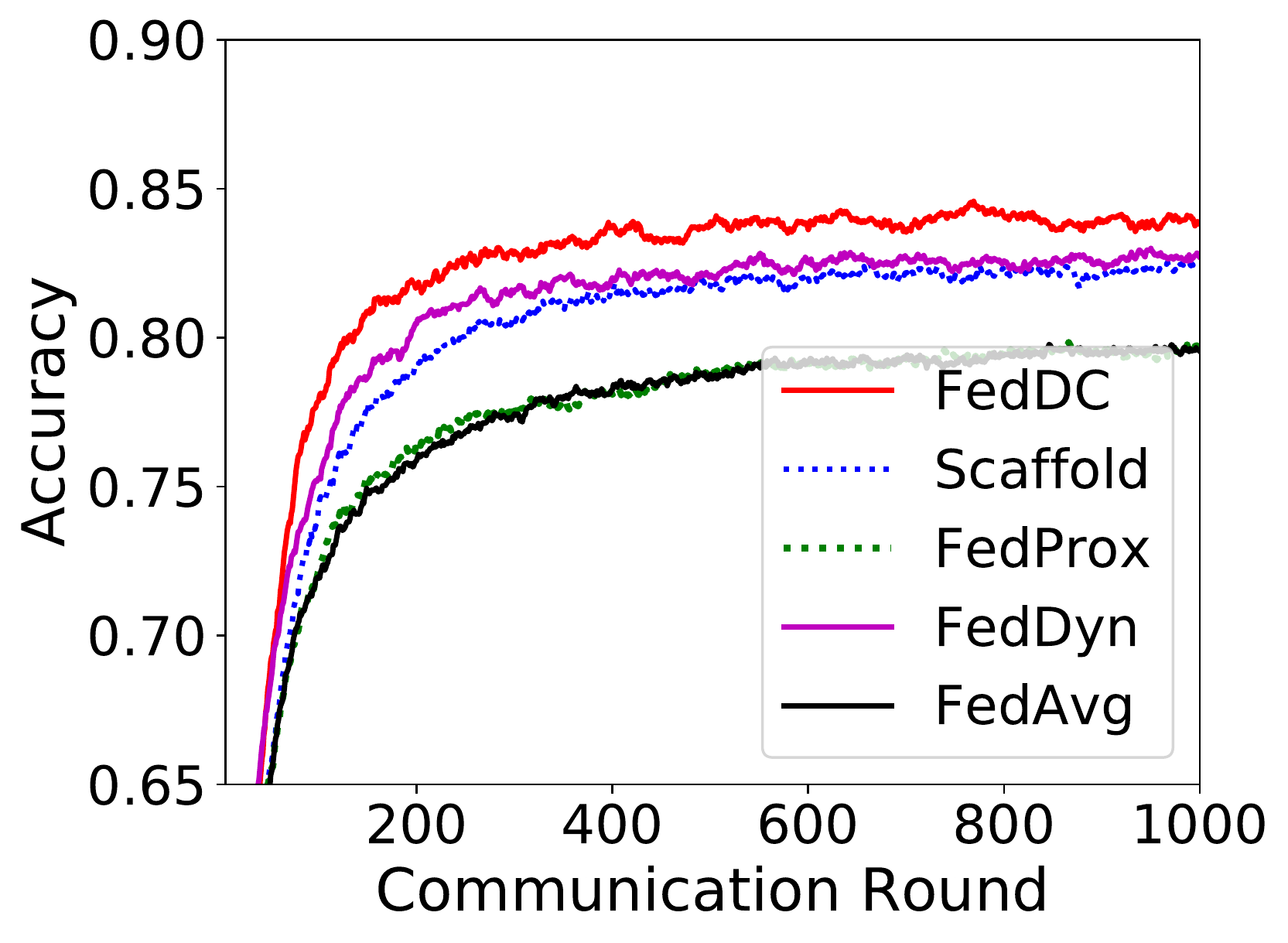}\caption{CIFAR10, D2, partial participation} \vskip -10pt
	\end{subfigure}
	\begin{subfigure}{0.32\linewidth}
		\includegraphics[width=1\linewidth]{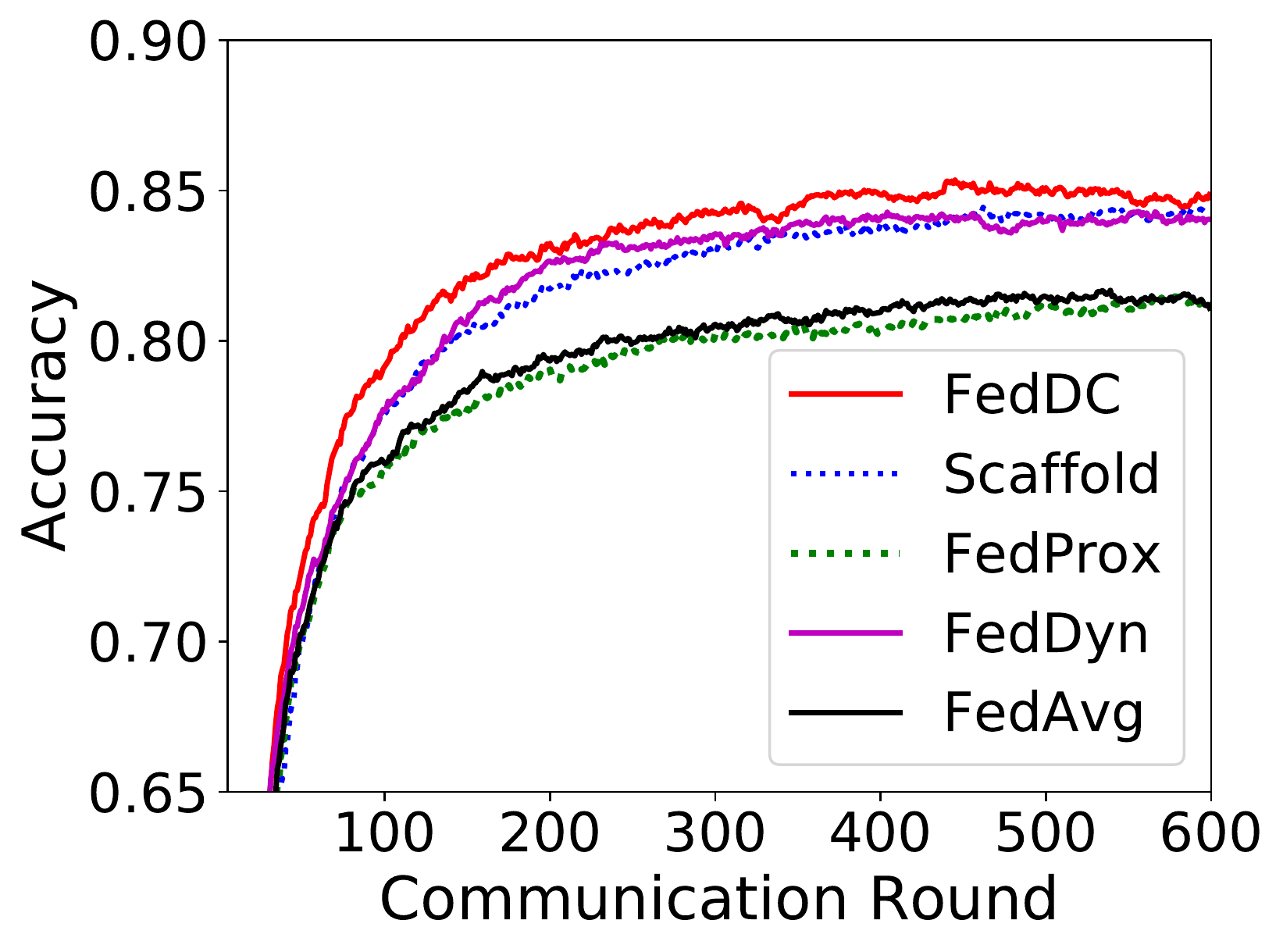}\caption{CIFAR10, unbalance, partial participation} \vskip -10pt
	\end{subfigure}
	\\
	\begin{subfigure}{0.32\linewidth}
		\includegraphics[width=1\linewidth]{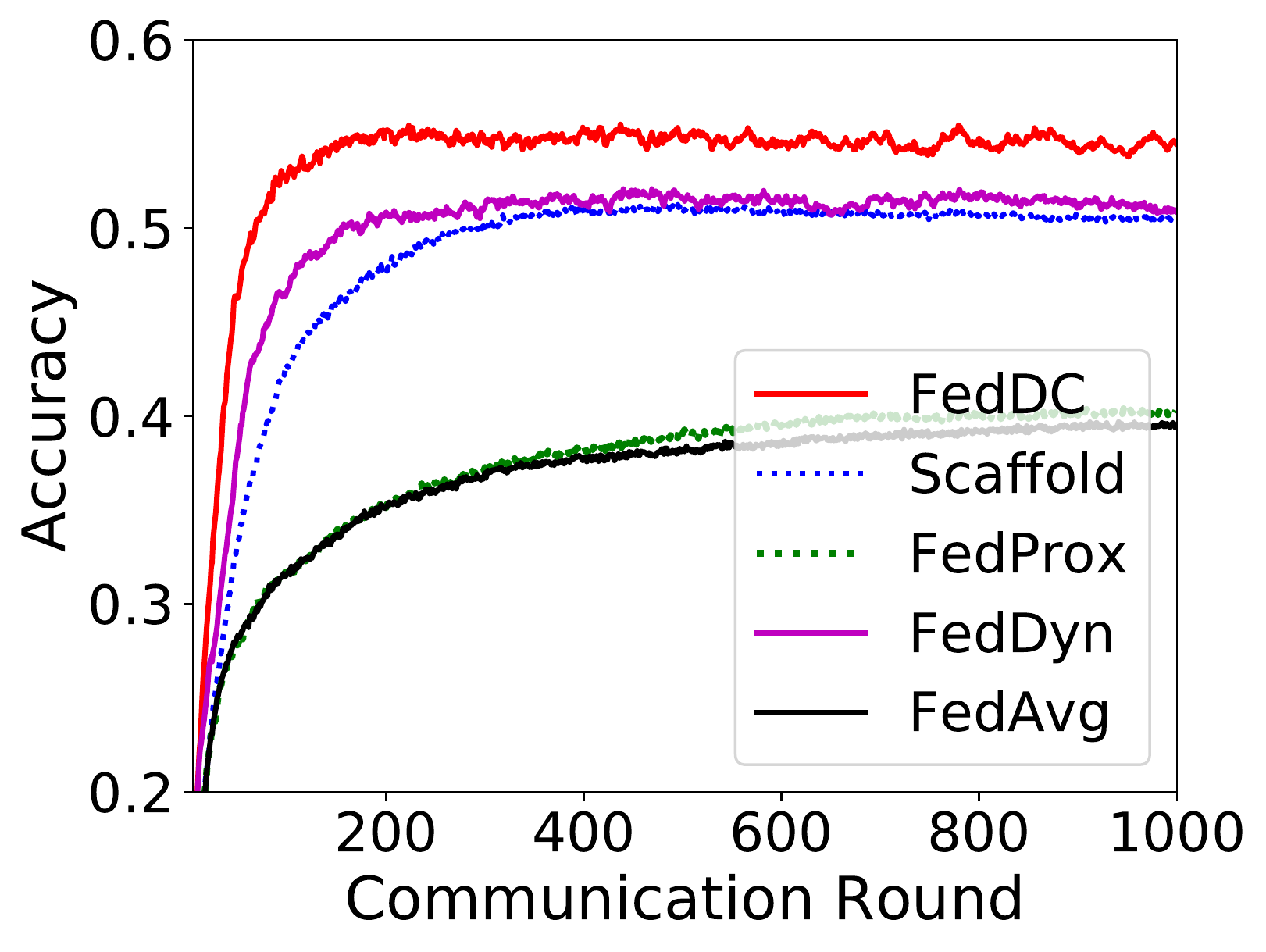}\caption{CIFAR100, iid, full participation}\vskip -10pt
	\end{subfigure}
	\begin{subfigure}{0.32\linewidth}
		\includegraphics[width=1\linewidth]{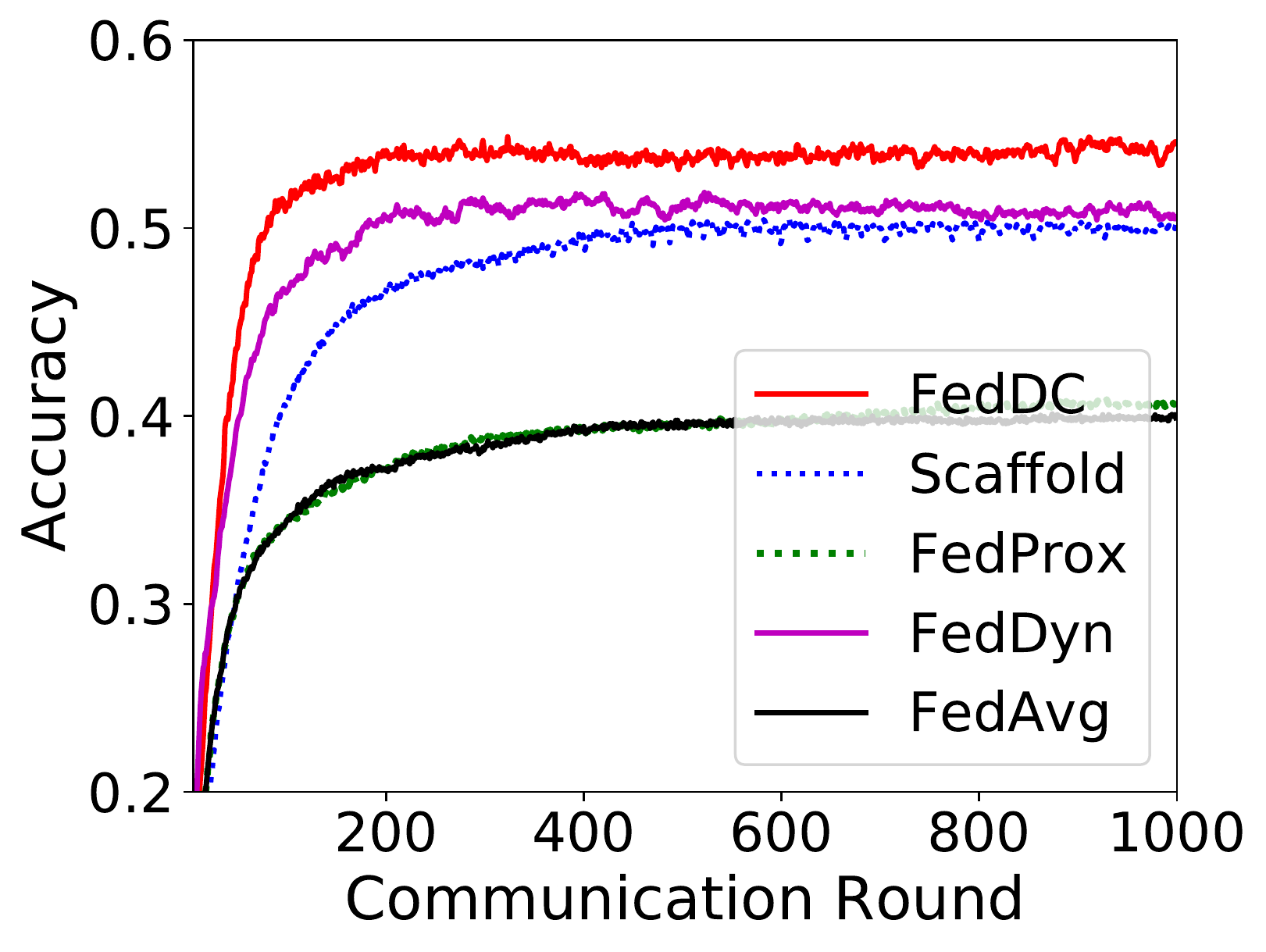}\caption{CIFAR100, D2, full participation}\vskip -10pt
	\end{subfigure}
	\begin{subfigure}{0.32\linewidth}
		\includegraphics[width=1\linewidth]{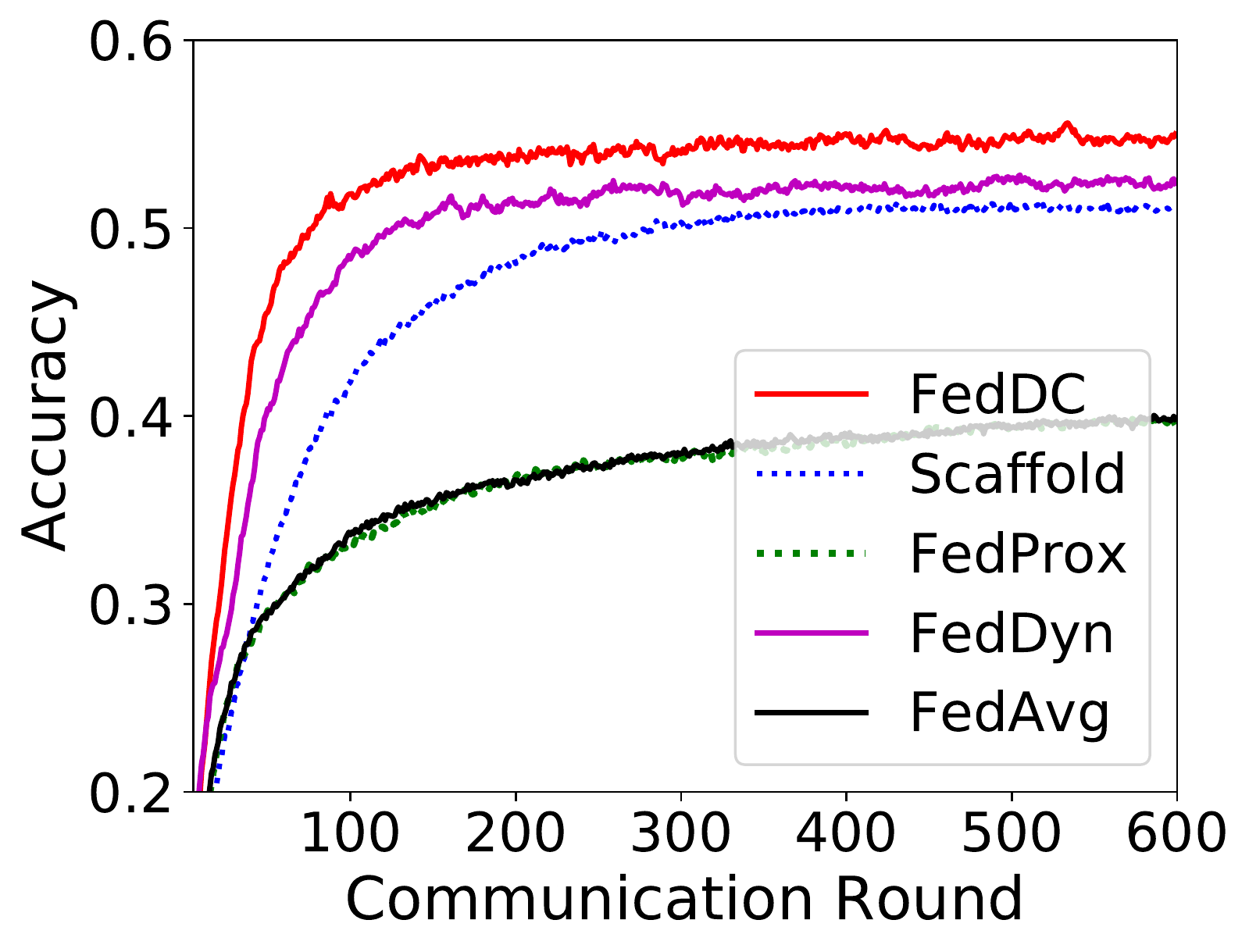}\caption{CIFAR100, unbalance, full participation}\vskip -10pt
	\end{subfigure}
	\vskip -5pt
	\caption{Convergence plots for FedDC and other baselines in different settings that with $100$ clients partial ($15\%$) client participating on iid, D2 non-iid (Dirichlet-0.3) and unbalanced data of CIFAR10 and CIFAR100 datasets. (a), (b) and (c) are training on CIFAR10 with partial participation. (d), (e) and (f) are training on CIFAR100 with full participation. }\label{figure_converge}
	\vskip -10pt
\end{figure*}

\textbf{Fast convergence of FedDC.}
Table \ref{tab_communication_saving_fashion} compares the convergence speed of FedDC and the mentioned baselines. The results show that FedDC is the best one to handle the local drift and speeds up the convergence speed compared with other methods. 
Specifically, FedDC could achieve a target accuracy using fewer communication rounds than the FedAvg, FedProx, Scaffold and FedDyn. For instance, in the iid setting, FedDC spends 37 communication rounds to achieve $40\%$ accuracy while 100 clients full participating in training on CIFAR100, while FedAvg spending over 1000 rounds to achieve $40\%$ accuracy in the same setting. That is, the convergence speed of FedDC relative to FedAvg is faster over $27.03 \times$.
We may attribute this to the fact that FedDC bridges the local drift and efficiently optimizes the objectives. The convergence speedup also leads to proportional communication-saving.
And Figure \ref{figure_converge} shows more vivid results of the convergence plots, in which FedDC is consistently the fastest one in all settings.
Figure \ref{figure_converge} (a, d) show the convergence plots in iid settings on CIFAR10 and CIFAR100. Figure \ref{figure_converge} (b, e) are accuracy plots on non-iid settings. From these convergence plots, we intuitively observe that FedDC achieves better accuracy and greatly speeds up the convergence speed than baselines. It is obvious that convergence speedup of FedDC relative to baselines is larger on non-iid settings than in iid settings. 
As the increasing of data heterogeneity, the local models suffer from more significant client drift. FedDC handles the drift by bridging the gap using the local drift variables that are learned on the client-side, so that FedDC show an obvious advantage in convergence speed over other baselines. The results confirm that FedDC has a stronger ability to handle heterogeneous data.  
Figure \ref{figure_converge} (c, f) are convergence plots on unbalanced data set settings. 
The unbalanced data introduces another type of system heterogeneity, making the convergence speed slower than in the balanced data. The results show FedDC's superiority in both model performance and convergence speed in unbalanced settings, we find that FedDC also has the potential to handle the heterogeneity caused by unbalanced data.
In addition, a widespread trend in these figures is that as the target accuracy improves, the communication-saving of FedDC relative to other methods become bigger. Another trend is that the improvement of FedDC over baselines in CIFAR100 is bigger than in CIFAR10 in the same settings. We attribute it to the fact that as the difficulty of optimization increases, FedDC's robustness advantage over other methods is further highlighted. FedDC can utilize the local drift variable to capture the system heterogeneity in clients' local datasets and capture the subtle features needed to classify confusing samples. 

\textbf{Better performance of FedDC.}
Table \ref{tab-best-accuracy} compares the best accuracy of FedDC with baselines on evaluation datasets with various settings. On CIFAR10 and CIFAR100, FedDC always achieves the best test accuracy, where FedAVG and FedProx have the least. For instance, when training on the data of 0.3-Dirichlet distribution (Dw) CIFAR10 with $100$ clients full participating, the test accuracy of FedDC is $84.32\%$, the accuracy of FedAvg achieves $79.14\%$ and the accuracy of Scaffolf achieves $82.96\%$. 
FedDC also achieves appreciable improvement in top-1 test accuracy on the unbalanced settings. 
Besides, the results in setting 2 (500 clients) and setting 3 (20 clients) indicate that FedDC is efficient in the practically relevant massively distributed settings.
The improvements of FedDC indicates that tracking and correcting client drift effectively prevent the model performance from decreasing. Compared with Scaffold, FedDC not only uses the gradient correction term to reduce gradient drift but also introduces the local drift variable to track the deviation between the global model and local models, so that FedDC is the best one to prevent the accuracy reduction.
Table \ref{figure_tiny} shows the accuracy of ResNet18 that training for 15 rounds on Tiny ImageNet, where the ResNet18 is started from an ImageNet pre-trained model. The performance of FedDC significantly outperforms the baselines in all settings. This shows that FedDC is still efficient in tasks that use pre-trained models.

\textbf{Robustness on heterogeneous data.}
A more extensive non-iid data or unbalanced data can greatly slow down the model convergence \cite{pmlr-v54-mcmahan17a}. Comparing the convergence plots of Figure \ref{figure_converge} (a,b,d,e), the results show that the data distribution has a prominent influence on both the model convergence speed and accuracy. It reveals that the convergence speed on the iid data is faster than on the non-iid data in which the local dataset can not well approximate the overall distribution. 
As shown in Table \ref{tab-best-accuracy}, FedDC outperforms baselines on iid, non-iid and unbalanced settings. 
FedDC gets more communication-saving gains relative to other methods when we increase the target accuracy or training on a harder task. 
The data heterogeneity does damage to the model performance of all methods. While training with $100$ clients and full participation on CIFAR100, the accuracy of FedDC is $85.71\%$ in iid setting, $84.77\%$ in 0.6-Dirichlet (D1) distribution, and $84.58\%$ in 0.3-Dirichlet distribution (D2, it is more non-iid than 0.6-Dirichlet). 
However, even in these heterogeneous data settings, FedDC maintains its competitive advantage compared with baselines because it is able to neutralize the local drifts.

\textbf{Robustness to massive clients.}
We conduct experiments to analyze the effectiveness of FedDC while adopting different amounts of clients to participate in the training process. We report the model accuracy in Table \ref{tab-best-accuracy} with $100$ and $500$ clients of both partial participation and full participation on CIFAR10 and CIFAR100 datasets.
FedDC achieved the best performance consistently. 
FedDC converges to a better stationary point than other methods. In the setting with $100$ clients (setting 1) and full client participating, the test accuracy of FedDC is $84.93\%$ on CIFAR10, while FedAVG only achieves $74.43\%$ ($11.5\%$ lower than FedDC) on CIFAR10. Scaffold and FedDyn methods always get intermediate accuracy. Moreover, the performance gap between FedDC and other methods increases when the client size increases from $100$ to $500$. We attribute it to that a smaller number of samples per device (with massive clients) brings a greater risk of optimization dispersion.

\textbf{Robustness to client sampling.}
The devices in FL are heterogeneous and flexible, which may join and exit at any time. 
To show that FedDC is resilient for clients sampling, we set the experiments with full participation and partial sampling participation (in this setting we randomly sample $15\%$ client join training each round). We compared the final performance of FedDC and the baseline algorithms in Table \ref{tab-best-accuracy}. Partial client participating means the active data is only a subset of all training data, which leads to unstable and slower convergence. In full clients participating, the accuracy of FedDC with 100 clients on iid CIFAR10 is $86.18\%$, and in $15\%$ client sampling, the accuracy decreases to $85.71\%$.
Moreover, the results turn out that keeping all clients active is not necessary for FedDC, where the partial client participating could achieve similar accuracy as the full client participating. FedDC keeps the best accuracy in partial client participation compared to the other methods. 
Thus, FedDC is much resilient to client sampling compared to baselines as it utilizes the clients' parameter deviations to improve the performance of the global model. The clients in FedDC hold and update drift variables locally, so that occasionally interrupted training does not cause the loss of the drift state, which allows clients to train better in partial client participation settings. 


    \section{Conclusion}
%

In this work, we proposed a novel FL algorithm with local drift decoupling and correction, named FedDC, to solve the problem of local drift which caused by the heterogeneous data. 
FedDC dynamically bridges the gap between the local model and the global model with the learned local drift variable.
Through extensive experiments on various image classification datasets, we demonstrated that our FedDC provides better performance and faster model convergence in FL. 
Moreover, FedDC is robust and efficient in homogeneous or heterogeneous data, in both full client participation and partial client participation.  
    
\noindent\textbf{Acknowledgement  }
The work is supported by the National Natural Science Foundation (NSF) under grant 62072306 and 61872372, Open Fund of Science and Technology on Parallel and Distributed Processing Laboratory under grant 6142110200407, and A*STAR AI3 HTPO Seed Fund (C211118012).

{\small
\bibliographystyle{ieee_fullname}
\bibliography{./tex_file/flnips}
}

\onecolumn 
\appendix
\section{Appendix: More Experiment Results}\label{appendix_results}
We run experiments on the true world datasets of image classification tasks including CIFAR10, CIFAR100, MNIST, fashion MNIST, Tiny ImageNet and EMNIST-L datasets. We also evaluate on a Synthetic dataset. We explored multiple types of models: the FCN for MNIST and EMNIST-L, the CNN architecture network for CIFAR10 and CIFAR100, and a Multi-class logistic network for  Synthetic  dataset, ResNet18 for Tiny ImageNet. We conduct comprehensive investigations for the impact of client heterogeneity by designing iid and non-iid data scenarios, balance and unbalance data, full client participating and part client participating scenarios. We also justify the robustness of FedDC in flexible devices and large-scale setup by assigning the data to different amounts of clients. For comparison, we utilize the FedAvg, FedProx, Scaffold and FedDyn algorithms as baselines. We detailed describe the experiment settings, the models and datasets, the comparison methods in the following.  

\subsection{Synthetic Dataset}
We conduct experiments on a Synthetic dataset which adopts the same setting as \cite{acar2021federated}. We generate the samples for each clients $(x,y)\in{D_i}$, where the samples and labels follow the rule of $y=argmax(\theta_i x+b_i)$, the shape of $x$ is $30\times1$, $y$ contains $5$ categories, $\theta_i$ (the shape is $5\times30$) and $b_i$ (the shape is $5\times1$) are the best parameter to fitting the data distribution in $i$-th client. We use $\gamma_1$ to control the value of $(\theta_i,b_i)$ which sampled from $N(\mu_i,1)$ where $\mu_i\sim N(0,\gamma_1)$, and $\gamma_2$ to control the data distribution in each client.    
In the experiments on Synthetic dataset, we set only one of $\gamma_1,\gamma_2$ as $1$ to allow one type heterogeneity for one set of experiments, and we set all of them as $0$ to simulate the homogeneous settings. Thus, the settings for $(\gamma_1,\gamma_2)$ include $(0,0)$, $(0,1)$, $(1,0)$, that represent a homogeneous setting and two heterogeneous settings. In all experiments on Synthetic dataset, the amount of clients is $20$, the average amount of samples for each client is $200$.

\subsection{Real World Dataset}
\textbf{Datasets and models.}
We adopt the real-world datasets for the image classification task, including MNIST, EMNIST-L, CIFAR10, fashion MNIST, Tiny ImageNet and CIFAR100 datasets. The EMNIST-L is used for characters classification, which is a subset of the EMNIST dataset that only contains the first $10$ categories. The MNIST is the dataset for the classification of handwritten digits, which contains $10$ categories. FashionMNIST is an image dataset that replaces the MNIST handwritten digit set. Different from the MNIST handwriting data set, the Fashion-MNIST data set contains 10 categories of images, namely: t-shirt (T-shirt), trouser (jeans), pullover (pullover), dress (skirt), coat (coat) , Sandal (sandals), shirt (shirt), sneaker (sports shoes), bag (bag), ankle boot (short boots). The sample size of the EMNIST-L, fashion MNIST and MNIST is $(1\times28\times28)$. For MNIST and fashion MNIST, the sample amount in the training set is $60000$, and the sample number in the test set is $10000$. For EMNIST-L, the sample amount in the training set is $48000$, and the sample amount in the test set is $8000$. 
Both the CIFAR10 and CIFAR100 datasets contain $60000$ of $3\times32\times32$ images. For CIFAR10, there are $10$ categories, and there are $100$ categories on CIFAR100. For both the CIFAR10 and CIFAR100, the sample amount in the training set is $50000$, and the sample amount in the test set is $10000$. 

A fully-connected network (FCN) as \cite{pmlr-v54-mcmahan17a} is adopted for the classification of MNIST and EMNIST-L. The FCN includes an input layer, two fully connected hidden layers and an output layer. The two hidden layers both contain 200 neurons. A network with CNN-based structure is employed to classify samples on CIFAR10 and CIFAR100. The CNN follows similar setting as \cite{pmlr-v54-mcmahan17a}, which consists of the basic modules of CNN, including two conventional layers with 64 of $5\times5$ convolution kernels, each conventional layer followed a down—pooling larger, after that are two fully connected layers with $394$ and $192$ neurons and a softmax layer for prediction. The classic ResNet18 network is adopted in the Tiny ImageNet dataset.

\textbf{iid and non-iid setting.}
The experiments mainly contain three types of balanced data settings, including an iid setting and two non-iid settings. For the iid data distribution, all clients get the same number of samples that are independently identically distributed on the training dataset. For the non-iid settings, we obey the Dirichlet distribution to sample data. In non-iid settings, the label ratio of each client follows the Dirichlet distribution. For each client, its samples are sampled without replacement from the full training dataset according to the label ratio that obeys the Dirichlet distribution. A hyper-parameter of the Dirichlet distribution controls the data heterogeneity degree, and we set two types of Dirichlet distributions where the hyper-parameter is $0.3$ and $0.6$, respectively. Dirichlet-0.3 distribution is stronger non-iid than Dirichlet -0.6 distribution . In most experiments, we set $100$ clients in the experiments. Each of them contains $1\%$ samples of full training data in the balanced settings.

\textbf{Unbalanced setting.} In unbalanced data settings, the sample amount of clients are different from each other. To produce the unbalanced dataset, each client owns data points in which the amount follow a lognormal distribution. The hyper-parameter in the lognormal distribution is the variance of the distribution. In the balanced setting, the variance is $0$, and we set the variance as $0.3$ in the unbalanced setting.

\textbf{Hyper-parameter setting.}
We give the hyper-parameter settings in different datasets. For all the true world datasets including MNIST, EMNIST-L, CIFAR10 and CIFAR100, we set the batch size as $50$, the number of local epochs in one communication round as $5$, the initial learning rate as $0.1$ and the learning rate decay per round as $0.998$, the weight decay as $0.001$. We search the $\alpha$ of FedDC in $[0.001,0.005,0.01,0.05,0.1,0.2,0.5,1]$.
In experiments of CIFAR10 and CIFAR100, when the number of client is $100$ we set $\alpha=0.01$ for FedDC, and when the number of clients is $500$ we set $\alpha=0.05$ for FedDC. In experiments with $100$ clients, we set the hyper-parameter $\alpha=0.1$ for MNIST, fashion MNIST and $\alpha=0.2$ for EMNIST-L in FedDC, respectively. In experiments of MNIST with $500$ clients, we set the hyper-parameter $\alpha=0.2$ for FedDC. In experiments on Tiny ImageNet dataset, we set the client number as $20$, besides, the pretrained ResNet18 is adopted.
For the Synthetic dataset, we use a multi-class logistic classification model to classify samples, and we set the batch size as $10$, the epoch amount in each local communication round as $10$, the learning rate as $0.1$, the hyper-parameter in FedDC as $\alpha=0.005$.
For the hyper-parameters in baselines, $\alpha= 0.01$ in FedDyn, $\mu=0.0001$ and weight decay as $1e-5$ in FedProx.

We adopt the same hyper-parameter for a specific dataset for all iid or non-iid data, full client participating or $15\%$ client participating settings. In the experiments, we use "MNIST iid 100-clients-100\%" to represent the result on iid distributed MNIST dataset with $100$ clients and $100\%$ client participating, and so on.

\subsection{Results}
\begin{figure*}[!t]
	\centering 
	
	\begin{subfigure}{0.4\linewidth}
		\includegraphics[width=1.0\linewidth]{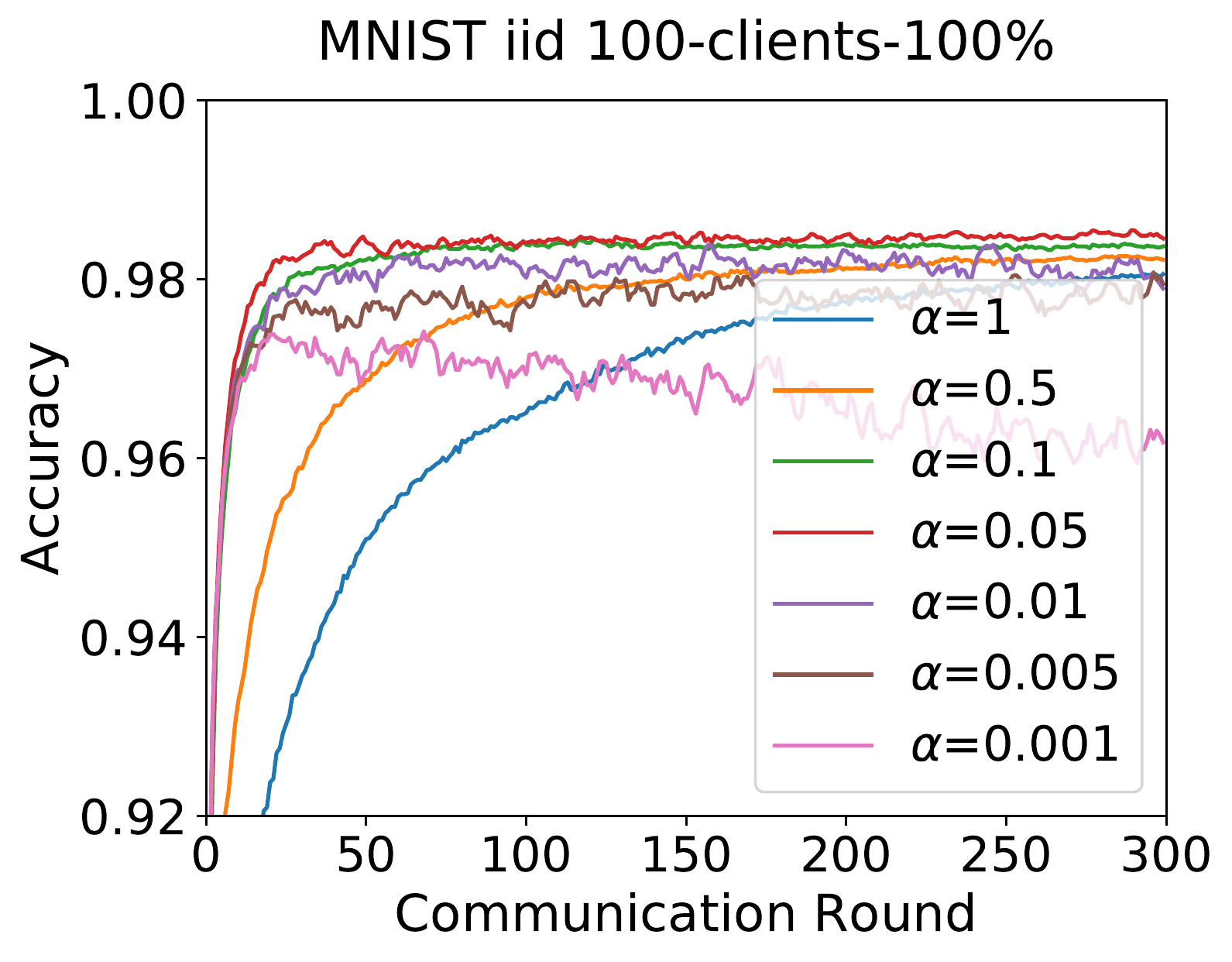}	\caption{}
	\end{subfigure}
	\begin{subfigure}{0.4\linewidth}
		\includegraphics[width=1.0\linewidth]{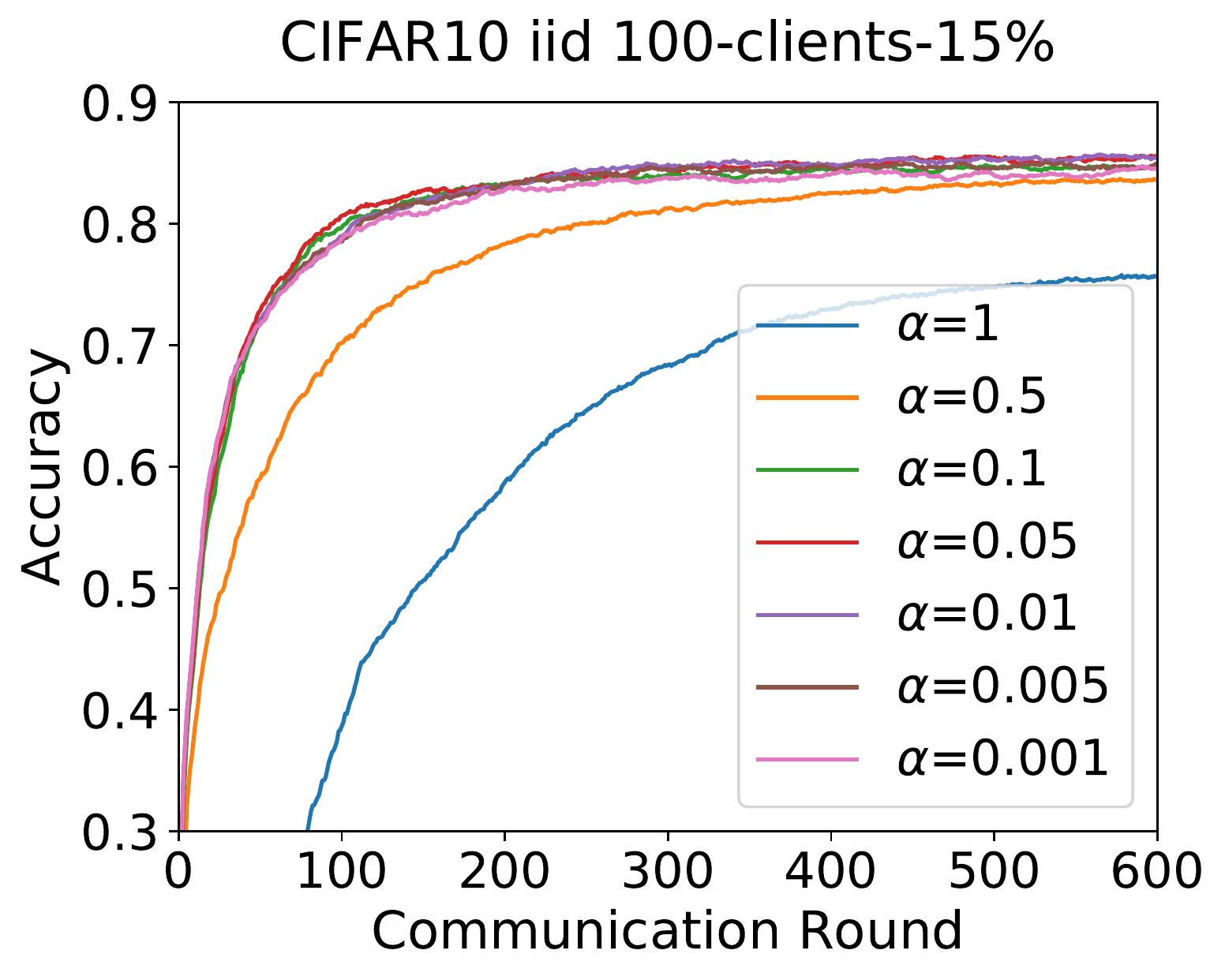}	\caption{}
	\end{subfigure}\\
	\begin{subfigure}{0.4\linewidth}
		\includegraphics[width=1.0\linewidth]{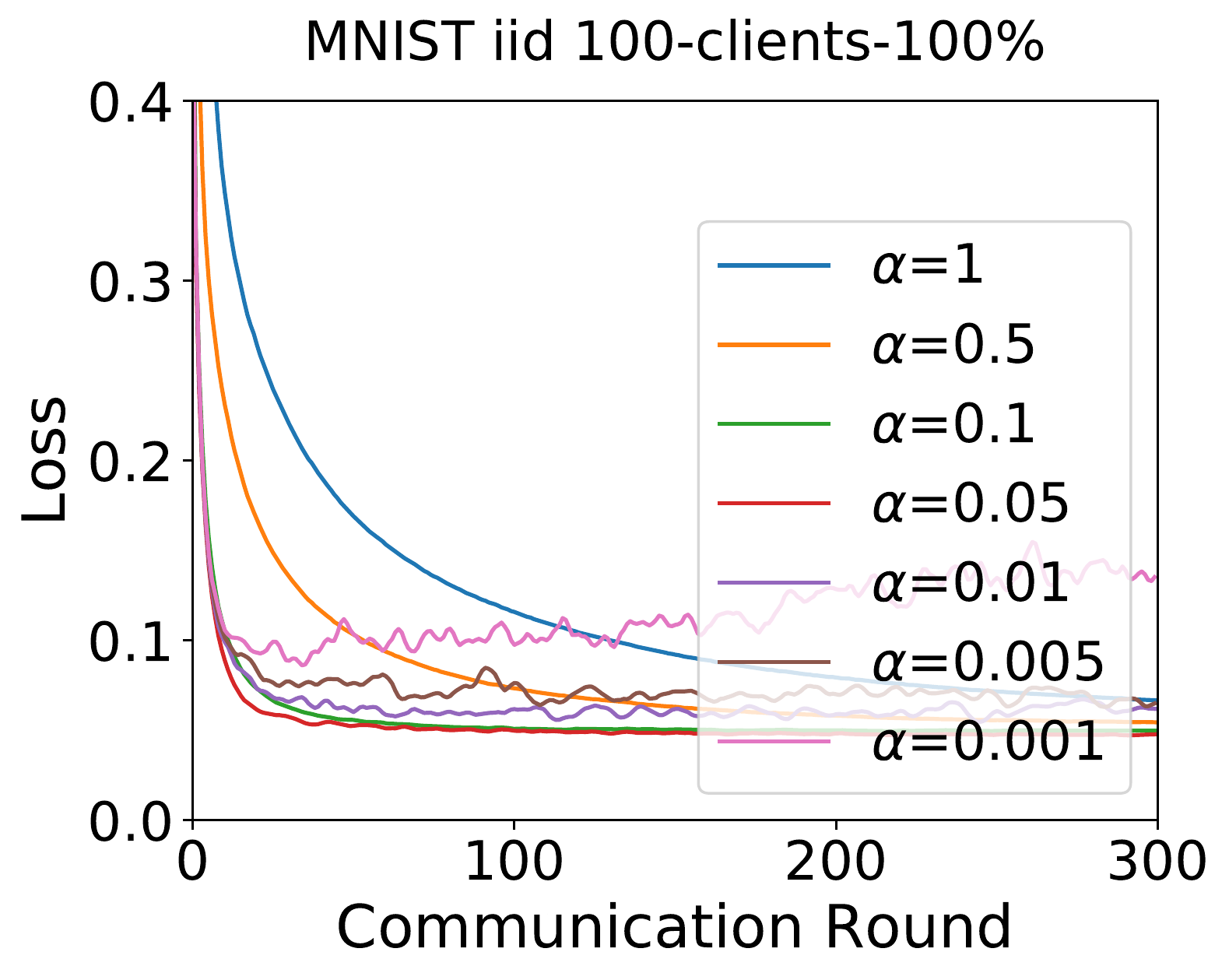}	\caption{}
	\end{subfigure}
	\begin{subfigure}{0.4\linewidth}
		\includegraphics[width=1.0\linewidth]{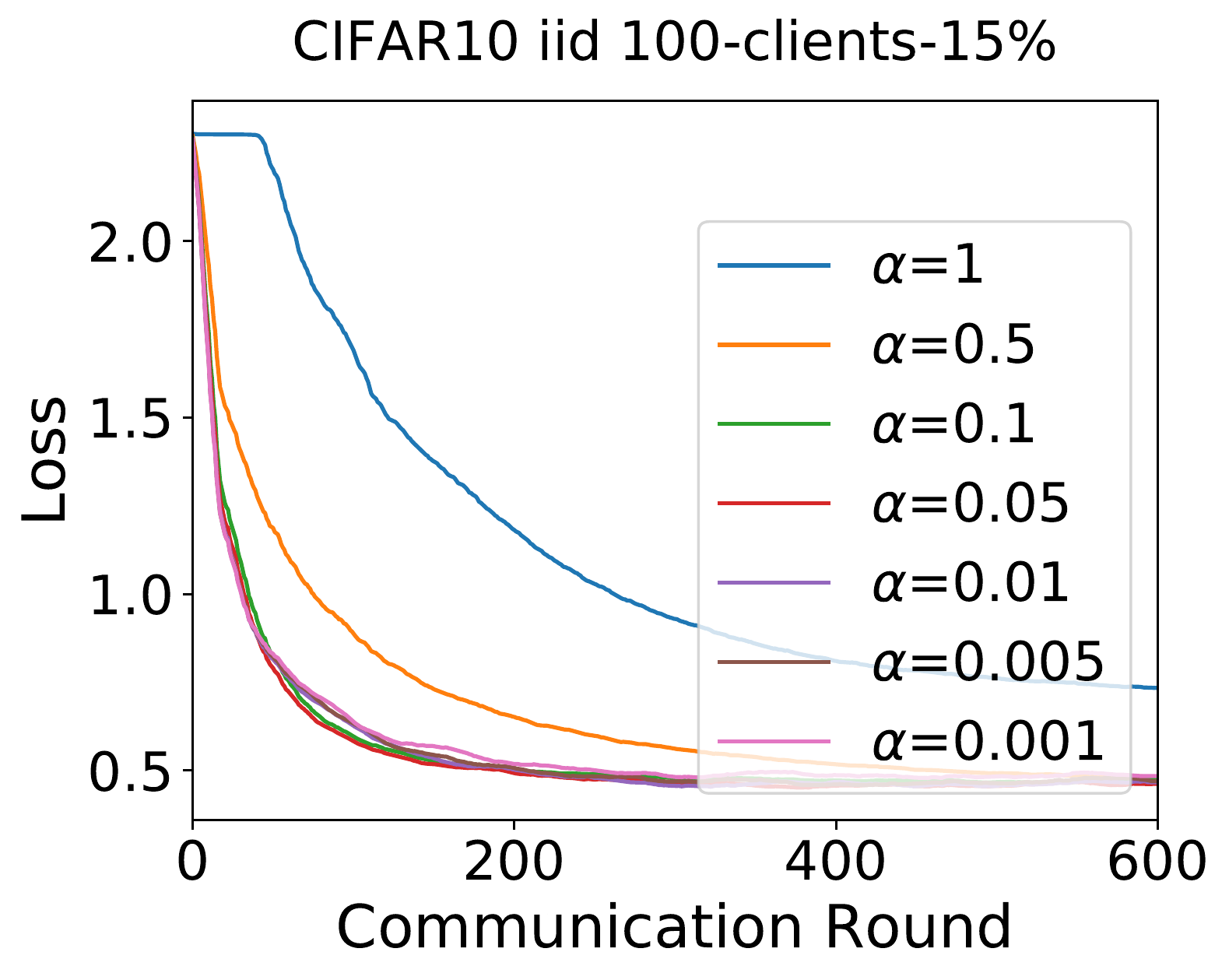}	\caption{}
	\end{subfigure}
	\caption{Convergence plots of FedDC for different hyper-parameter-$\alpha$ settings with 100 clients adopting $100\%$ and $15\%$ client participating settings on iid MNIST and CIFAR10.}\label{figure_alpha}
\end{figure*}

\textbf{Sensitive of hyper-parameter in FedDC.}
In FedDC, there is only one manually controlled parameter $\alpha$. It controls the weight of the penalized term in the local objective function. The $\alpha$ is related to the dissimilarity between local parameters and the global parameter.
To analyze the impact of $\alpha$, we run experiments with different $\alpha$. The hyper-parameter $\alpha$ is explored in $[0.001,0.005,0.01,0.05,0.1,0.5,1]$. The convergence plots of experiments use different $\alpha$ with 100 clients $15\%$ client participating on MNIST and CIFAR10 datasets are shown in Figure \ref{figure_alpha}. 
A large value of $\alpha$ increases the weight of the penalized term and that leads to less attention to the local experience loss, which will cause the model difficult to converge. As the figure shows, $\alpha=1$ and $\alpha=0.5$ get the worst performance. The value of $\alpha$ needs to trade off the empirical loss term and the penalized term, a more reasonable value of $\alpha$ for CIFAR10 is in the range $[0.1,0.01]$. All experiments of different $\alpha$ converge to a stationary point, but a carefully selected $\alpha$ better improve the model performance. 

\begin{figure}[!t]
	\centering

	\begin{subfigure}{0.4\linewidth}
		\includegraphics[width=1.0\linewidth]{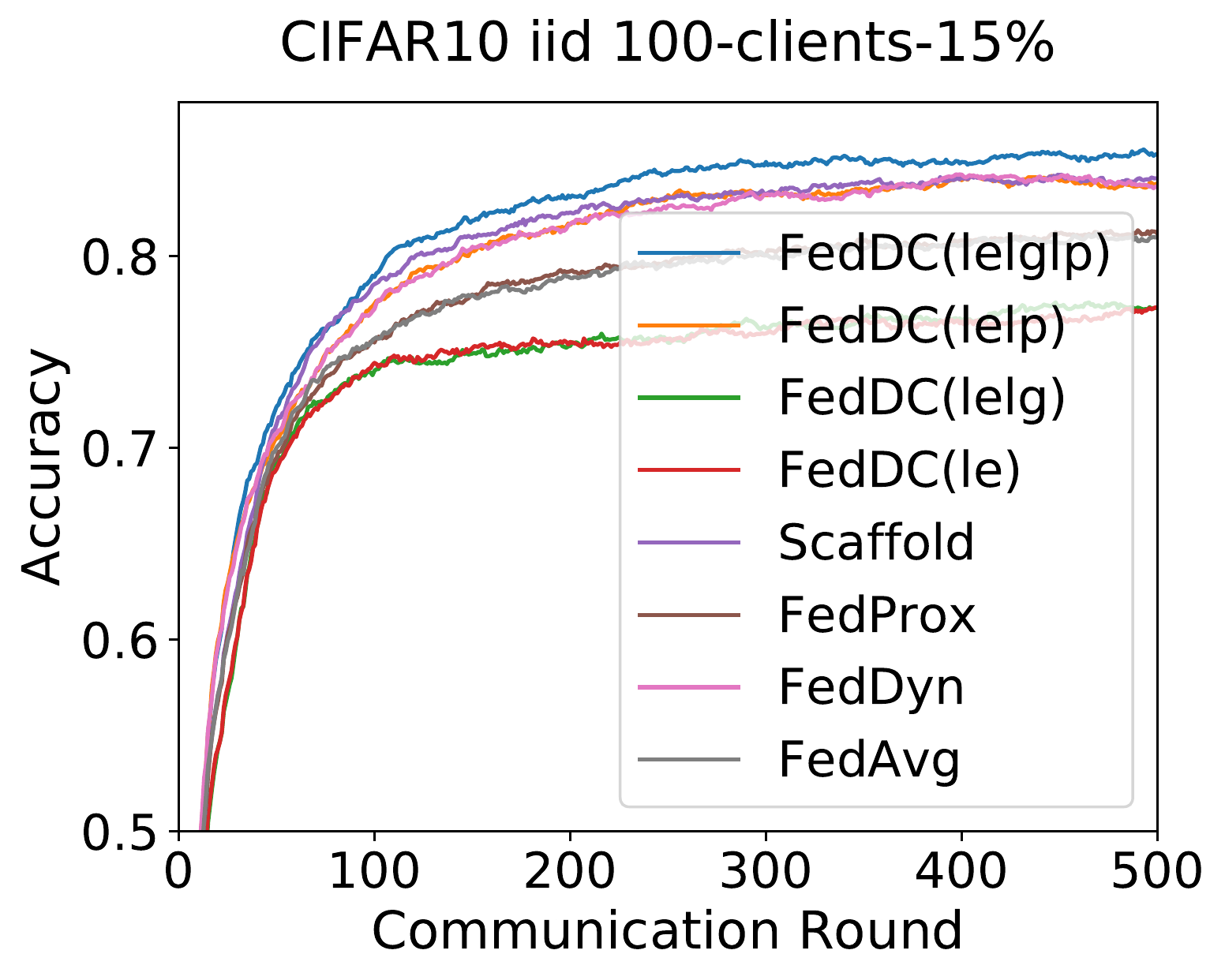}	\caption{}
	\end{subfigure}
	\begin{subfigure}{0.4\linewidth}
		\includegraphics[width=1.0\linewidth]{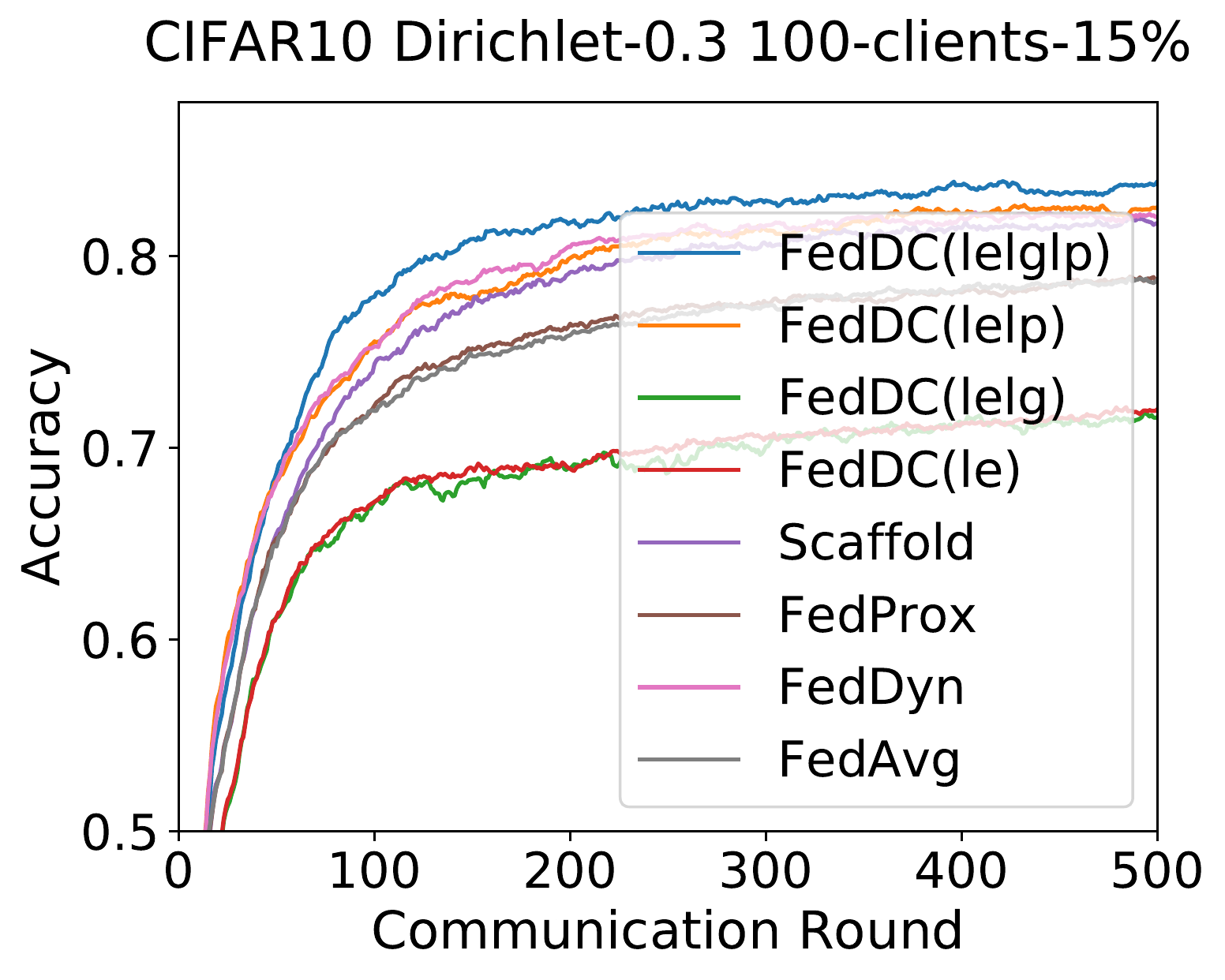}	\caption{}
	\end{subfigure}
	\caption{Ablation study for FedDC on CIFAR10. FedDC(le) adopts the proposed training process and uses the empirical loss as local objection function. FedDC(lelp) adopts the proposed training process and uses the sum of the empirical loss and the penalized term as local objection function. FedDC(lelg) adopts the proposed training process and uses the sum of the empirical loss and the gradient correction term as local objection function. FedDC(lelglp) adopts the proposed training process and uses the sum of the empirical loss, the gradient correction term and the penalized term as local objection function.  }\label{figure_ablation}
\end{figure}

\textbf{Ablation study}
In order to show the effects of the gradient correction term and the penalized term in FedDC, respectively, we conduct the ablation study.
We denote the standard local empirical loss as $le$, the gradient correction term as $lg$, and the penalized term as $lp$.
The training process of FedDC is different with FedAvg because we use the drift variable to decouple the local models and the global model, in which the global parameters are the sum of local parameters and drift variables. We compare the results of using the process of FedDC with different combinations of local objective functions. $FedDC(le)$ represents the method that adopts the training process of FedDC and uses the standard local empirical loss as the local objective function. With FedDC's training process, $FedDC(lelg)$ and $FedDC(lelp)$ represent adding a gradient correction term and a penalized term to the local objective function, respectively. $FedDC(lelglp)$ is the proposed method in which the local objective function includes the empirical loss, the penalized term and the gradient correction term. Figure \ref{figure_ablation} shows the convergence plots of the ablation study on iid distributed and non-iid distributed CIFAR10 dataset with $100$ clients $15\%$ client participating. From the figure, we can observe that the performance of the $FedDC(le)$ and $FedDC(lelg)$ methods are the worst, which even significantly lower than the standard FedAvg. The local objective functions of both $FedDC(le)$ and $FedDC(lelg)$ do not contain the penalized term to limit the drift variable, so that the global parameters can not be treated as the sum of the local parameters and the drift variables. Thus, the drift variable does not take effect; instead, it leads to worse performance. Compared with FedAvg, $FedDC(lelp)$ method greatly improves the performance, which indicates that decoupling the local parameters and global parameters with effective drift variables improves the model performance. The $FedDC(lelplg)$ method achieves the best performance compared with all the ablation methods and baselines, which indicates that the gradient correction term can reduce the risk of randomness and make the tracking of parameter drift more accurate.

\begin{figure}[!t]
	\centering

	\begin{subfigure}{0.3\linewidth}
		\includegraphics[width=1.0\linewidth]{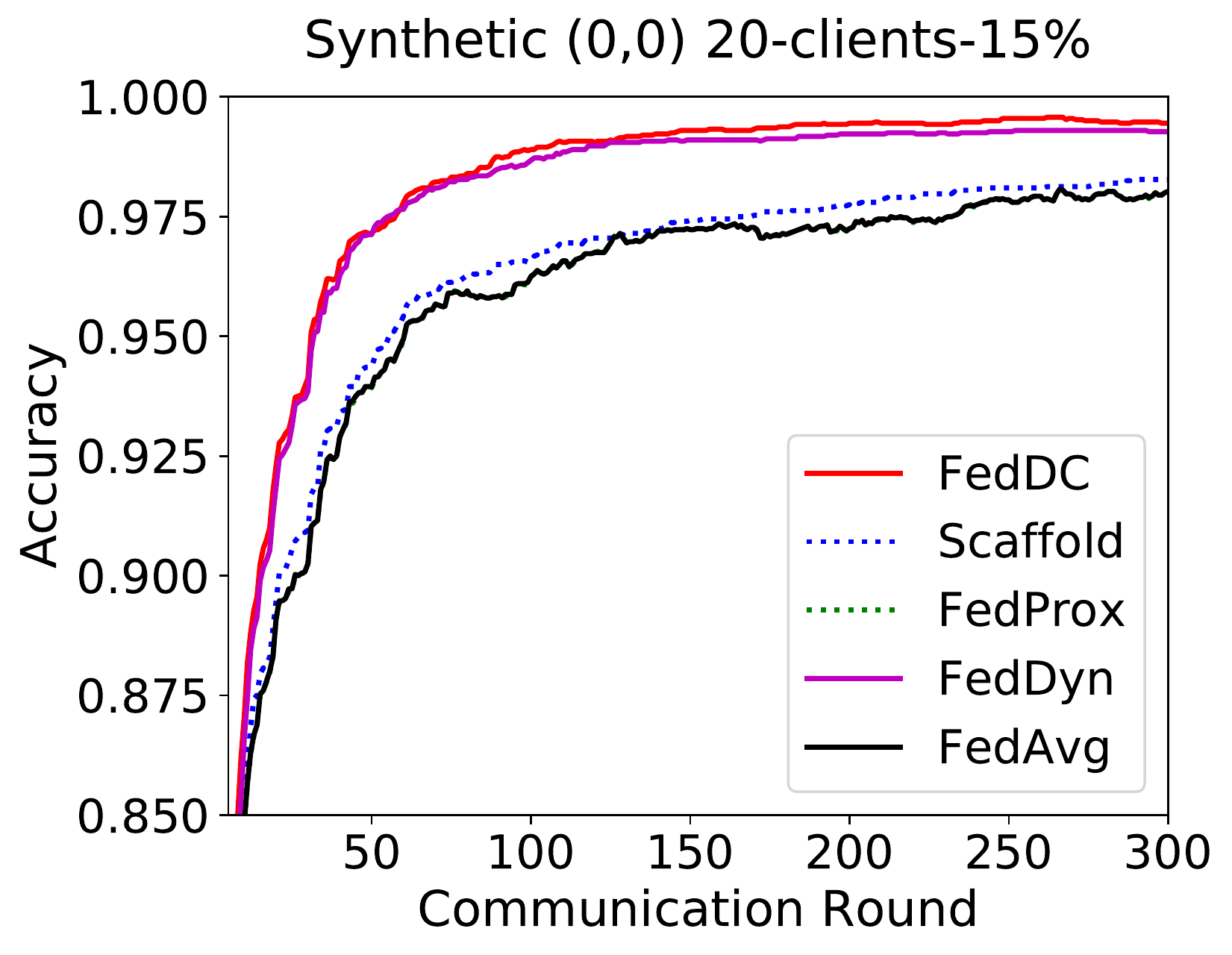}	\caption{}
	\end{subfigure}
	\begin{subfigure}{0.3\linewidth}
		\includegraphics[width=1.0\linewidth]{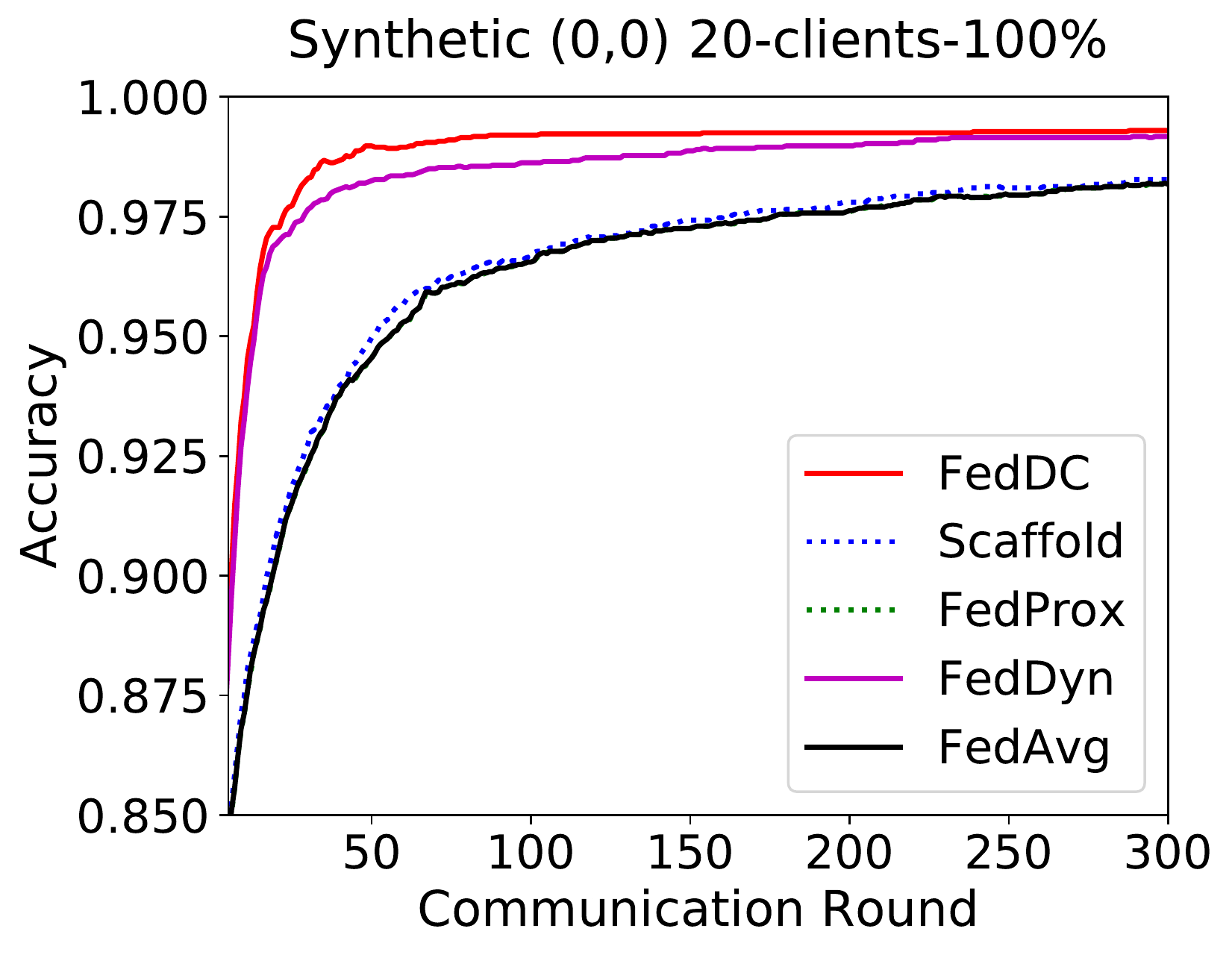}	\caption{}
	\end{subfigure}
	\begin{subfigure}{0.3\linewidth}
		\includegraphics[width=1.0\linewidth]{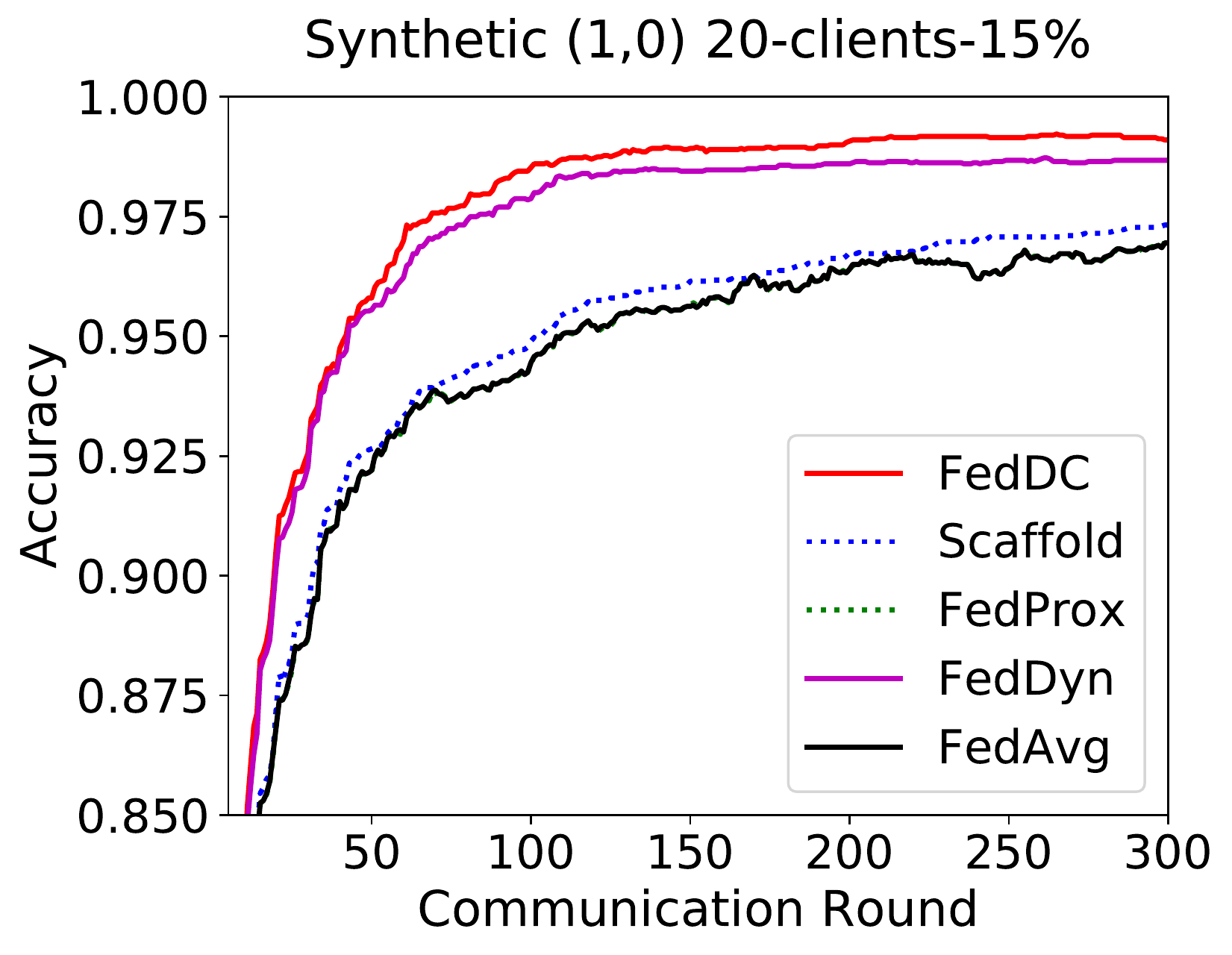}	\caption{}
	\end{subfigure}\\
	\begin{subfigure}{0.3\linewidth}
		\includegraphics[width=1.0\linewidth]{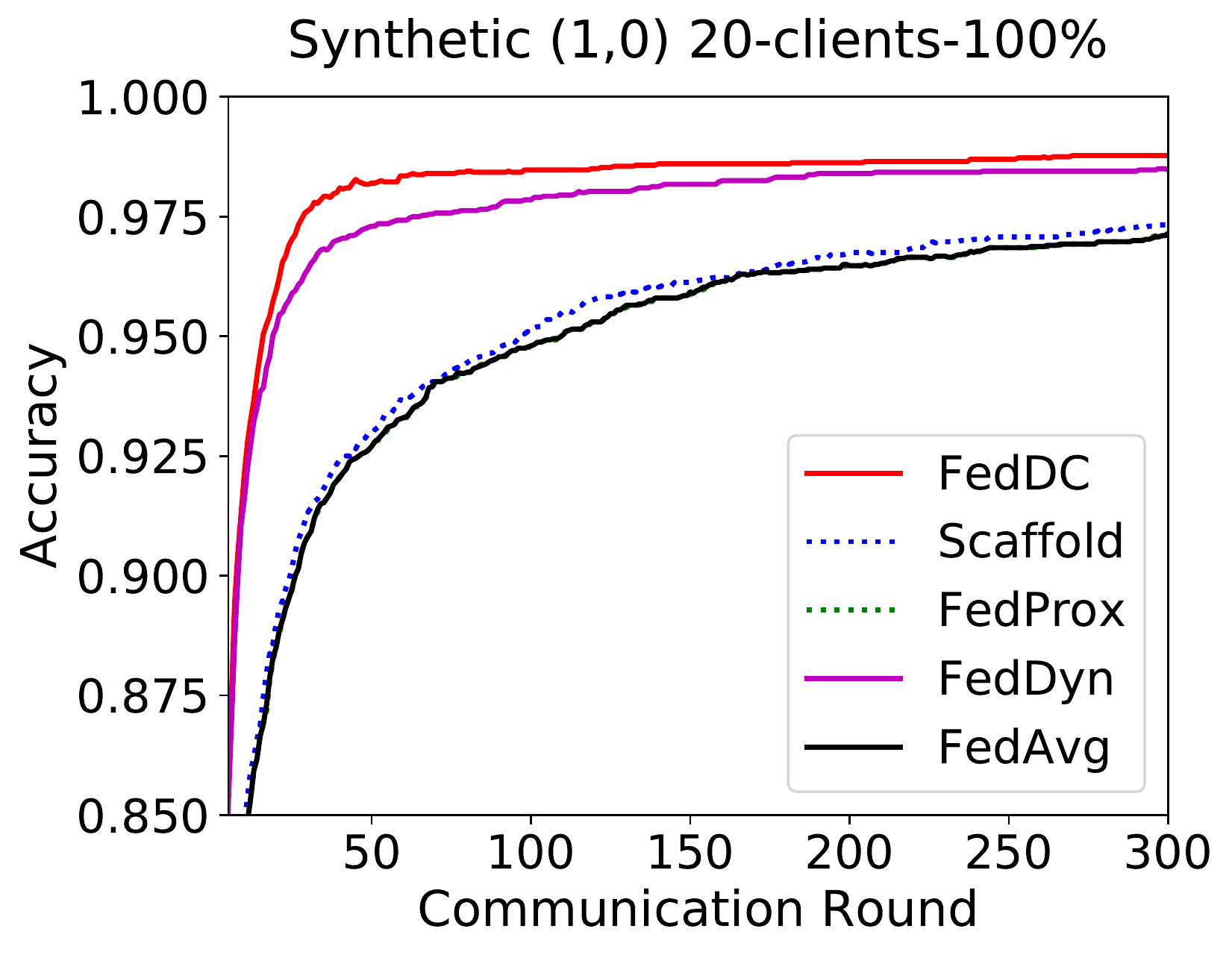}	\caption{}
	\end{subfigure}
	\begin{subfigure}{0.3\linewidth}
		\includegraphics[width=1.0\linewidth]{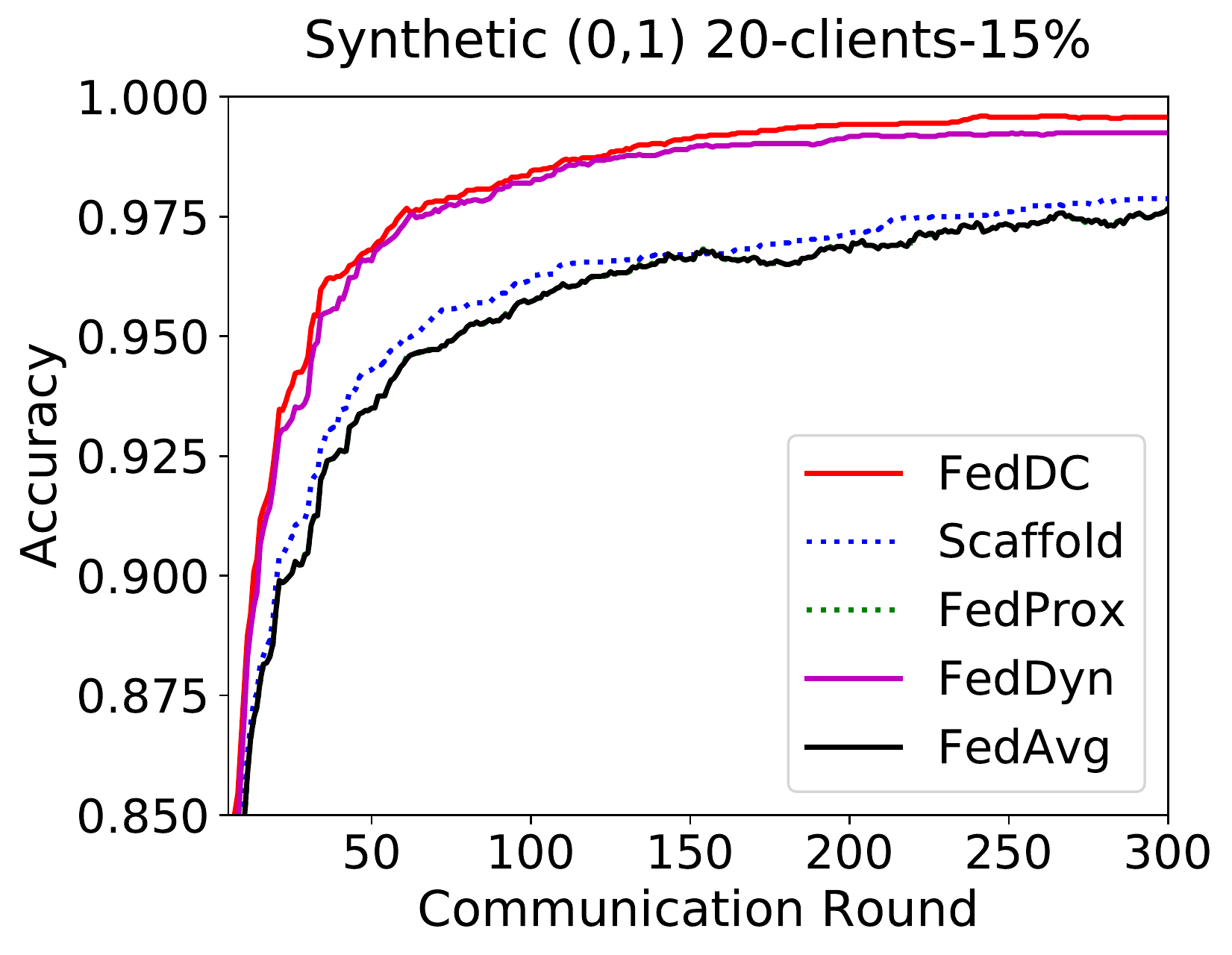}	\caption{}
	\end{subfigure}
	\begin{subfigure}{0.3\linewidth}
		\includegraphics[width=1.0\linewidth]{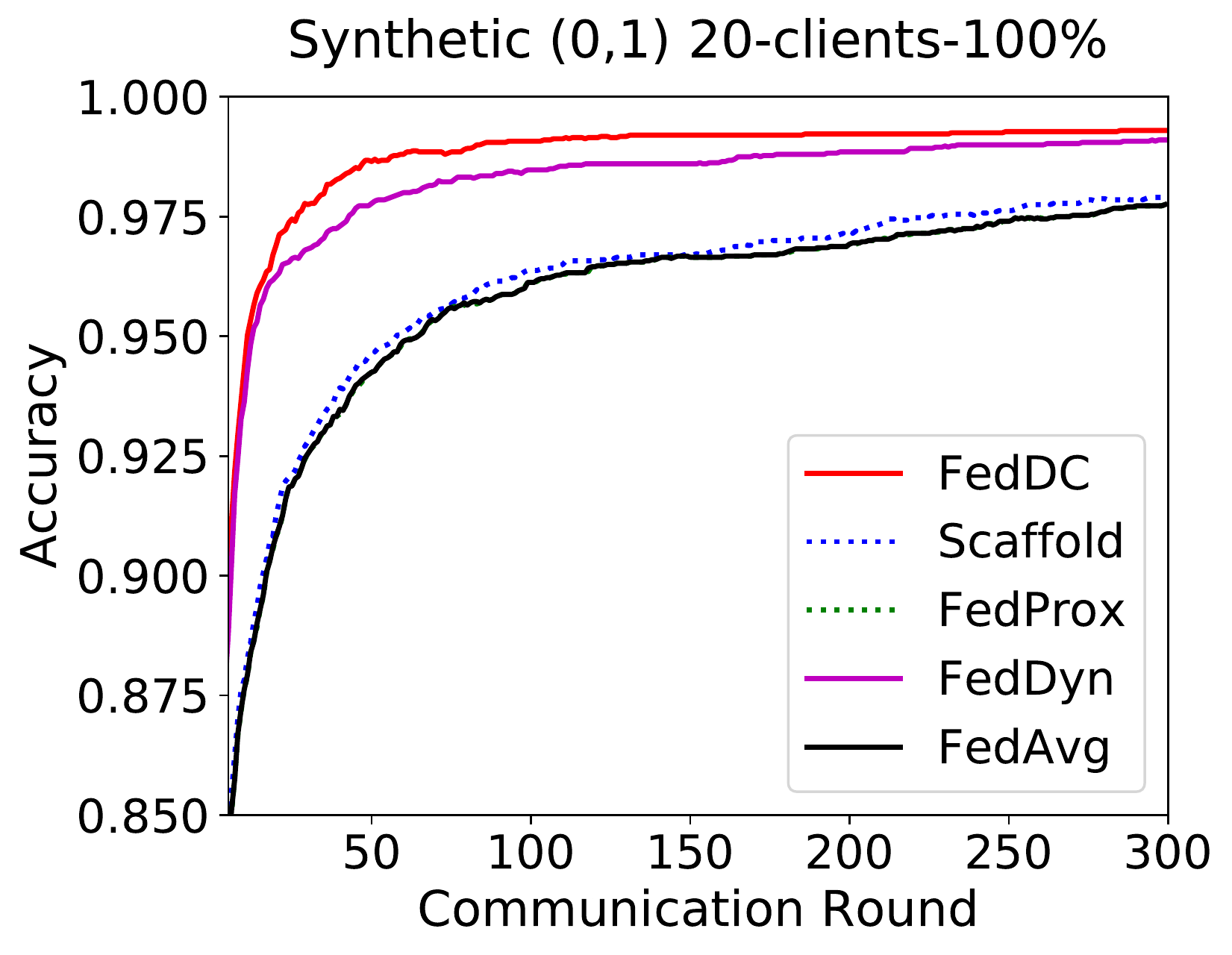}	\caption{}
	\end{subfigure}
	\caption{Convergence plots on Synthetic dataset. There are three types of settings, including the homogeneous setting where $(\gamma_1,\gamma_2)$ equal $(0,0)$, and two heterogeneous settings where $(\gamma_1,\gamma_2)$ equal $(1,0)$ and $(0,1)$, respectively.}\label{fig_syn}
\end{figure}

\begin{figure}[!t]
	\centering

	\begin{subfigure}{0.3\linewidth}
		\includegraphics[width=1.0\linewidth]{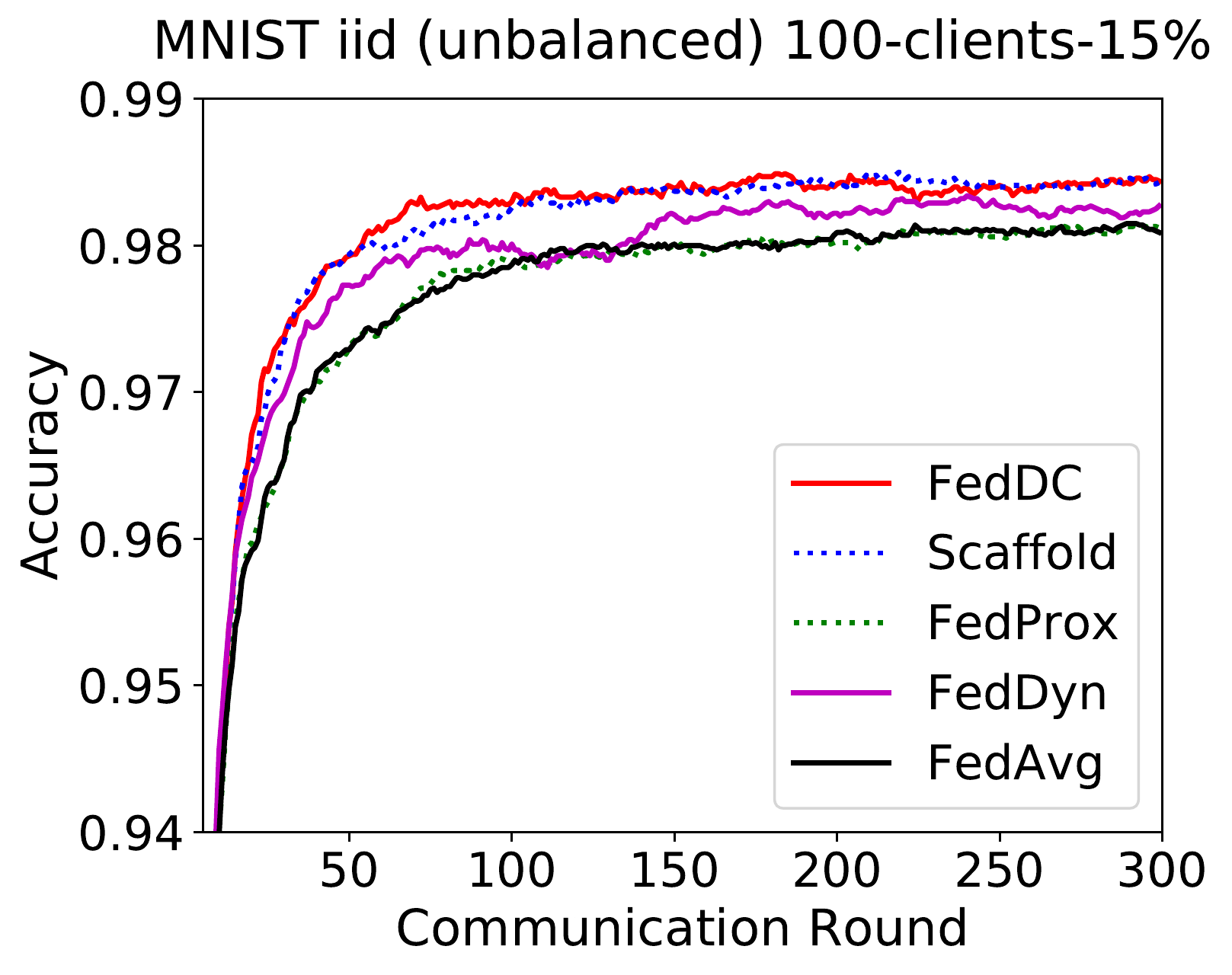}	\caption{}
	\end{subfigure}
	\begin{subfigure}{0.3\linewidth}
		\includegraphics[width=1.0\linewidth]{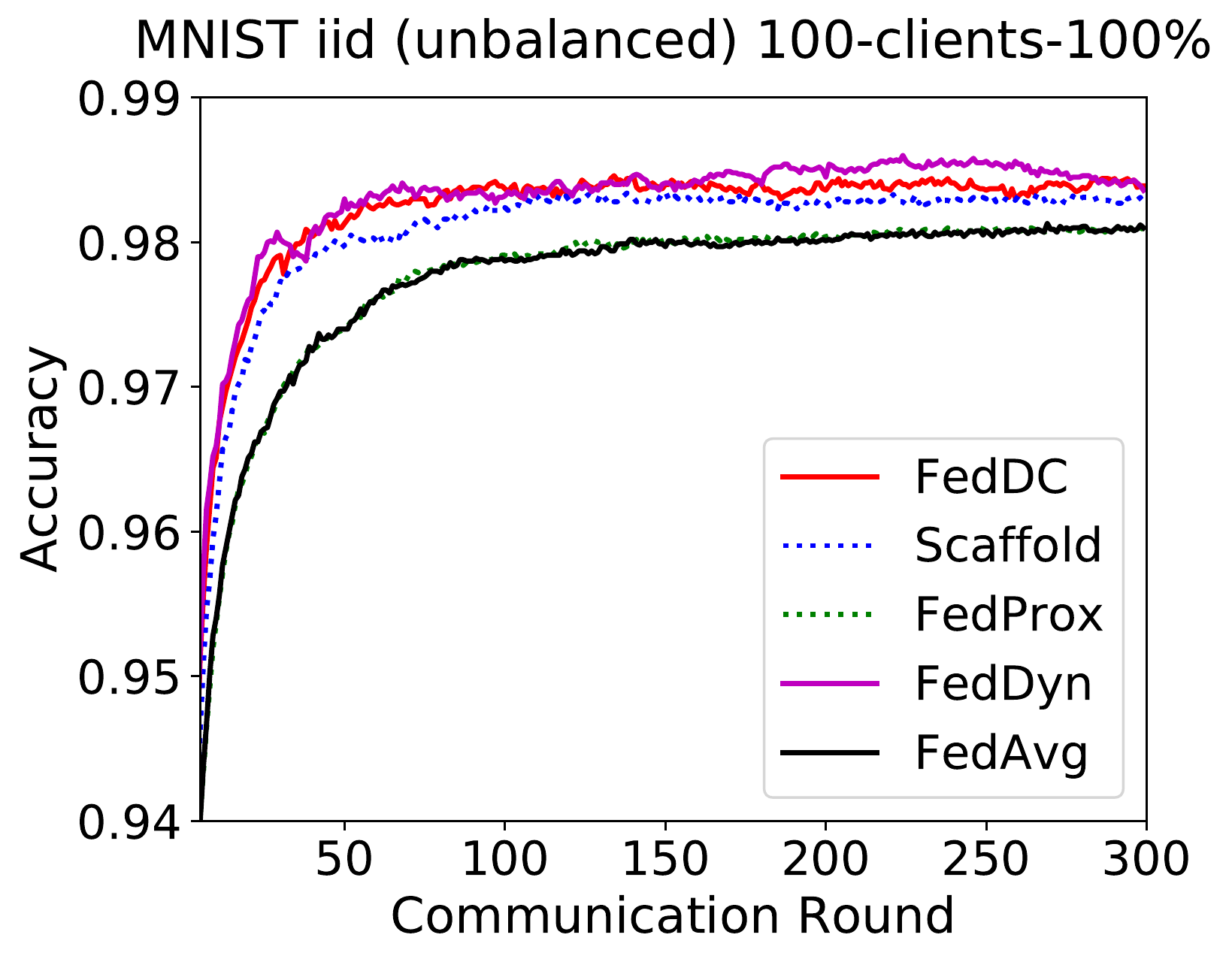}	\caption{}
	\end{subfigure}
	\begin{subfigure}{0.3\linewidth}
		\includegraphics[width=1.0\linewidth]{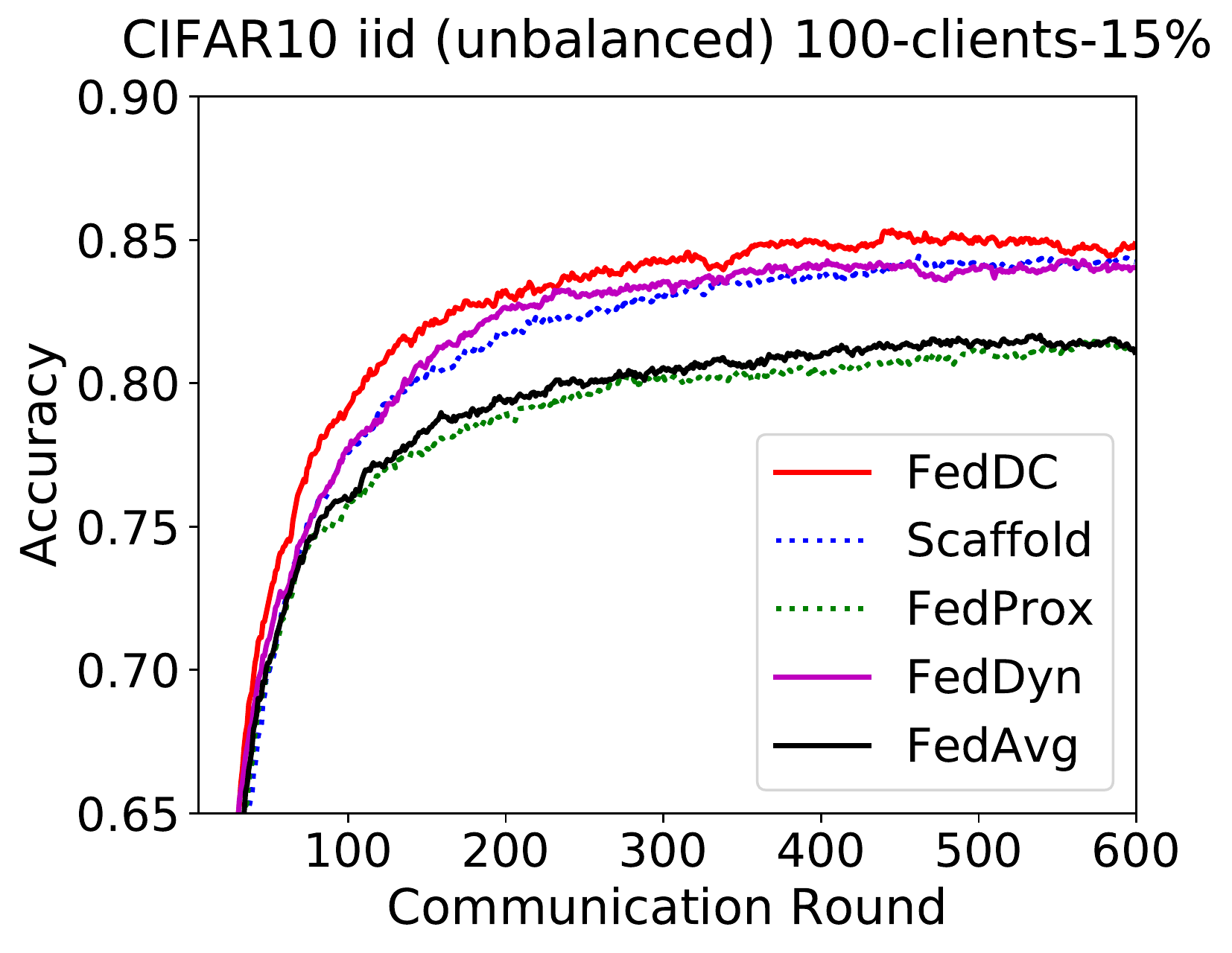}	\caption{}
	\end{subfigure}
\\
	\begin{subfigure}{0.3\linewidth}
		\includegraphics[width=1.0\linewidth]{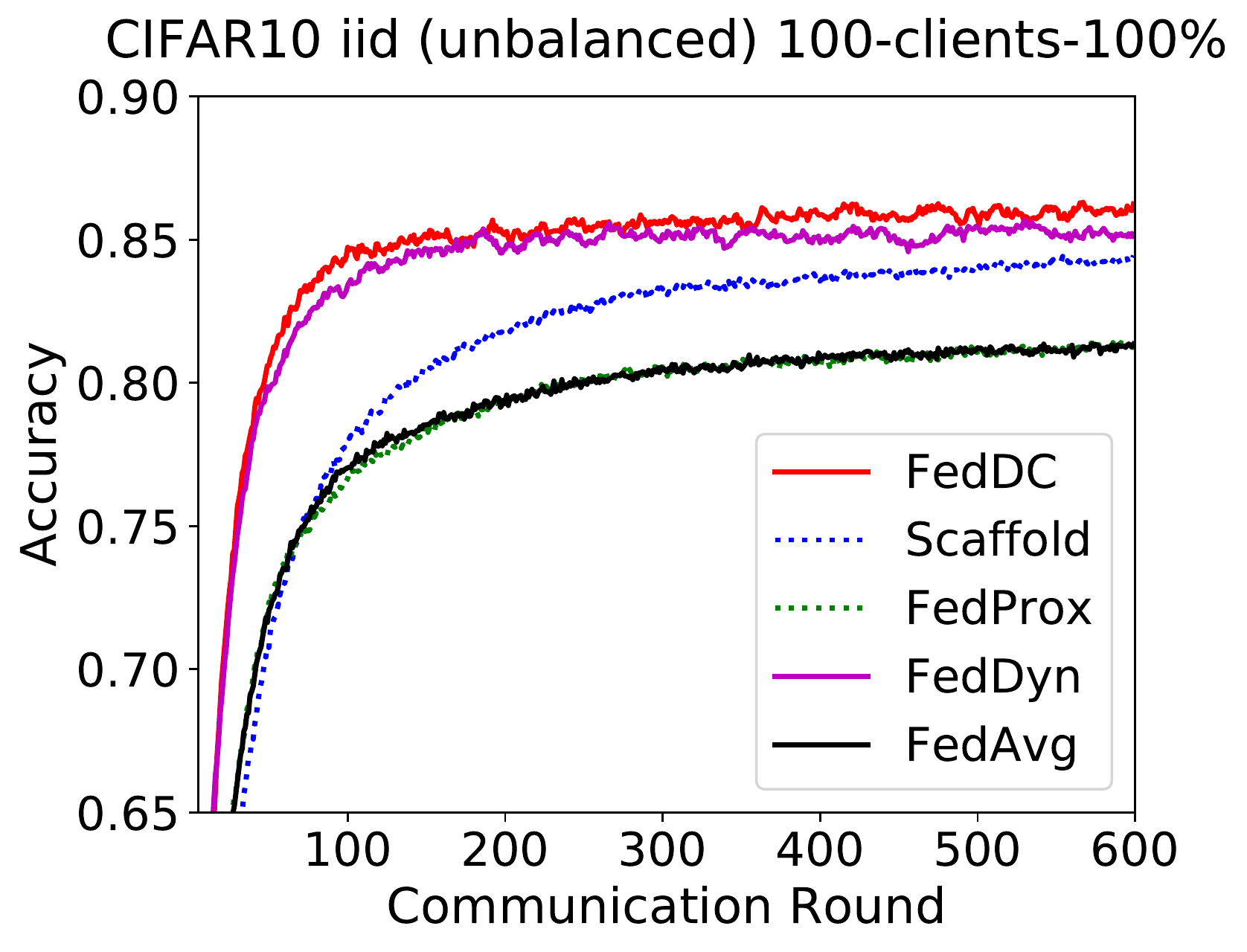}	\caption{}
	\end{subfigure}
	\begin{subfigure}{0.3\linewidth}
		\includegraphics[width=1.0\linewidth]{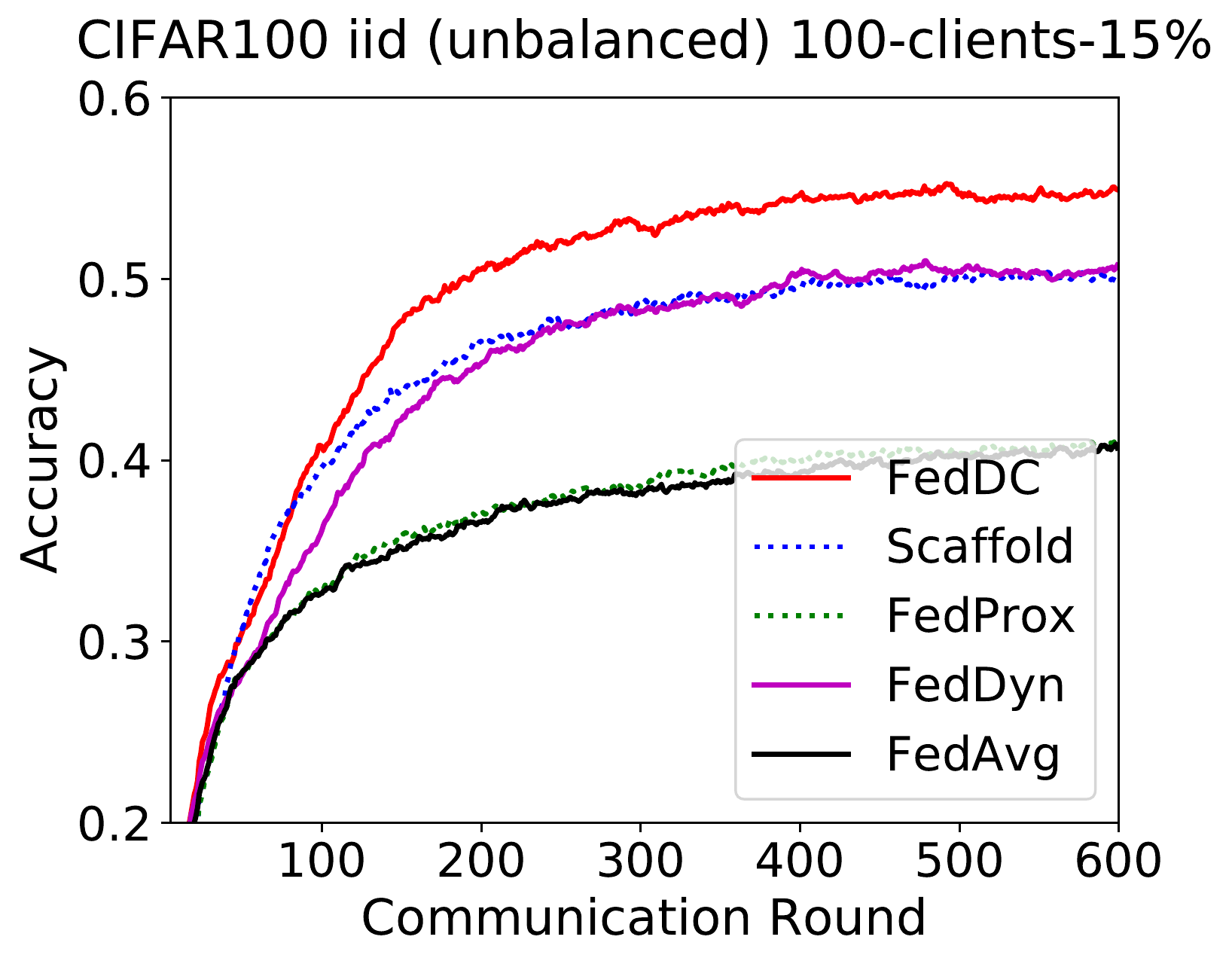}	\caption{}
	\end{subfigure}
	\begin{subfigure}{0.3\linewidth}
		\includegraphics[width=1.0\linewidth]{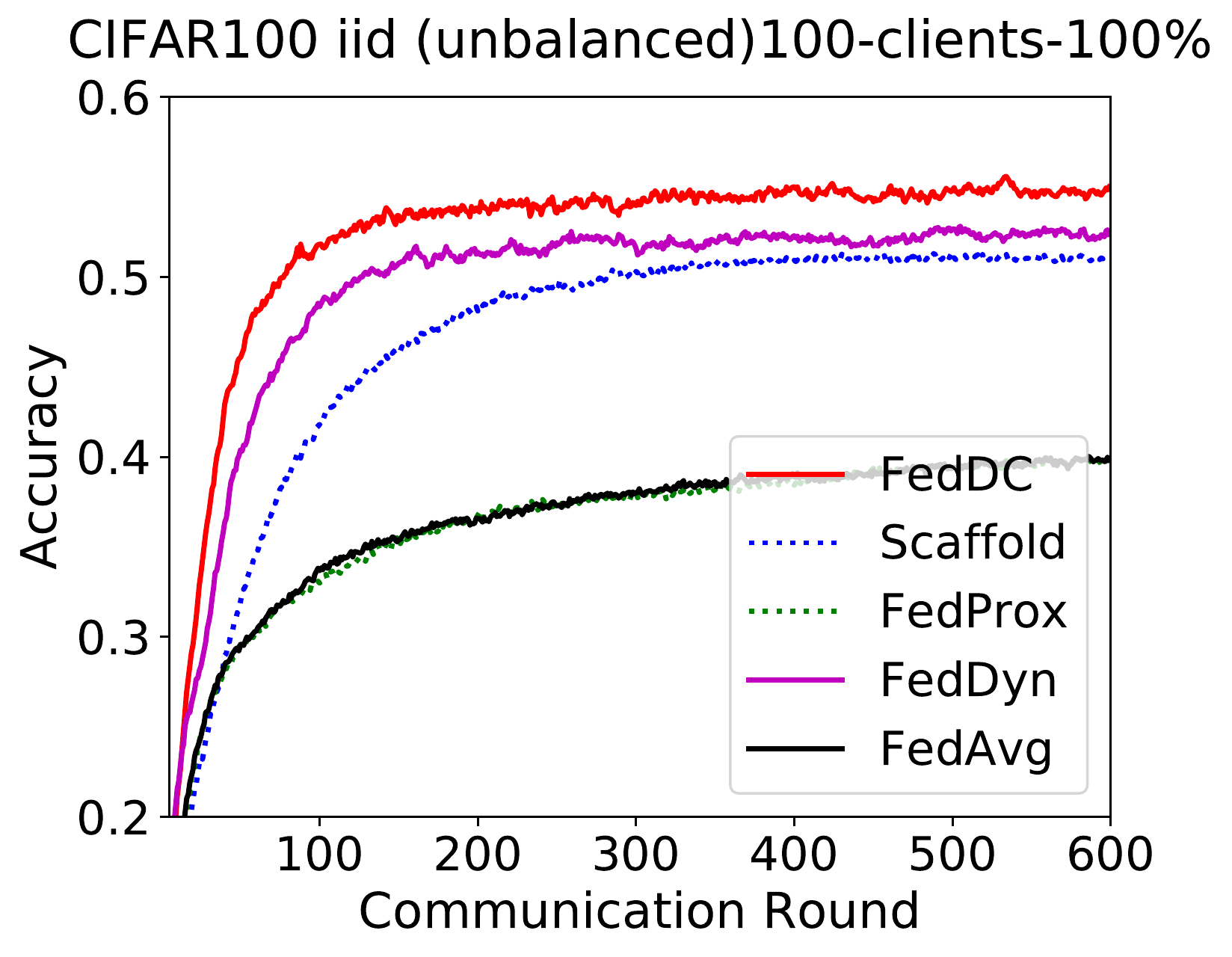}	\caption{}
	\end{subfigure}
	\caption{Convergence plots with 100 clients adopting $100\%$ and $15\%$ client participating settings on unbalanced data of MNIST, CIFAR10 and CIFAR100.}\label{figure_unbalance}
\end{figure}

\begin{figure}[!t]
	\centering

	\begin{subfigure}{0.3\linewidth}
		\includegraphics[width=1.0\linewidth]{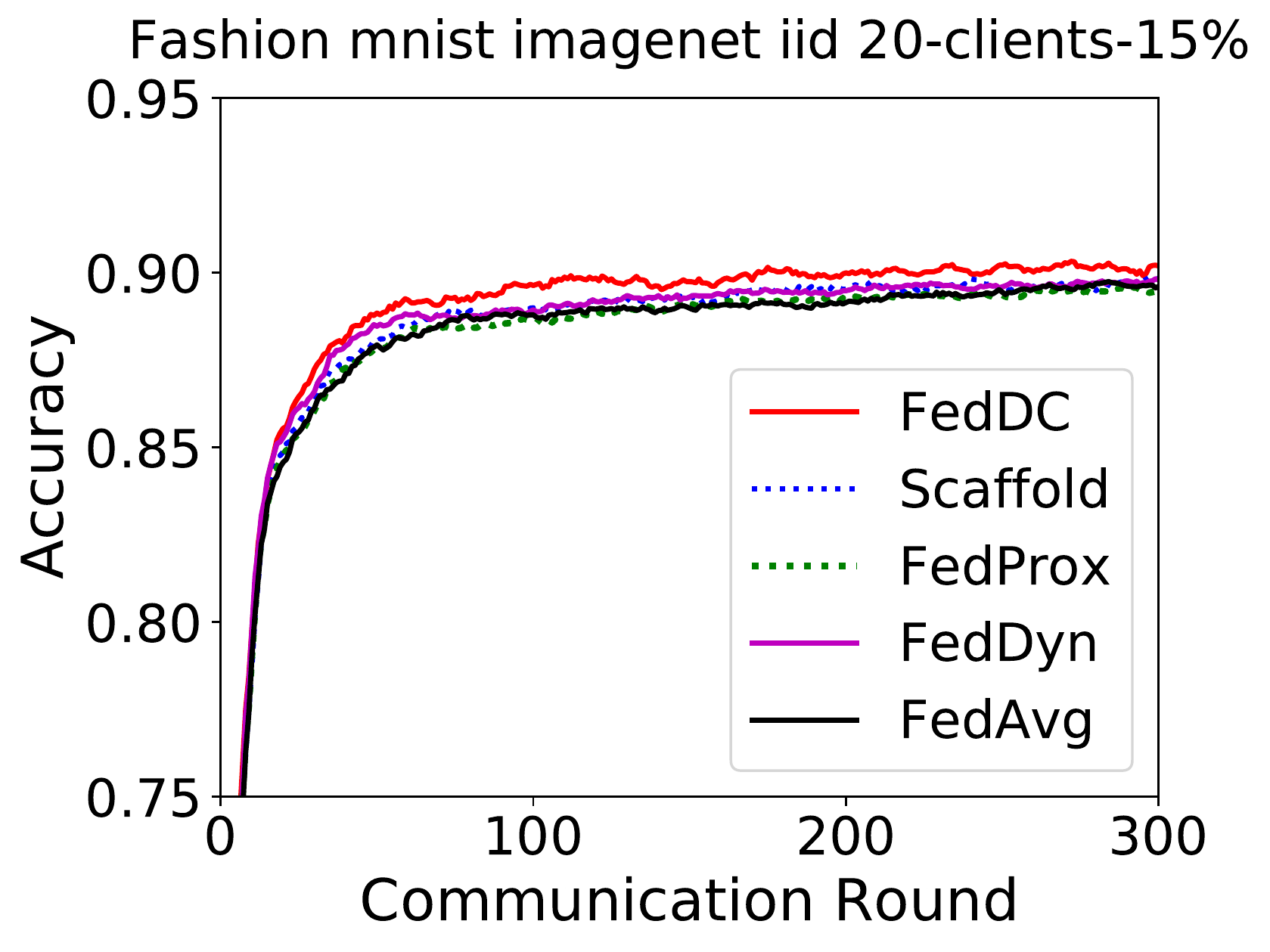}	\caption{}
	\end{subfigure}
	\begin{subfigure}{0.3\linewidth}
		\includegraphics[width=1.0\linewidth]{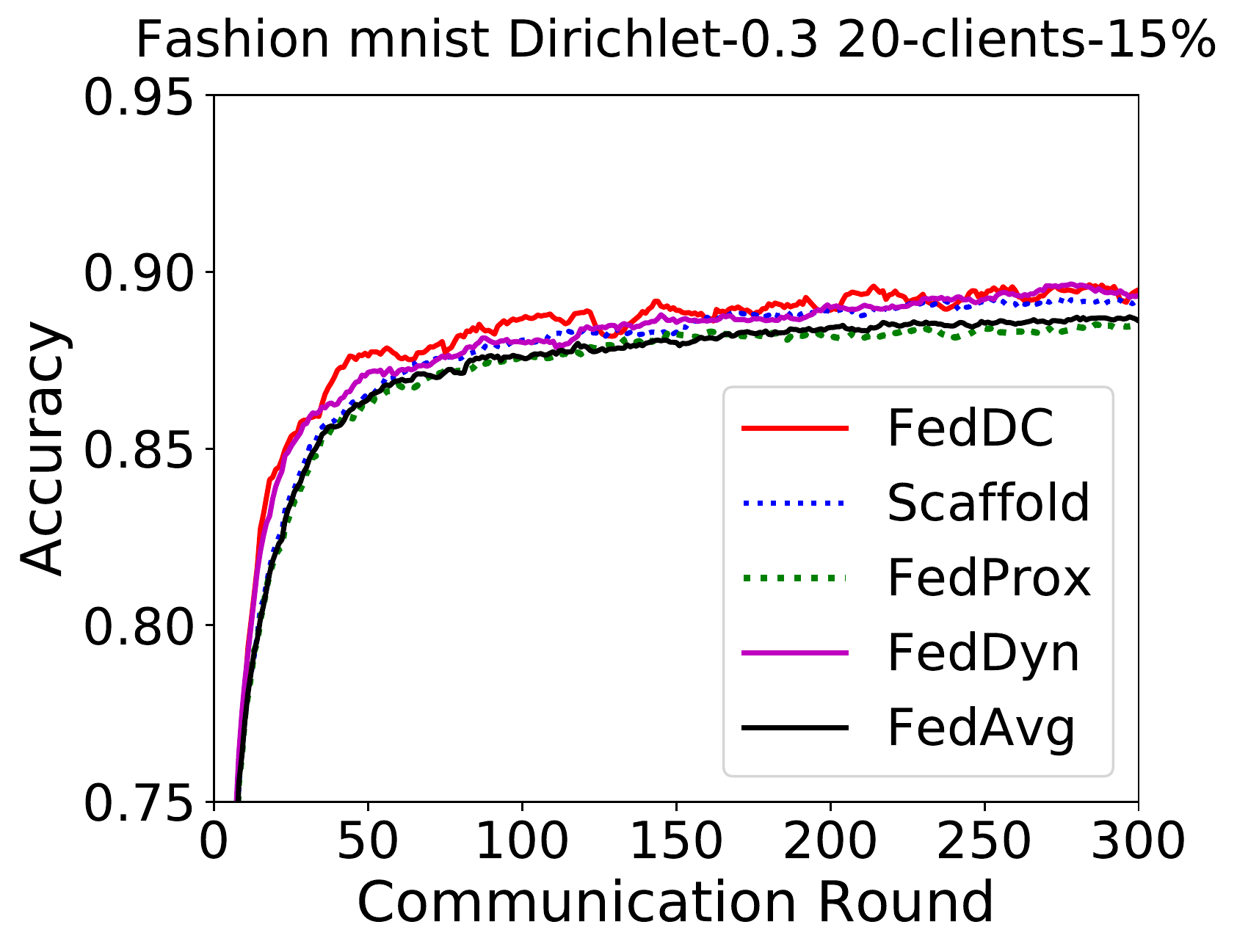}	\caption{}
	\end{subfigure}
	\begin{subfigure}{0.3\linewidth}
		\includegraphics[width=1.0\linewidth]{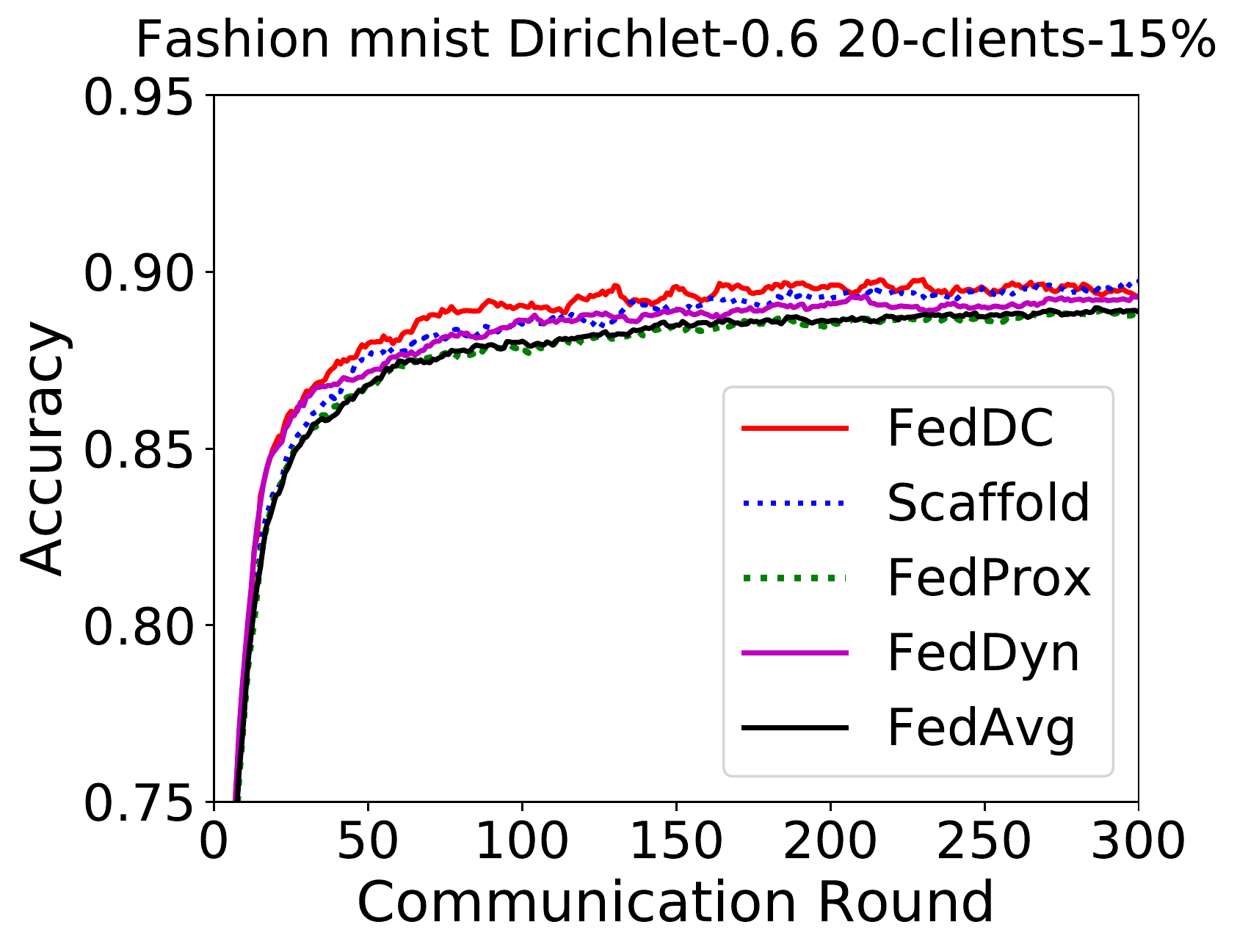}	\caption{}
	\end{subfigure}
	\\
	\begin{subfigure}{0.3\linewidth}
		\includegraphics[width=1.0\linewidth]{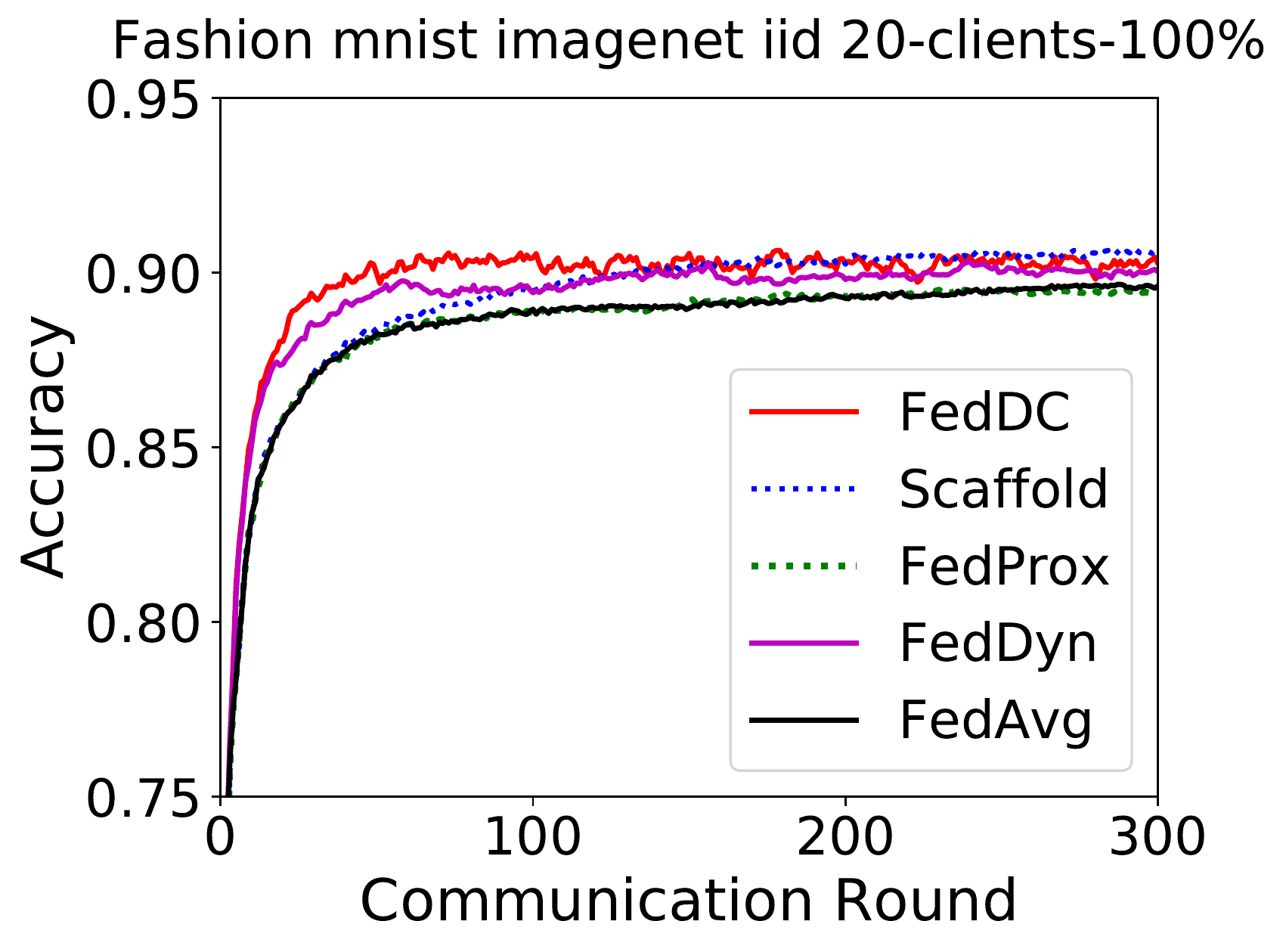}	\caption{}
	\end{subfigure}
	\begin{subfigure}{0.3\linewidth}
		\includegraphics[width=1.0\linewidth]{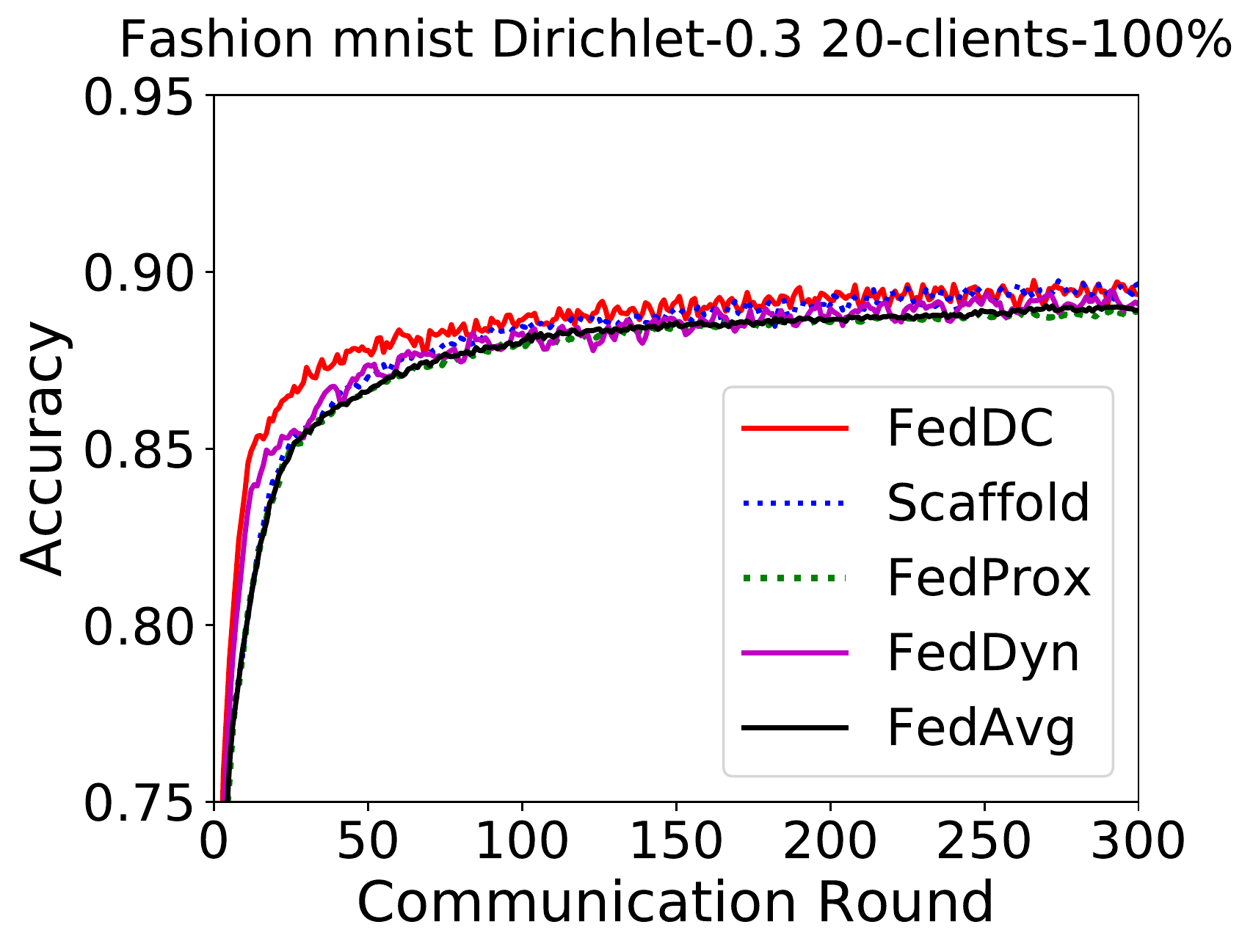}	\caption{}
	\end{subfigure}
	\begin{subfigure}{0.3\linewidth}
		\includegraphics[width=1.0\linewidth]{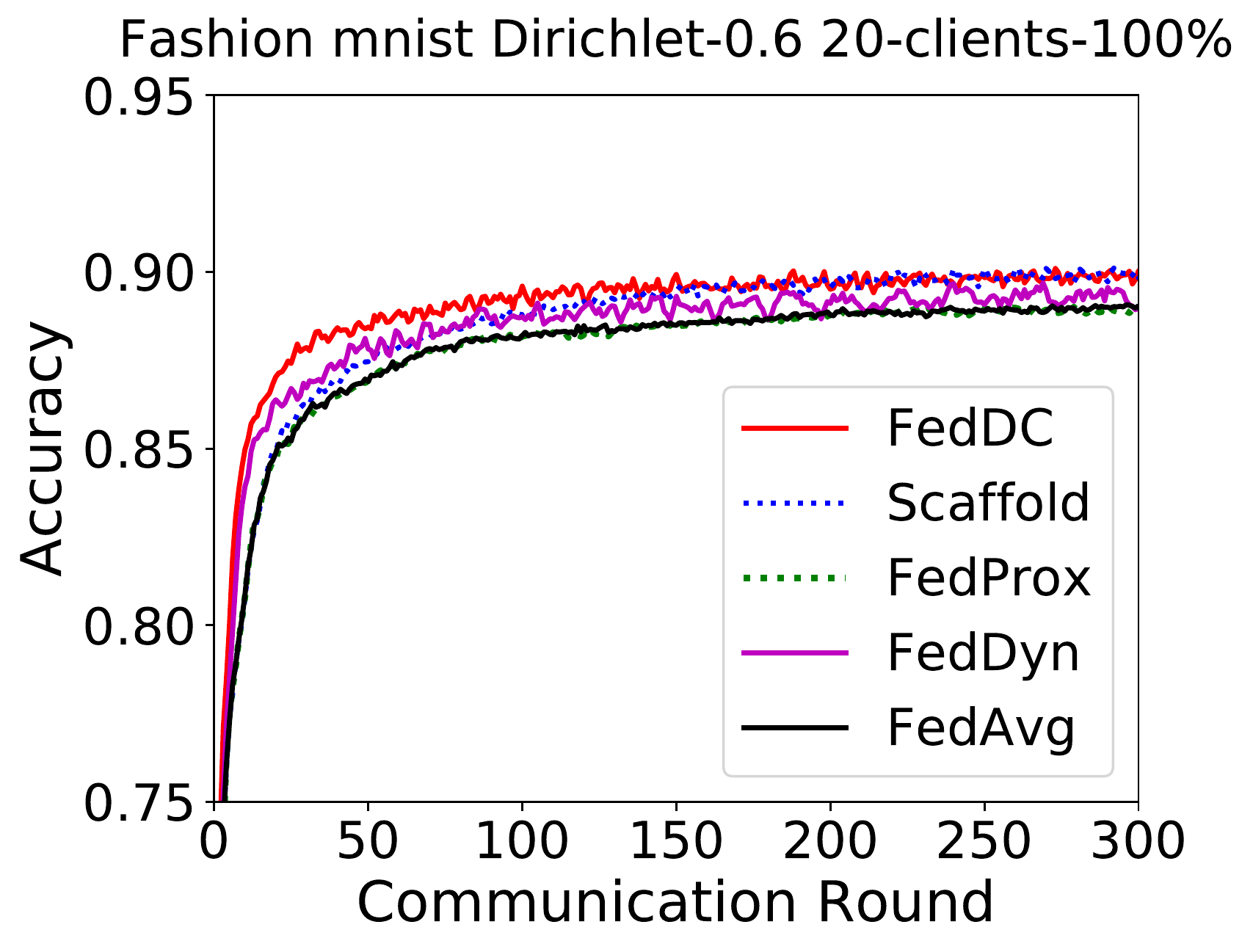}	\caption{}
	\end{subfigure}
	\caption{Convergence plots for iid and non-iid data with 100 clients adopting $100\%$ and $15\%$ client participating settings on fashion MNIST.}\label{figure_fashion}
\end{figure}

\begin{figure}[!t]
	\centering

	\begin{subfigure}{0.3\linewidth}
		\includegraphics[width=1.0\linewidth]{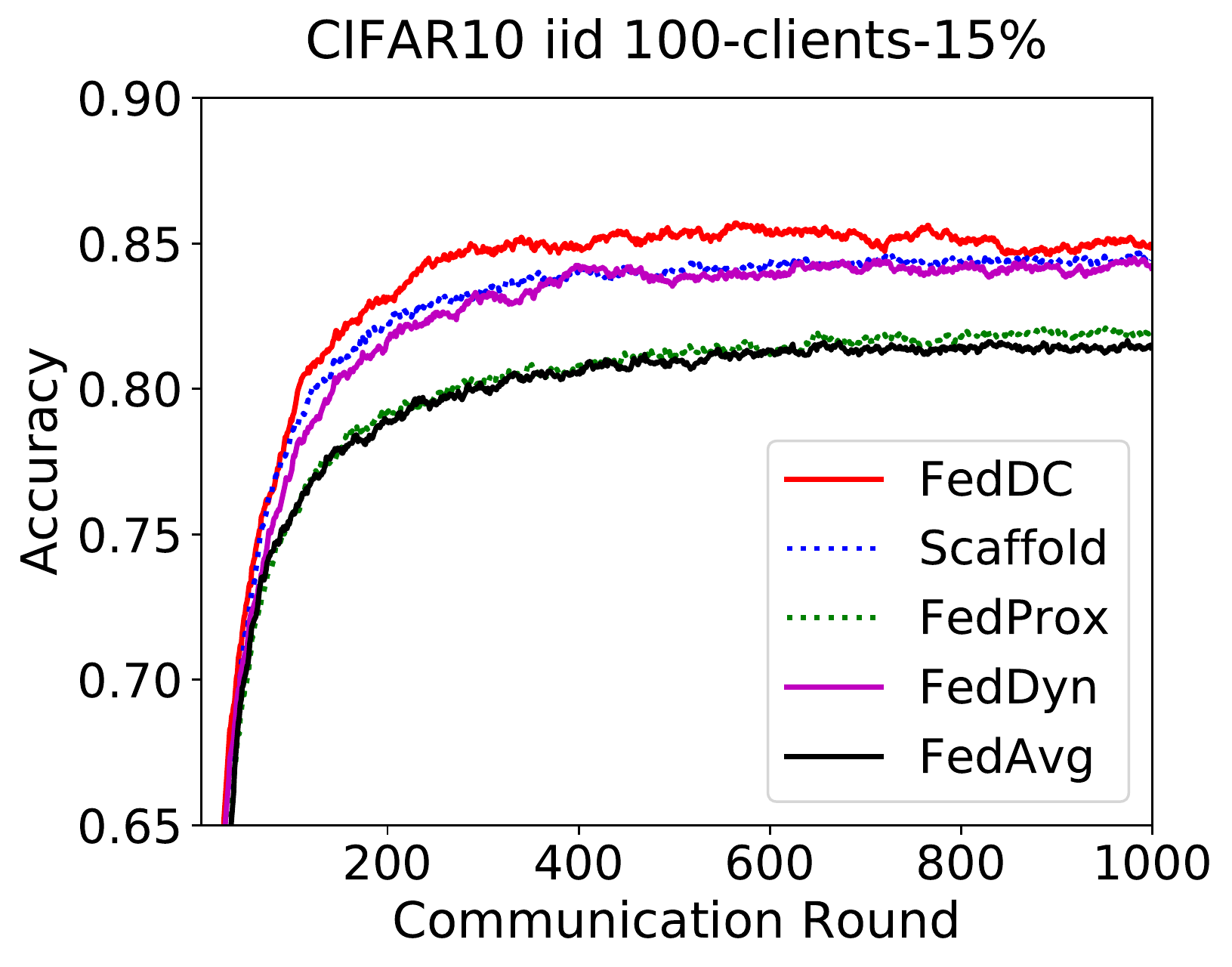}	\caption{}
	\end{subfigure}
	\begin{subfigure}{0.3\linewidth}
		\includegraphics[width=1.0\linewidth]{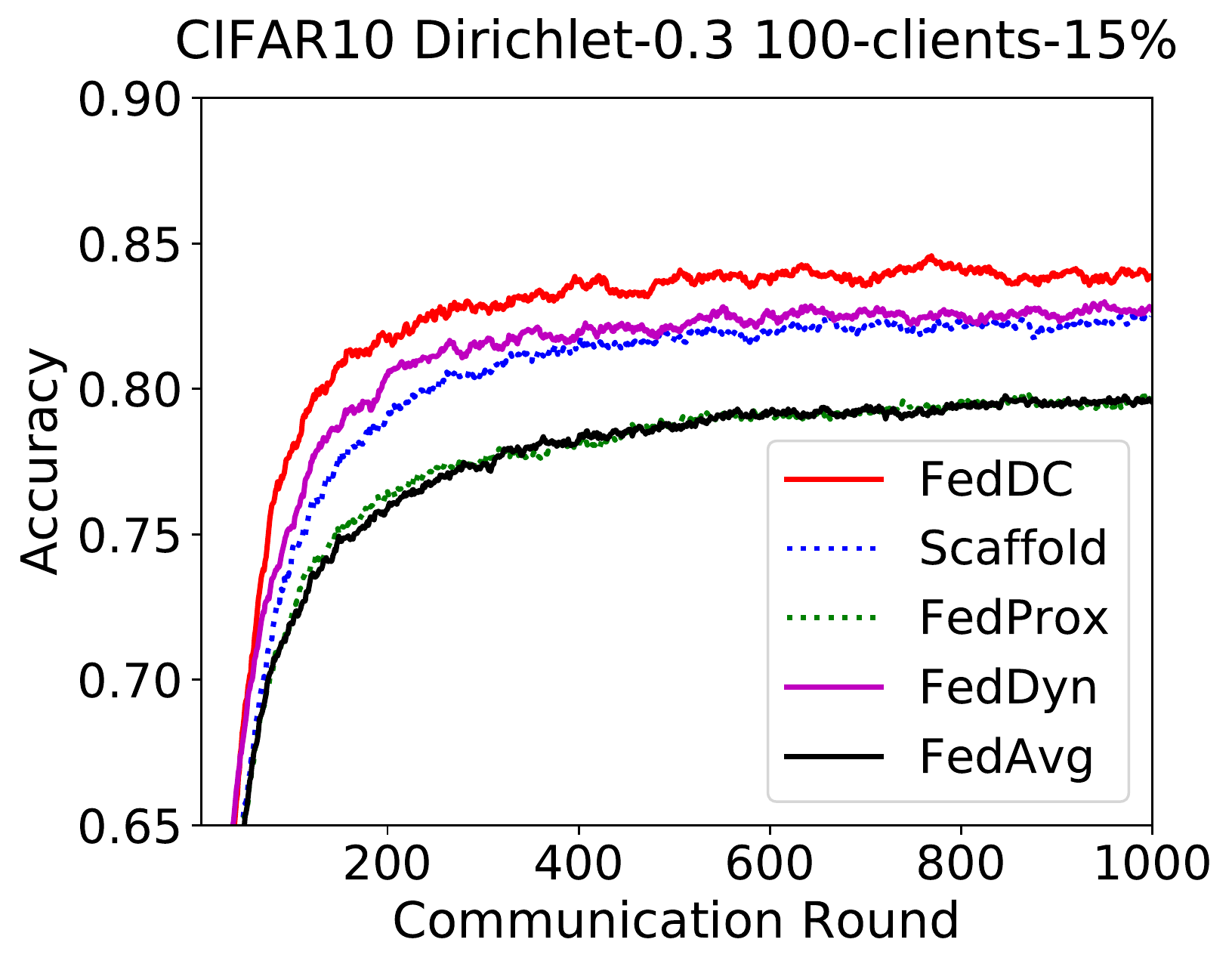}	\caption{}
	\end{subfigure}
	\begin{subfigure}{0.3\linewidth}
		\includegraphics[width=1.0\linewidth]{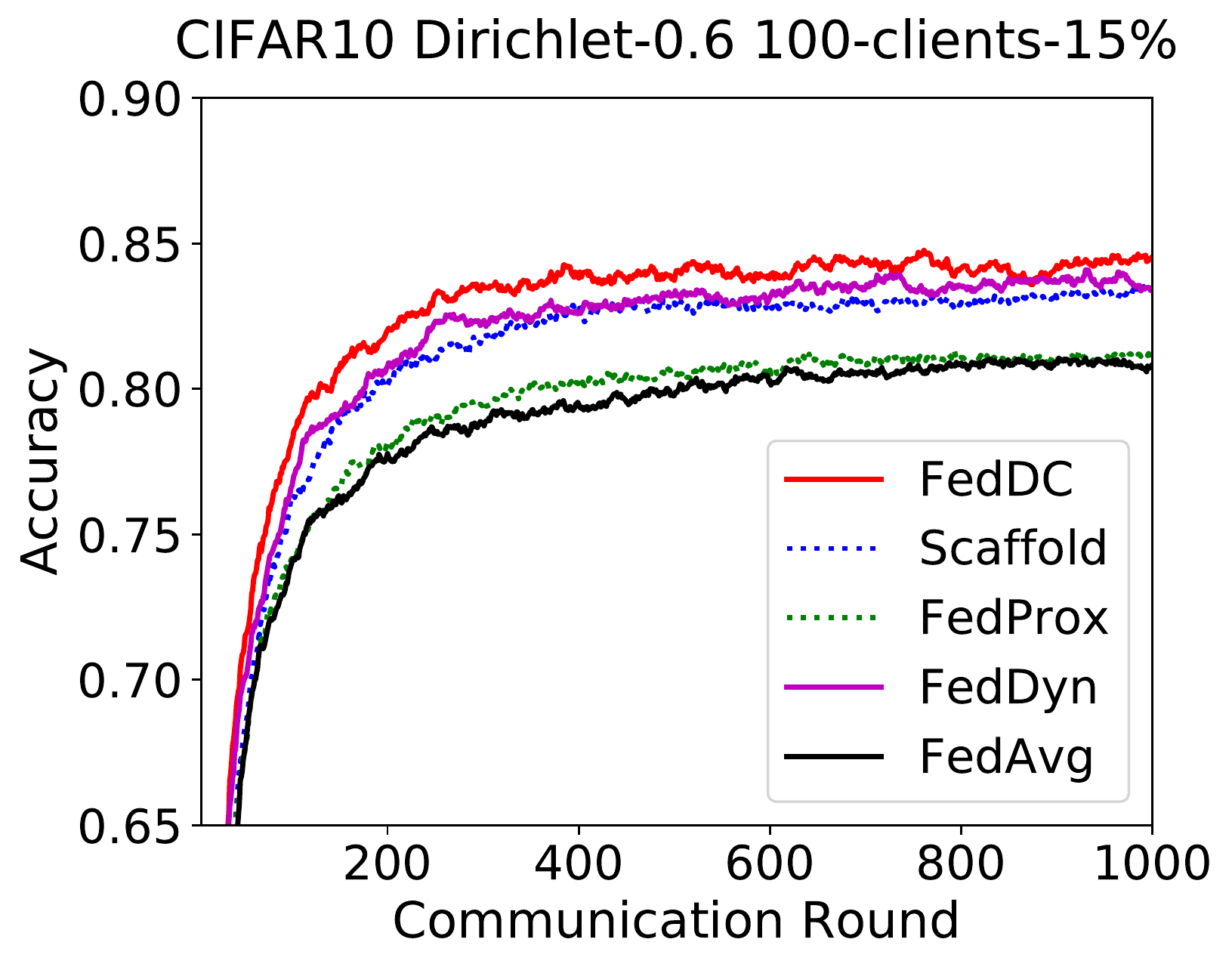}	\caption{}
	\end{subfigure}
	\\
	\begin{subfigure}{0.3\linewidth}
		\includegraphics[width=1.0\linewidth]{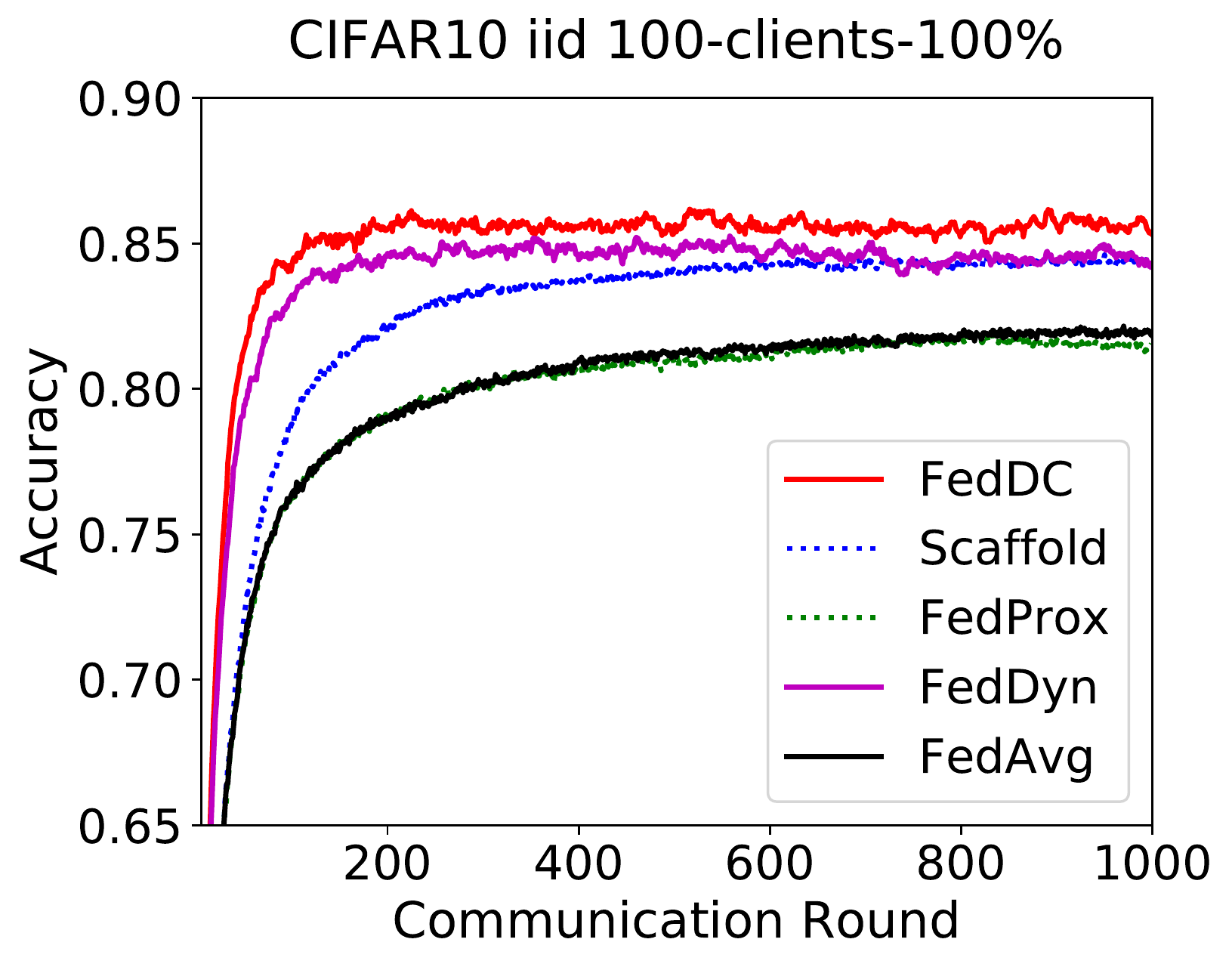}	\caption{}
	\end{subfigure}
	\begin{subfigure}{0.3\linewidth}
		\includegraphics[width=1.0\linewidth]{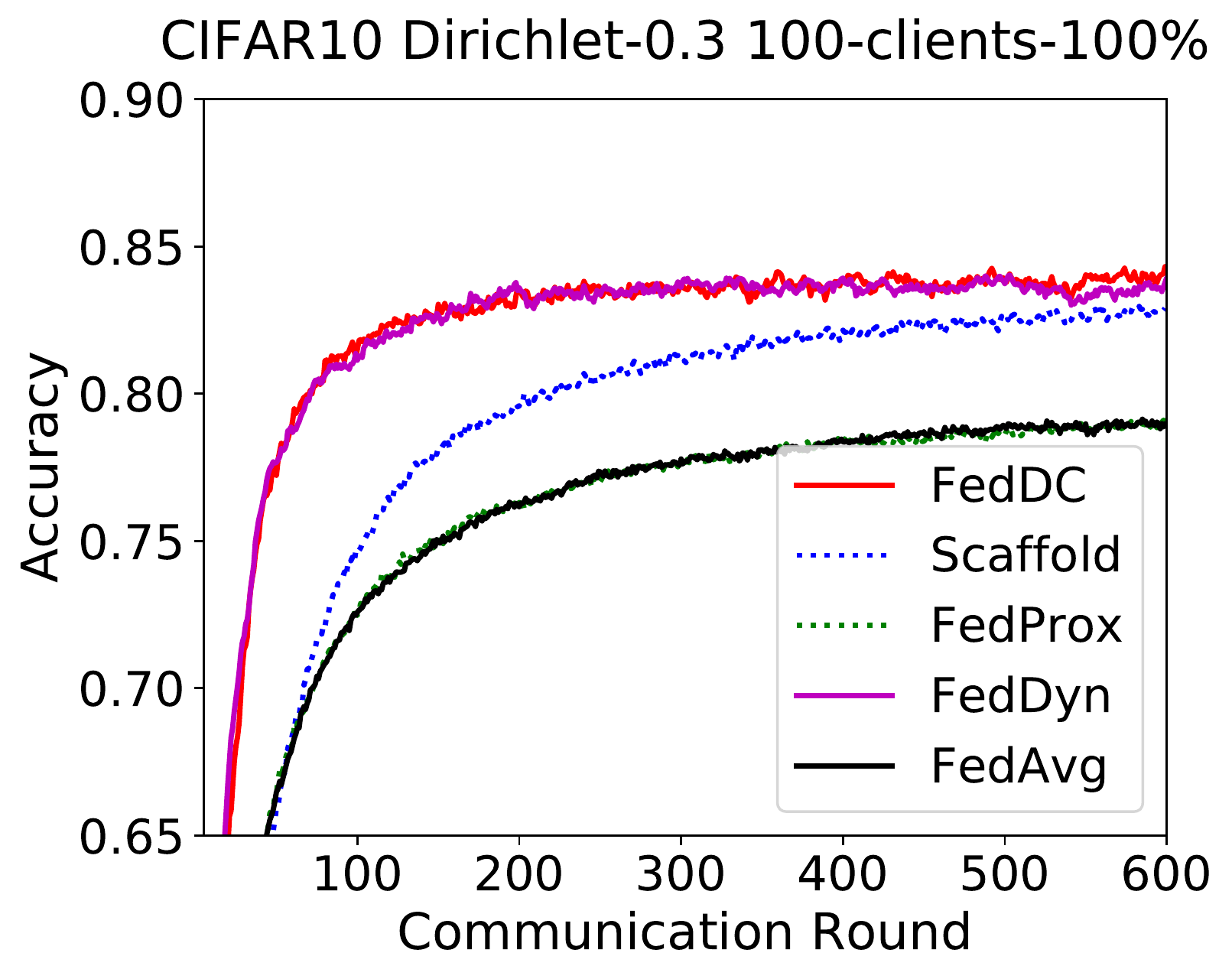}	\caption{}
	\end{subfigure}
	\begin{subfigure}{0.3\linewidth}
		\includegraphics[width=1.0\linewidth]{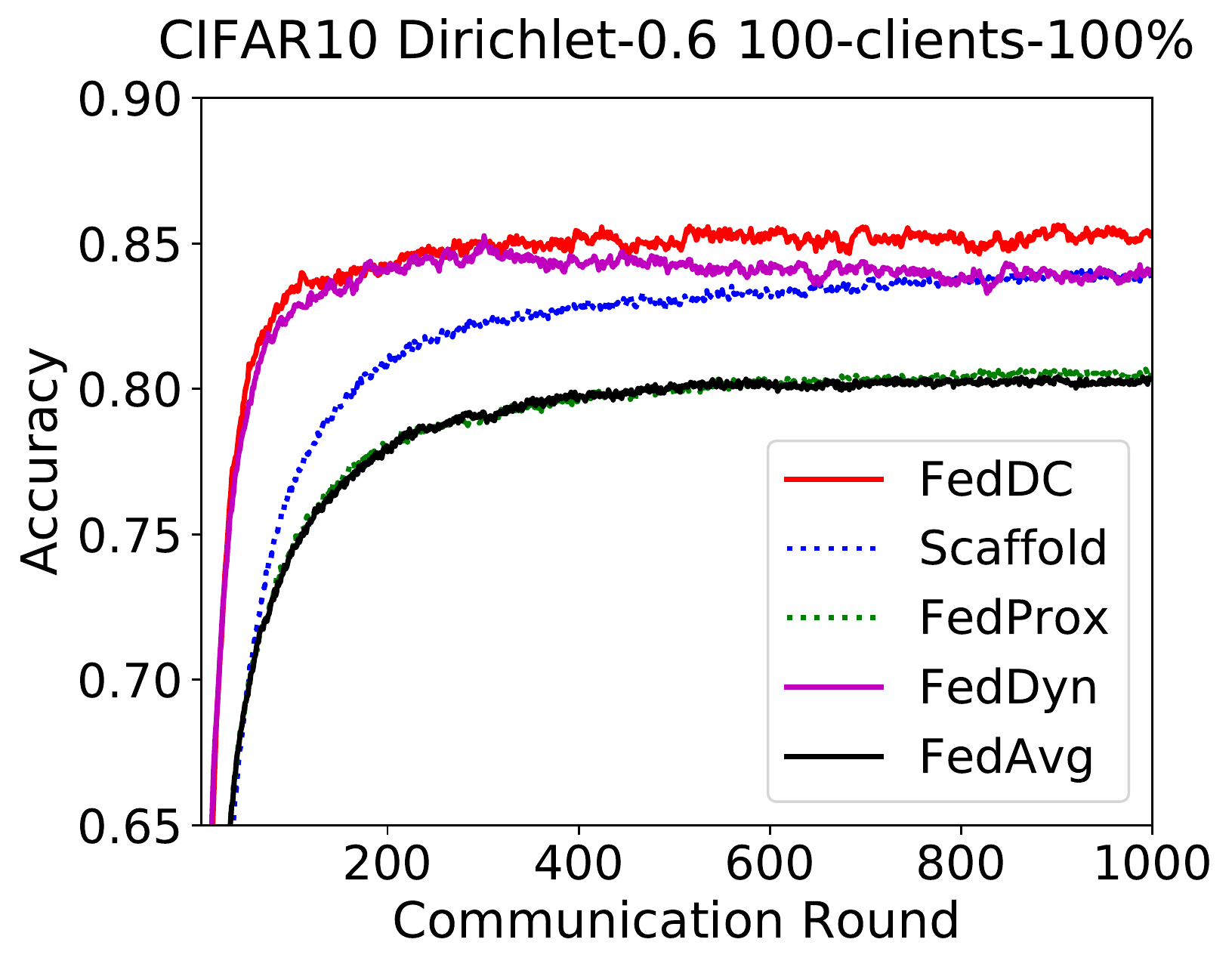}	\caption{}
	\end{subfigure}
	\caption{Convergence plots for iid and non-iid data with 100 clients adopting $100\%$ and $15\%$ client participating settings on CIFAR10.}\label{figure_cifr10}
\end{figure}

\begin{figure}[!t]
	\centering

	\begin{subfigure}{0.3\linewidth}
		\includegraphics[width=1.0\linewidth]{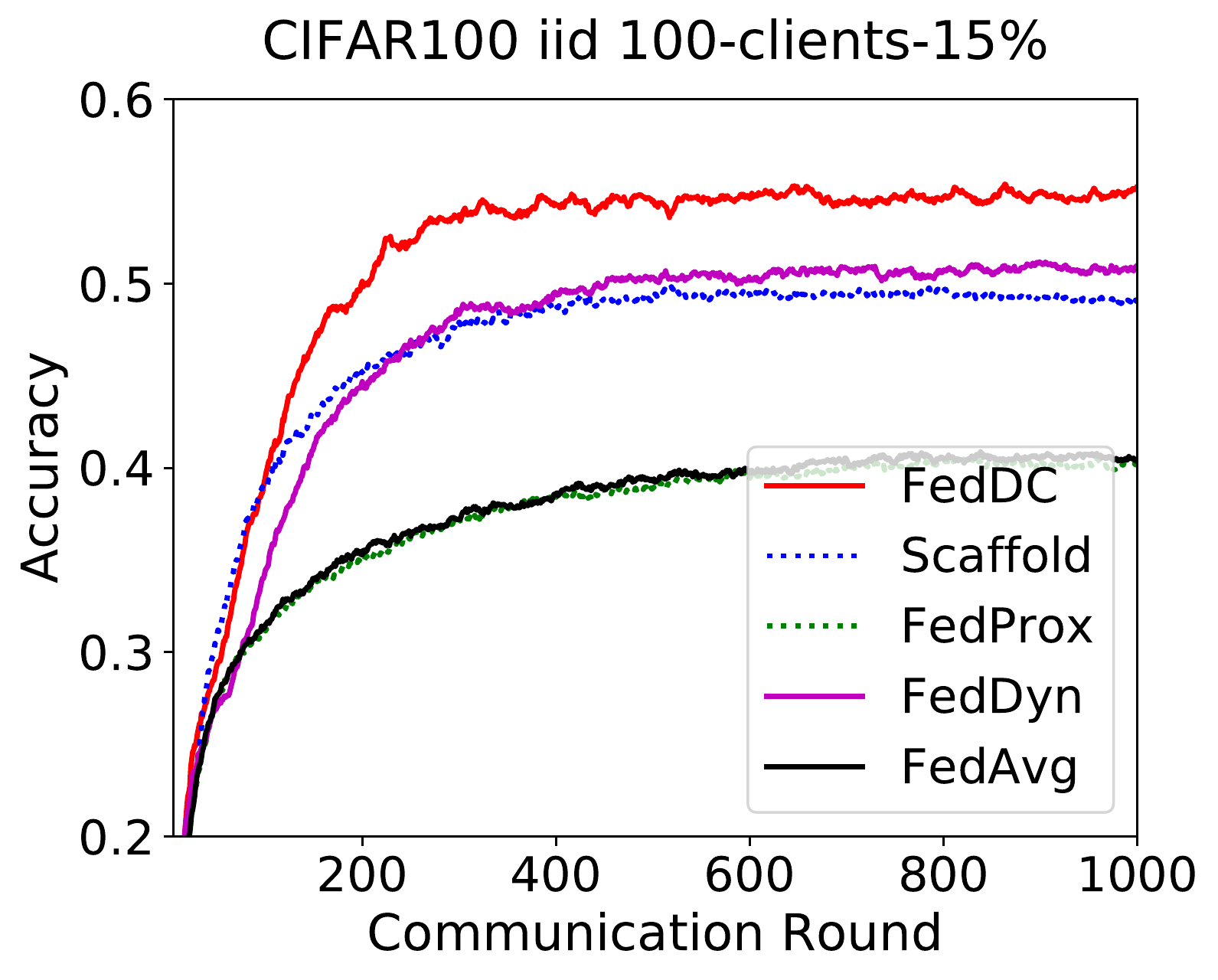}	\caption{}
	\end{subfigure}
	\begin{subfigure}{0.3\linewidth}
		\includegraphics[width=1.0\linewidth]{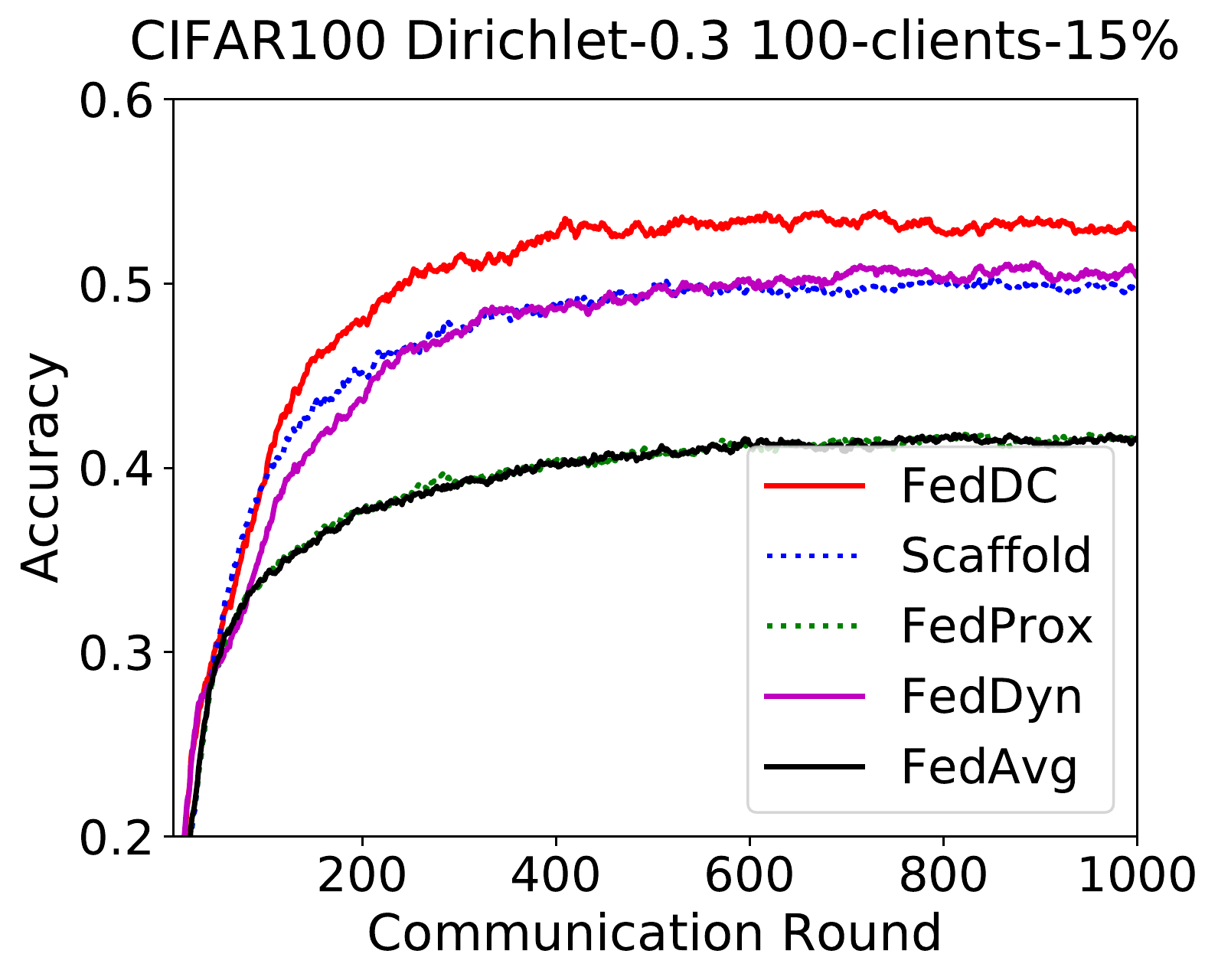}	\caption{}
	\end{subfigure}
	\begin{subfigure}{0.3\linewidth}
		\includegraphics[width=1.0\linewidth]{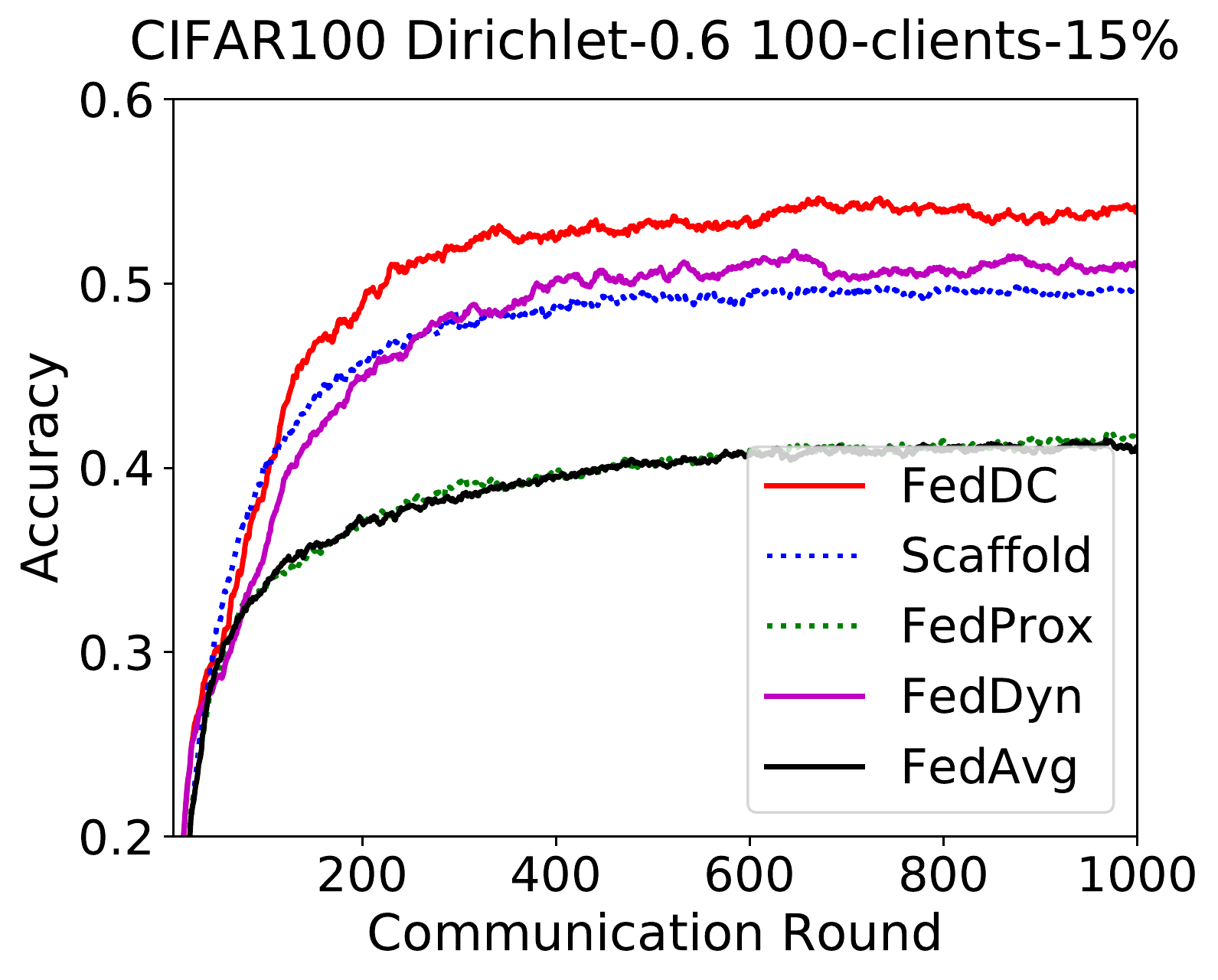}	\caption{}
	\end{subfigure}
	\\
	\begin{subfigure}{0.3\linewidth}
		\includegraphics[width=1.0\linewidth]{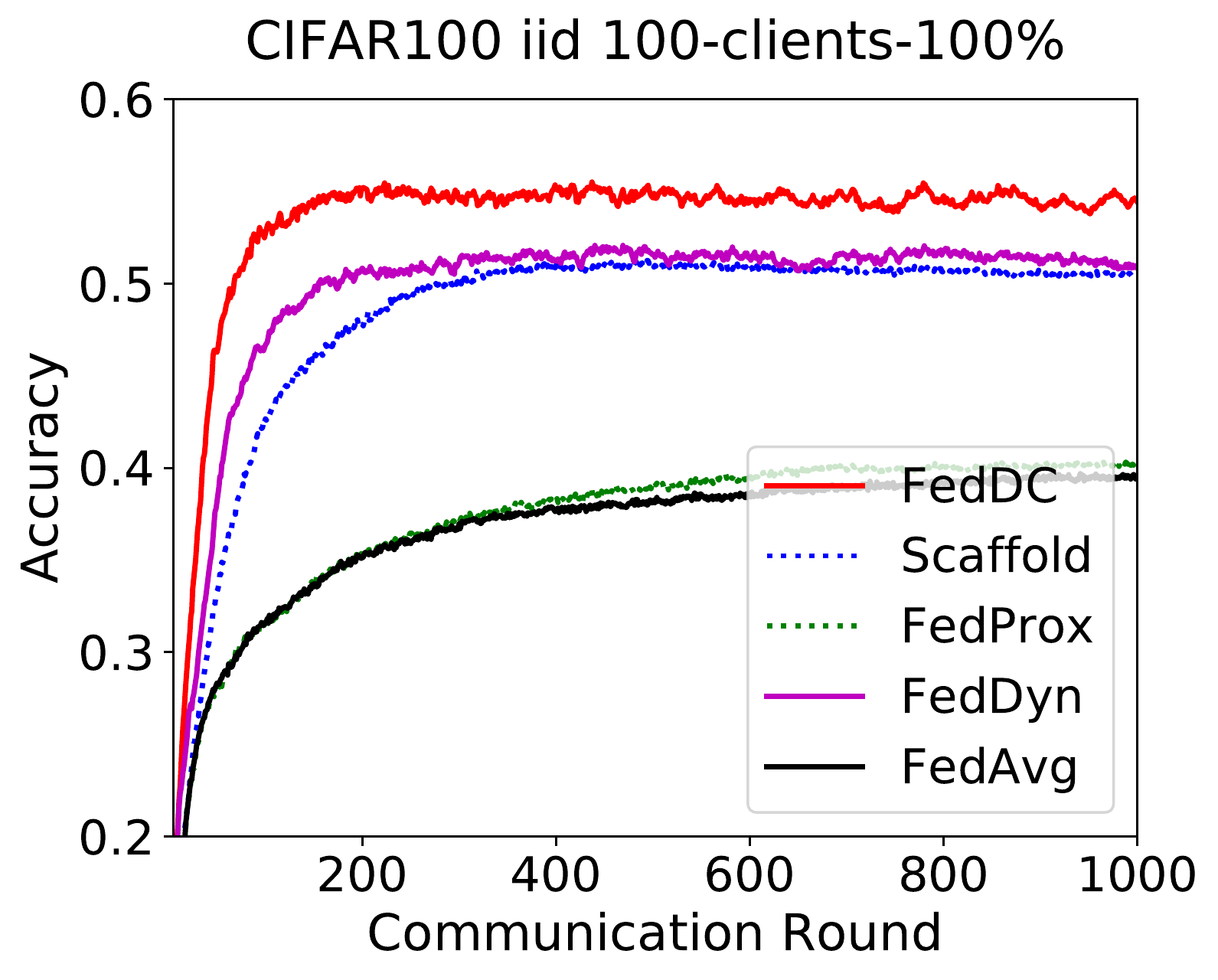}	\caption{}
	\end{subfigure}
	\begin{subfigure}{0.3\linewidth}
		\includegraphics[width=1.0\linewidth]{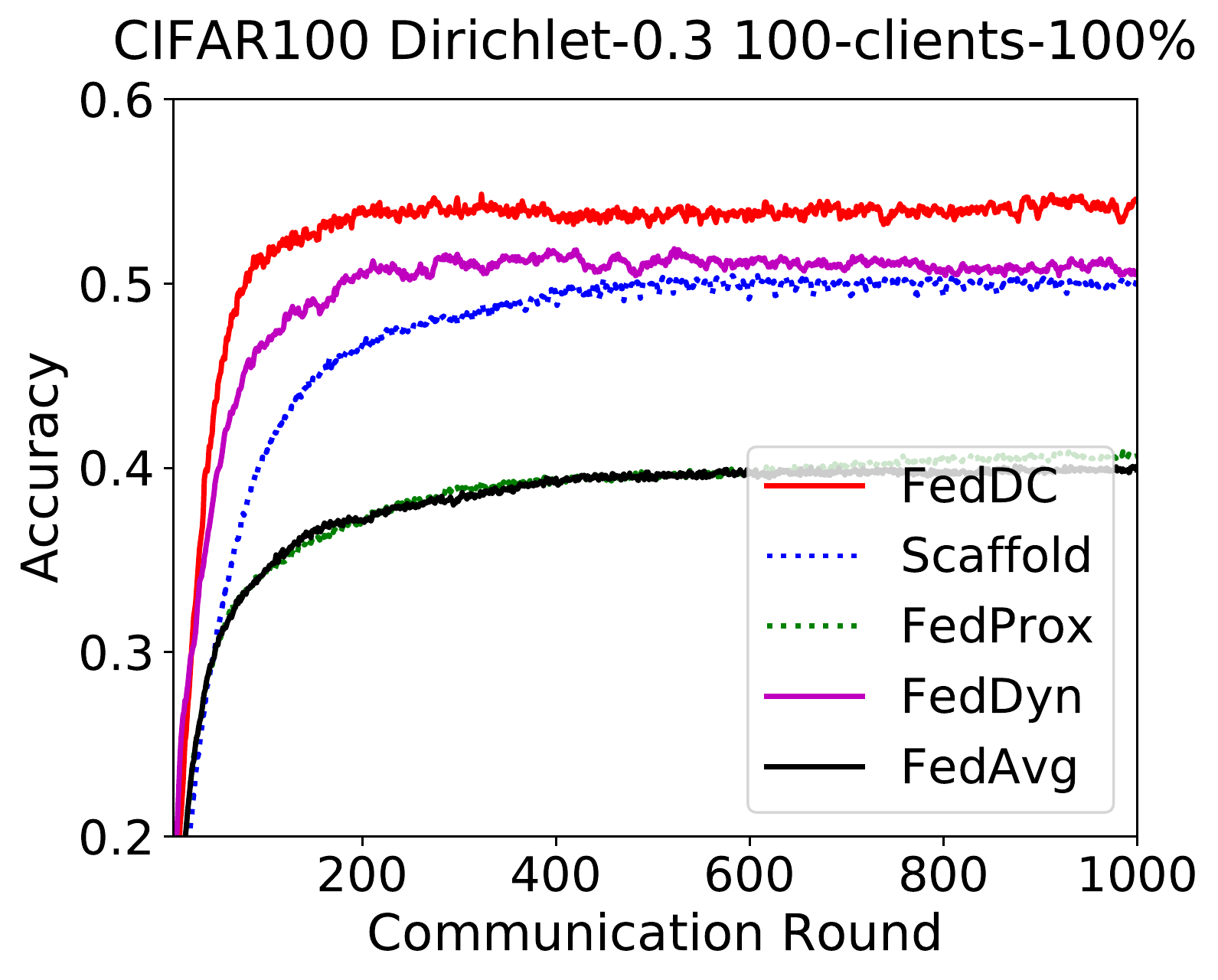}	\caption{}
	\end{subfigure}
	\begin{subfigure}{0.3\linewidth}
		\includegraphics[width=1.0\linewidth]{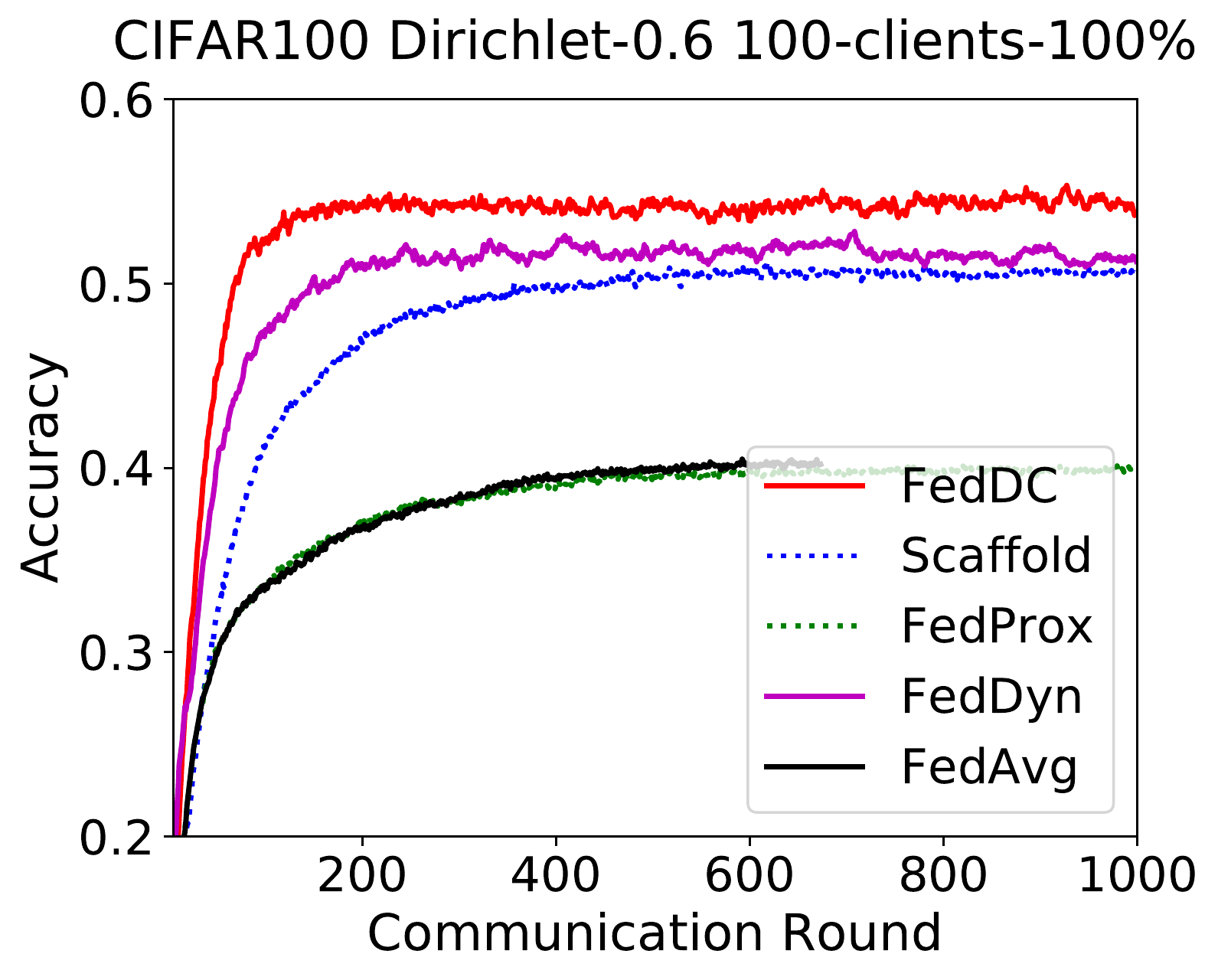}	\caption{}
	\end{subfigure}
	\caption{Convergence plots for iid and non-iid data with 100 clients adopting $100\%$ and $15\%$ client participating settings on CIFAR100.}\label{figure_cifr100}
\end{figure}

\begin{figure}[!t]
	\centering

	\begin{subfigure}{0.3\linewidth}
		\includegraphics[width=1.0\linewidth]{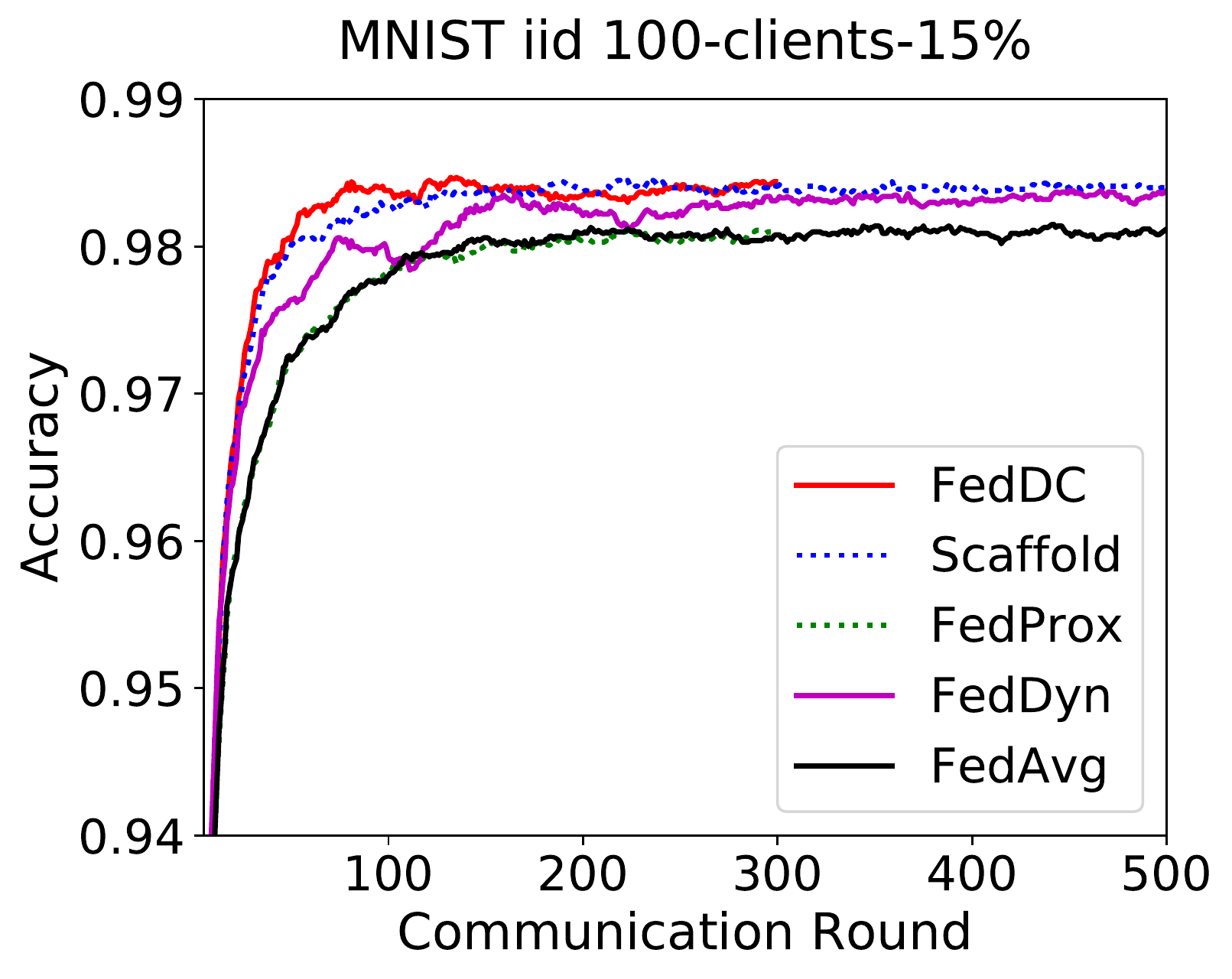}	\caption{}
	\end{subfigure}
	\begin{subfigure}{0.3\linewidth}
		\includegraphics[width=1.0\linewidth]{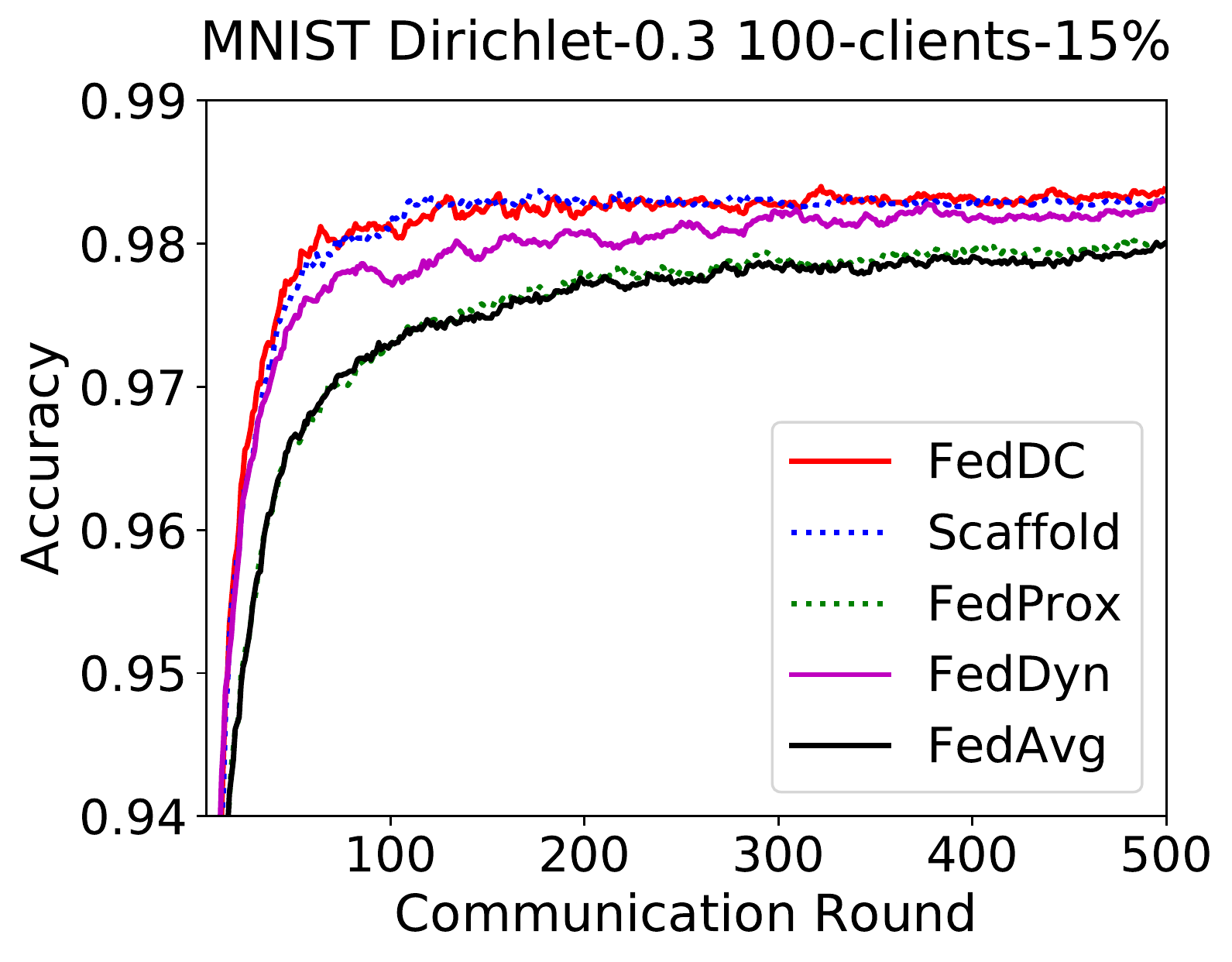}	\caption{}
	\end{subfigure}
	\begin{subfigure}{0.3\linewidth}
		\includegraphics[width=1.0\linewidth]{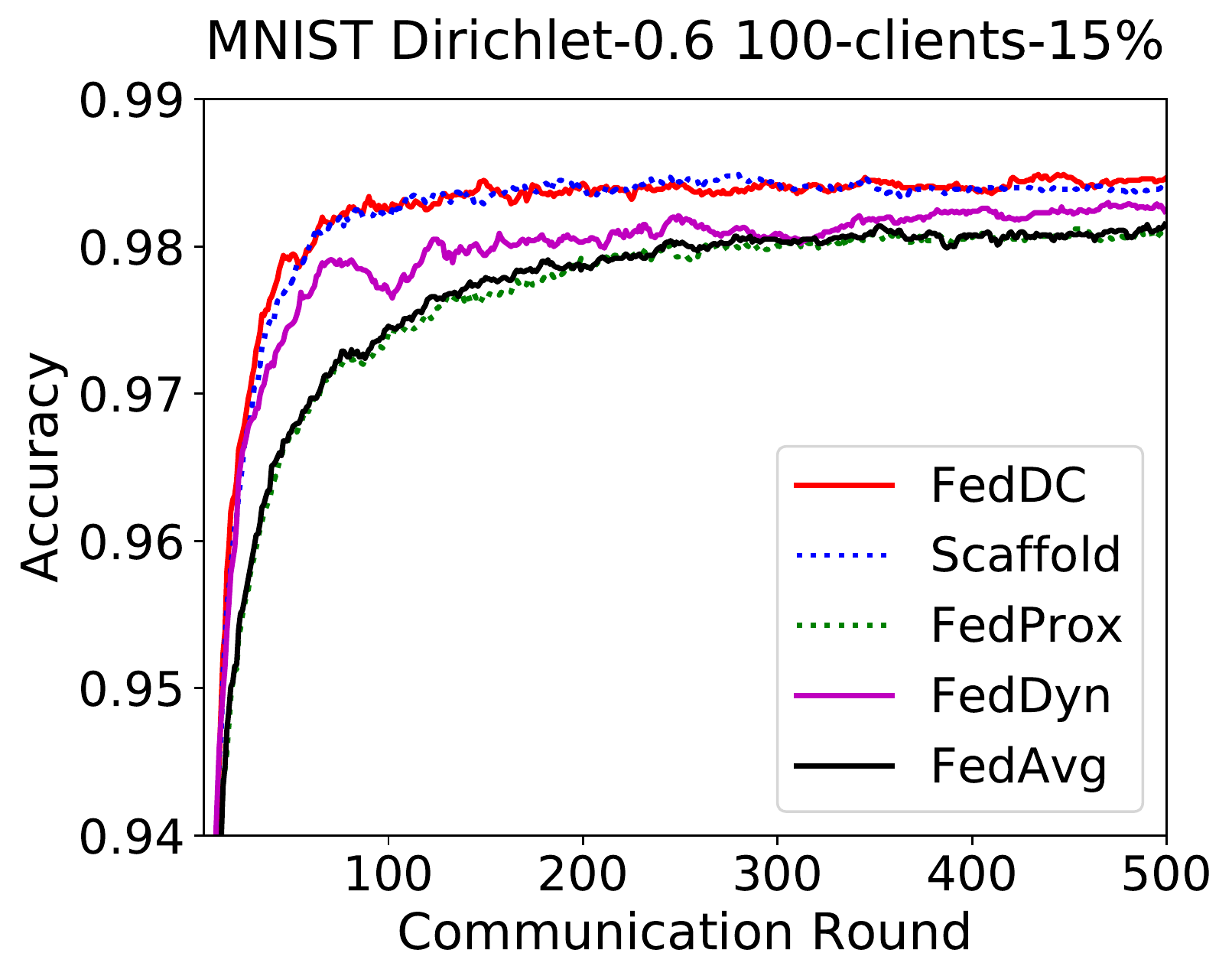}	\caption{}
	\end{subfigure}
	\\
	\begin{subfigure}{0.3\linewidth}
		\includegraphics[width=1.0\linewidth]{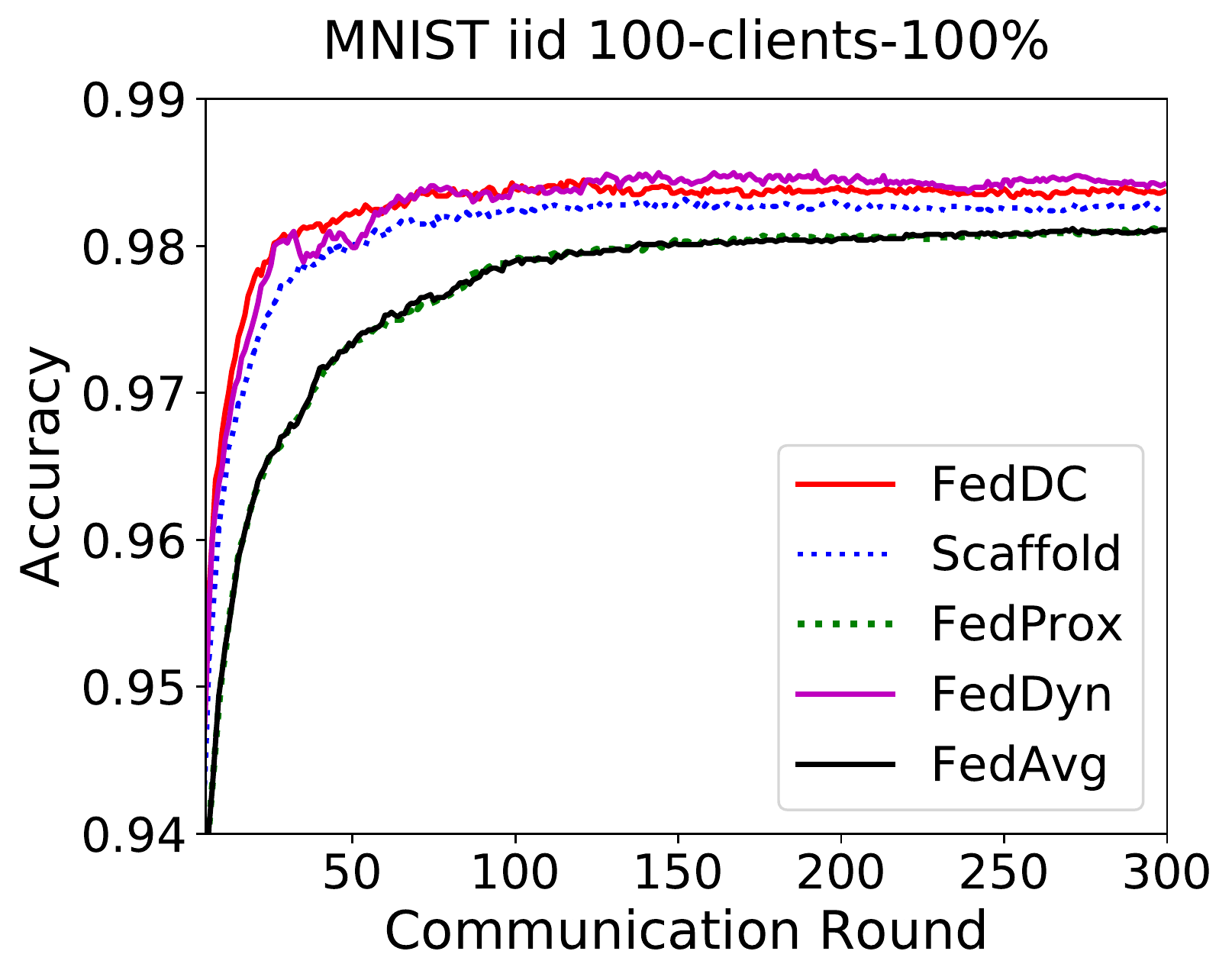}	\caption{}
	\end{subfigure}
	\begin{subfigure}{0.3\linewidth}
		\includegraphics[width=1.0\linewidth]{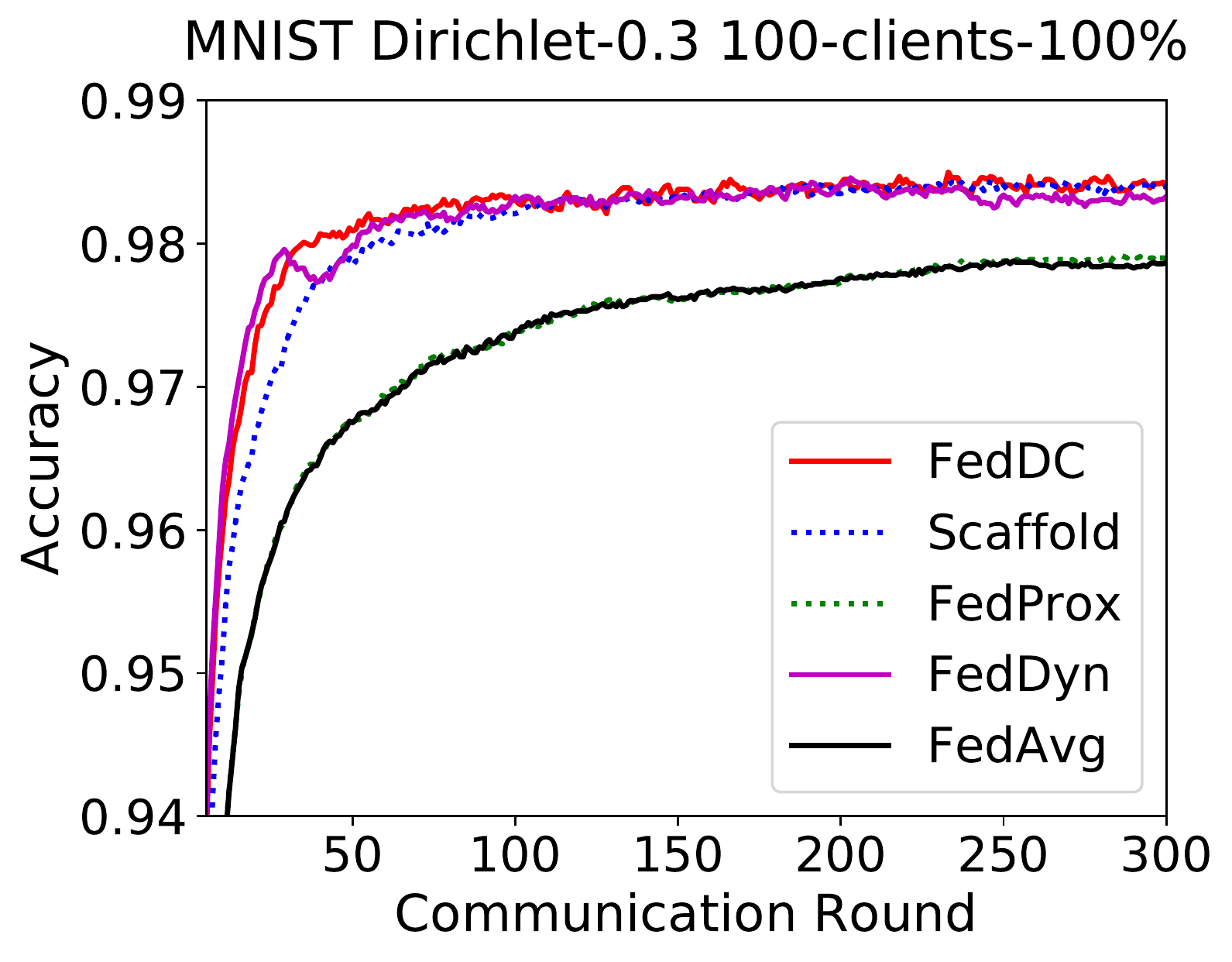}	\caption{}
	\end{subfigure}
	\begin{subfigure}{0.3\linewidth}
		\includegraphics[width=1.0\linewidth]{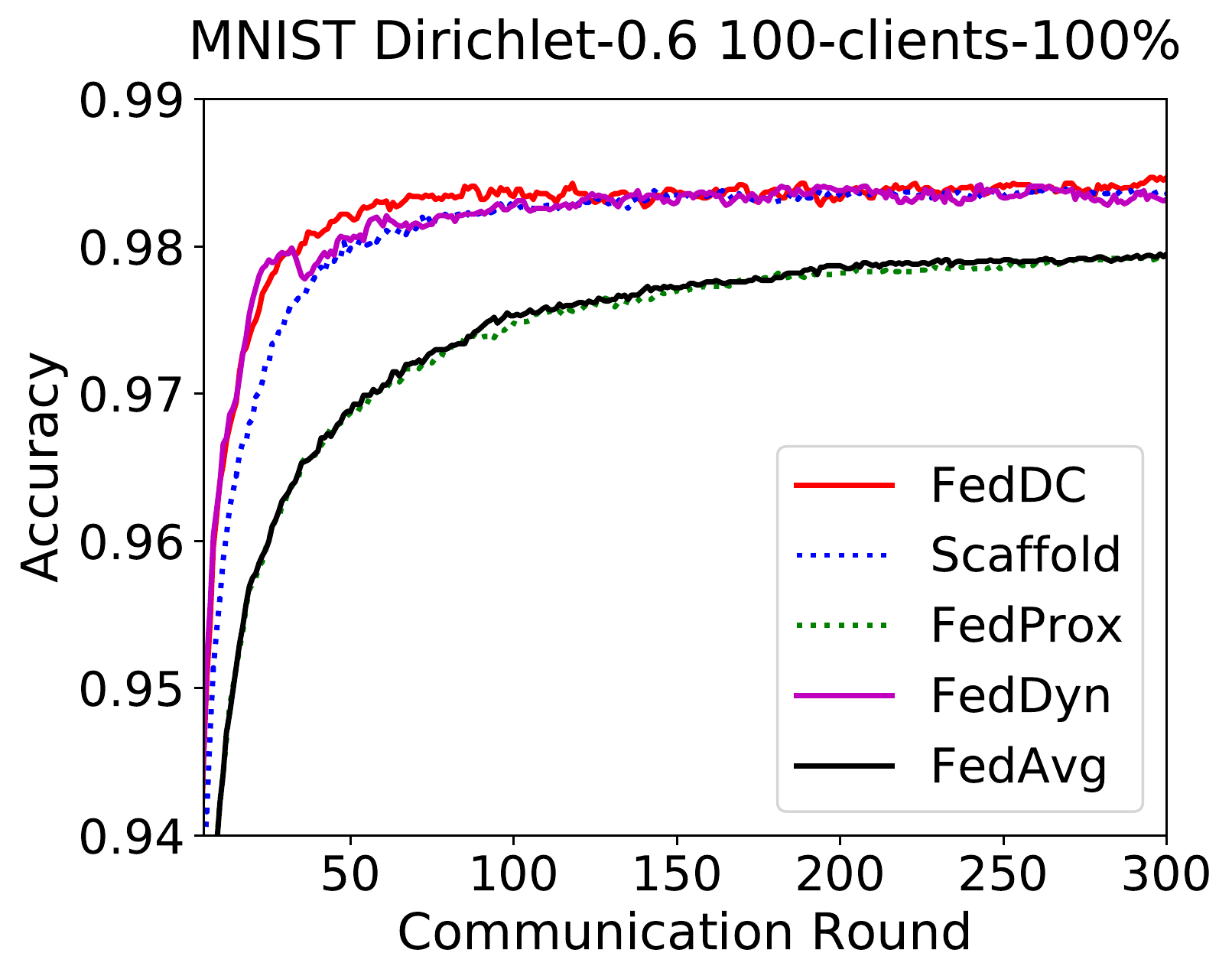}	\caption{}
	\end{subfigure}
	\caption{Convergence plots for iid and non-iid data with 100 clients adopting $100\%$ and $15\%$ client participating settings on MNIST.}\label{figure_mnist}
\end{figure}

\begin{figure}[!t]
	\centering

	\begin{subfigure}{0.3\linewidth}
		\includegraphics[width=1.0\linewidth]{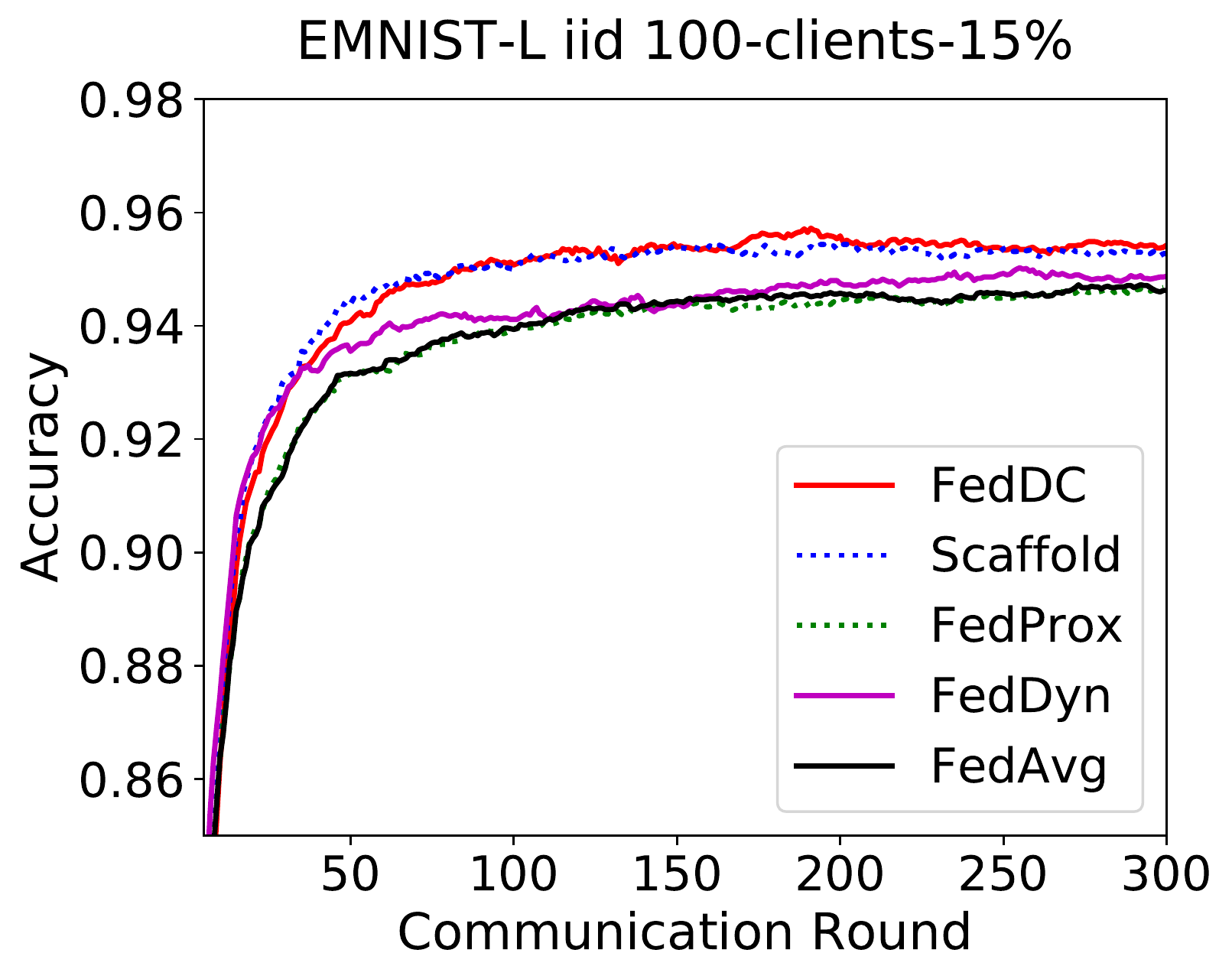}	\caption{}
	\end{subfigure}
	\begin{subfigure}{0.3\linewidth}
		\includegraphics[width=1.0\linewidth]{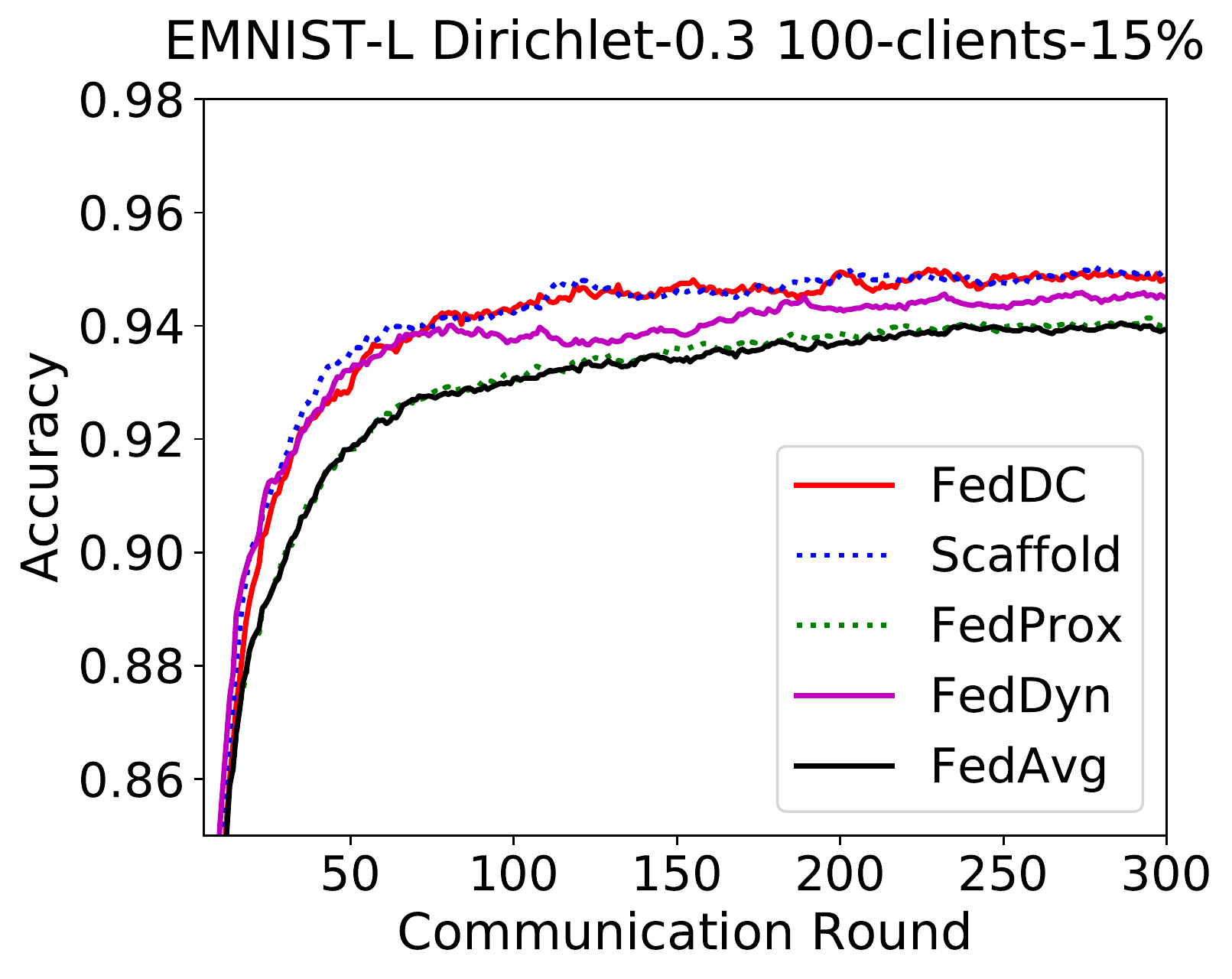}	\caption{}
	\end{subfigure}
	\begin{subfigure}{0.3\linewidth}
		\includegraphics[width=1.0\linewidth]{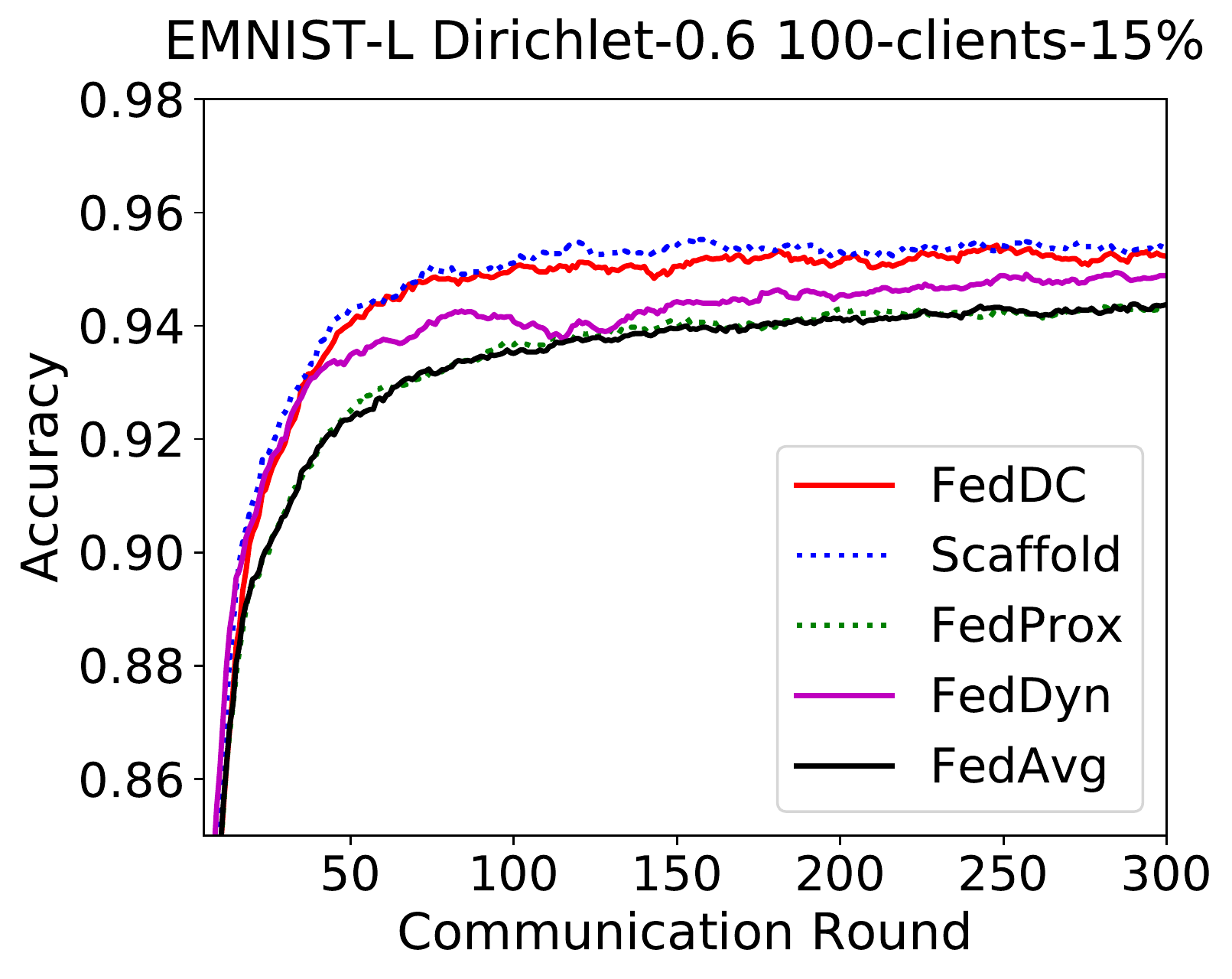}	\caption{}
	\end{subfigure}
	\\
	\begin{subfigure}{0.3\linewidth}
		\includegraphics[width=1.0\linewidth]{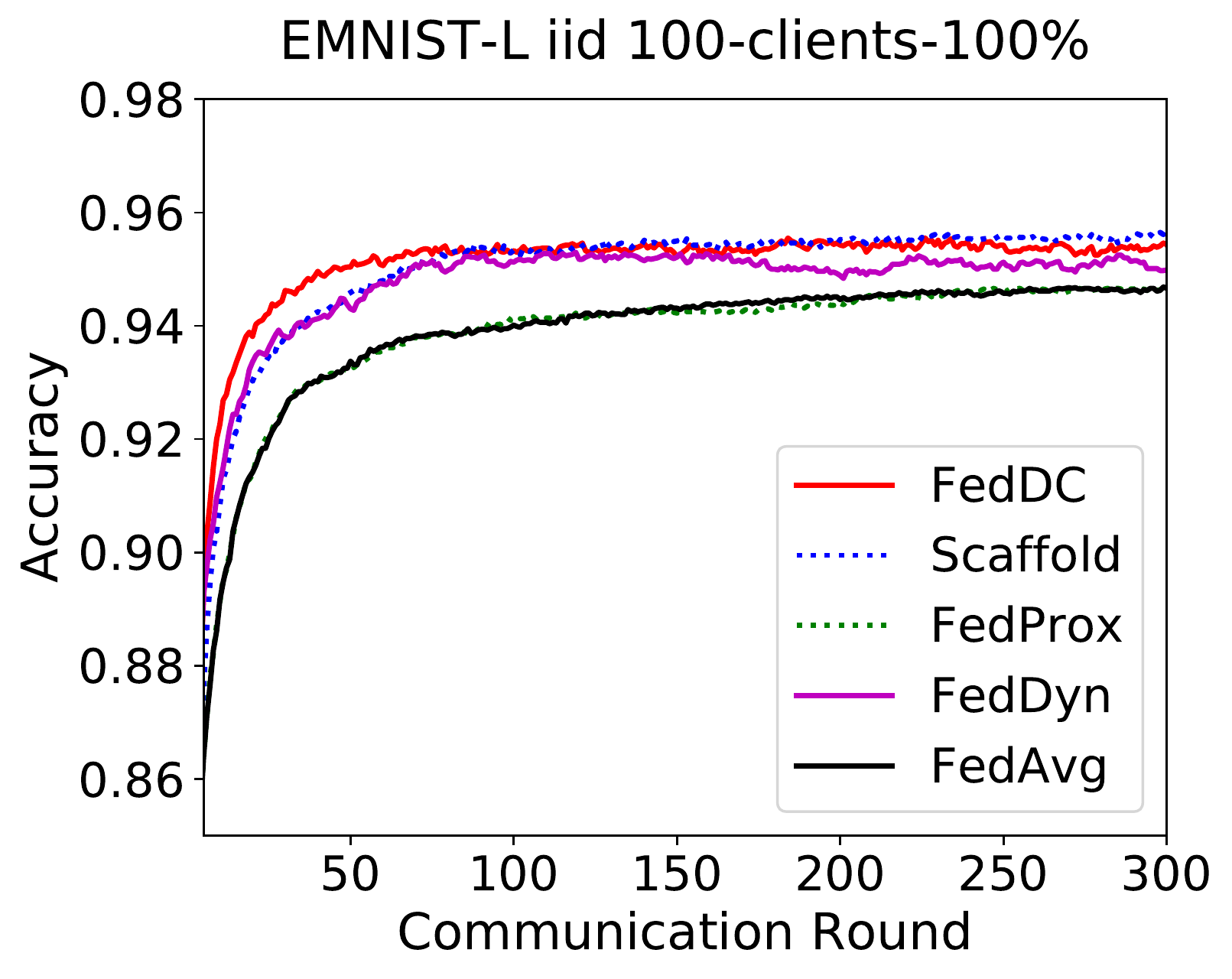}	\caption{}
	\end{subfigure}
	\begin{subfigure}{0.3\linewidth}
		\includegraphics[width=1.0\linewidth]{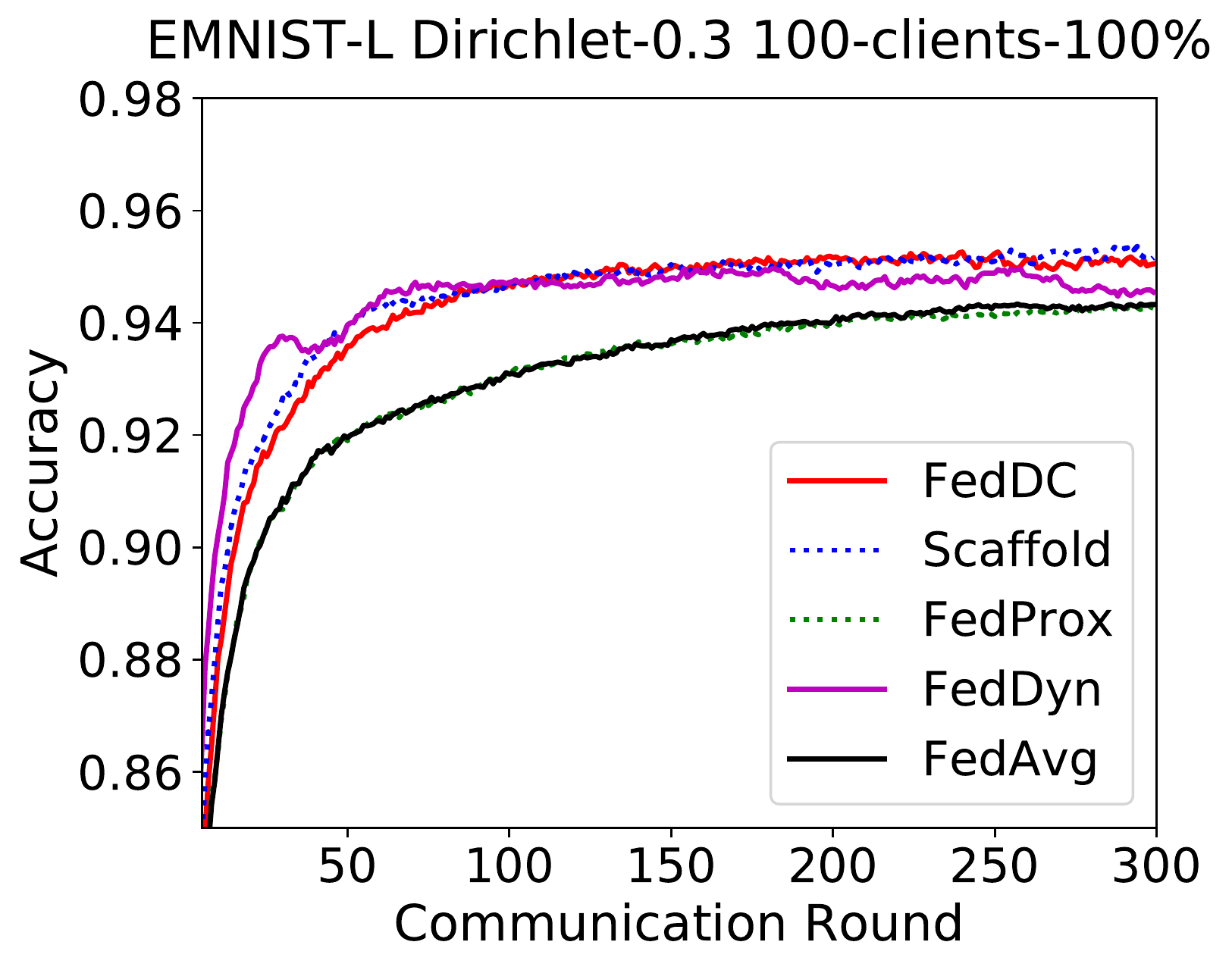}	\caption{}
	\end{subfigure}
	\begin{subfigure}{0.3\linewidth}
		\includegraphics[width=1.0\linewidth]{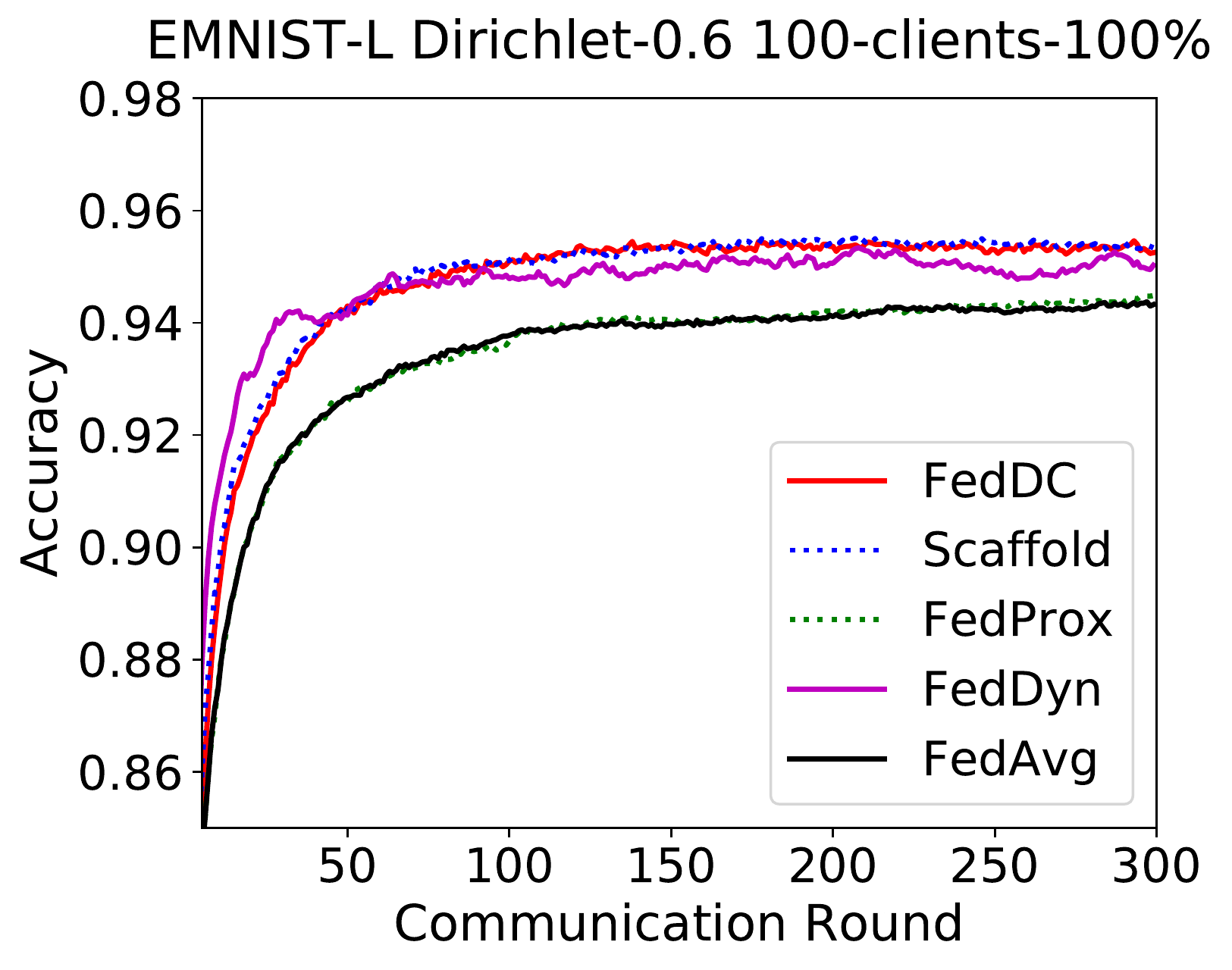}	\caption{}
	\end{subfigure}
	\caption{Convergence plots for iid and non-iid data with 100 clients adopting $100\%$ and $15\%$ client participating settings on EMNIST-L.}\label{figure_emnist}
\end{figure}

\textbf{Convergence plots.} 
We display a lot of convergence plots that trained in different settings and different datasets to verify the robustness and effectiveness of FedDC.
Figure \ref{fig_syn} shows the convergence plots on the Synthetic dataset with $20$ clients $15\%$ and $100\%$ client participating in three types data settings.
The results indicate that FedDC is consistently the best compared with the baselines on the Synthetic dataset.
Figure \ref{figure_client_number} shows the convergence curves with the massive devices which contains $500$ clients on MNIST, CIFAR10 and CIFAR100 datasets. The results show that FedDC is robust to the setting with large-scale clients.
Figure \ref{figure_unbalance} shows the convergence plots of models which are trained on the unbalanced datasets.
The results indicate that FedDC is robust on unbalanced settings, FedDC gets the best performance on the unbalanced CIFAR10 and CIFAR100 datasets in both settings of $15\%$ client participating and $100\%$ client participating.
Figure \ref{figure_fashion}, \ref{figure_cifr10} and \ref{figure_cifr100} show the convergence plots that trained with different client participating partitions and different data distributions on fashion MNIST, CIFAR10 and CIFAR100.
FedDC's accuracy outperforms the baselines, and FedDC also converges faster than baselines. 
Figure \ref{figure_mnist} and \ref{figure_emnist} are the convergence plots on MNIST and EMNIST datasets. In the experiments on MNIST and EMNIST, FedDC achieves competitive performance.

\textbf{The comparison of convergence speed.}
We compare the convergence speed of FedDC with baselines on CIFAR10, CIFAR100, MNIST and EMNIST-L datasets, the results are given in Table \ref{tab_communication_saving_cifar_100}, \ref{tab_communication_saving_cifar_15}, \ref{tab_communication_saving_mnist_100} and \ref{tab_communication_saving_mnist_15}. We report the communication speedup to achieve the target accuracy in which the benchmark is the number of communication rounds consumed by FedAvg. All results show that FedDC reaches the target accuracy using fewer communication rounds than other methods.
As a case, Table \ref{tab_communication_saving_cifar_100} reports the communication rounds of different methods (FedAvg, FedProx, Scaffold, FedDyn and FedDC) to achieve three accuracy degrees on CIFAR10 and CIFAR100 datasets with $100$ clients and $100\%$ client participating. 
In the experiments which train models in non-iid distributed data, we compare the FedDC's communication speedup of FedDC with FedAvg method. In the table, the number of communication round $(\textgreater{}1000)$ means the method can not achieve the target accuracy in $1000$ rounds, and the $SpeedUp$ with the symbol $(\textgreater{})$ means it is calculated with the benchmark of FedAvg where the communication rounds $(\textgreater{}1000)$.
The results indicate that FedDC outperforms all the comparison methods in both iid and non-iid ($0.6$-Dirichlet and $0.3$-Dirichlet distribution) settings. 
Tacking the settings with $100$ clients and $100\%$ client sampling ratio as an example, we can draw the following conclusions. In the experiments on CIFAR10 dataset, when FedAvg costs $149$ communication rounds to achieve the $78\%$ accuracy, FedDC only spends $35$ rounds in the same setting, where FedDC is $4.25\times$ faster than FedAvg. Besides, FedDyn and Scaffold are $3.47\times$ and $1.67\times$ faster than FedAvg, respectively.
In the non-iid setting, we increase the data heterogeneity over clients. In the setting of $0.6$-Dirichlet distributed data, FedDC is $4.77\times$ faster than FedAVG to reach $78\%$ accuracy on CIFAR10. When the data is more non-iid distributed ($0.3$-Dirichlet), the FedDC is over $18.86\times$ faster than FedAvg to reach $80\%$ accuracy on CIFAR10. 
The convergence curves of the FedAvg method is more stable, which indicates that the local optimization of the FedAvg is slower and causes less fluctuation in parameters.
All the above results demonstrate that FedDC outperforms the baselines in both convergence speed and model accuracy. 

\textbf{Comparison with other recent methods.} 
We added experiments to compare with the following three recent methods including FedAdam, FedYogi and FedAdagrad~\cite{DBLPafo1}. Table \ref{tab_three_other methods} shows the evaluation results. We can find that FedDC performs the best in different settings. They solve the difficulty of tune and exhibiting unfavorable convergence behavior with adaptive optimization methods. Specifically, they propose federated versions of adaptive optimizers, including ADAGRAD, ADAM, and YOGI to improve the convergence in the presence of heterogeneous data. These methods take effect in the parameter aggregation stage with gradient momentum update method, which reduces the negative effect of client drift and accelerates the convergence speed in a certain extent. However, they have no way to solve the parameter drift in the local training phase. FedDC decouples and learns client drift in the client training phase, and uses it to correct local parameters, which has better adaptability. FedDC takes effect in local training phase that is orthogonal to these improved aggregation methods (FedAdam, FedYogi, FedAdagrad etc.), and they can be used in combination.

	\begin{table}[th]
		\centering 
		\caption{Comparison of FedDC with FedAdam, FedYogi and FedAdagrad. There are 100 clients and $15\%$ of them randomly participate in training per round. The table shows the test accuracy on one iid and two non-iid  (D1, D2) settings of CIFAR100 dataset.}
		\label{tab_three_other methods}
		\vskip -10pt
		\begin{tabular}{llll}
			\toprule
			Method & iid & D1 & D2 \\\midrule
			FedAdam& 40.9\% & 41.5\% & 41.6\% \\
			FedYogi	& 42.3\% & 42.6\% & 42.7\% \\
			FedAdagrad& 42.5\% & 42.5\% & 42.1\% \\
			FedDC& 55.4\% & 54.7\% & 53.9\%\\\bottomrule
		\end{tabular}
	\end{table}

\subsection{Discussion}
\textbf{Impact of data heterogeneity.} We set various types of data heterogeneity settings including Dirichlet-0.3 distributed datasets, Dirichlet-0.6 distributed datasets and unbalanced datasets. 
From the results, we observe that the non-iid degree significantly impacts the model performance for federated learning. With a higher non-iid setting in Dirichlet-0.3 distributed datasets, both the accuracy and the convergence speed of the global model is lower than iid settings. That indicates that data heterogeneity makes the model convergence in federated learning more unstable and challenging. 
Fortunately, the proposed FedDC has an advantage over baselines in all non-iid settings. 
FedDC achieves the fastest convergence and the best accuracy compared with baselines, indicating FedDC is much robust to data heterogeneity.

\begin{figure}[!t]
	\centering

	\begin{subfigure}{0.3\linewidth}
		\includegraphics[width=1.0\linewidth]{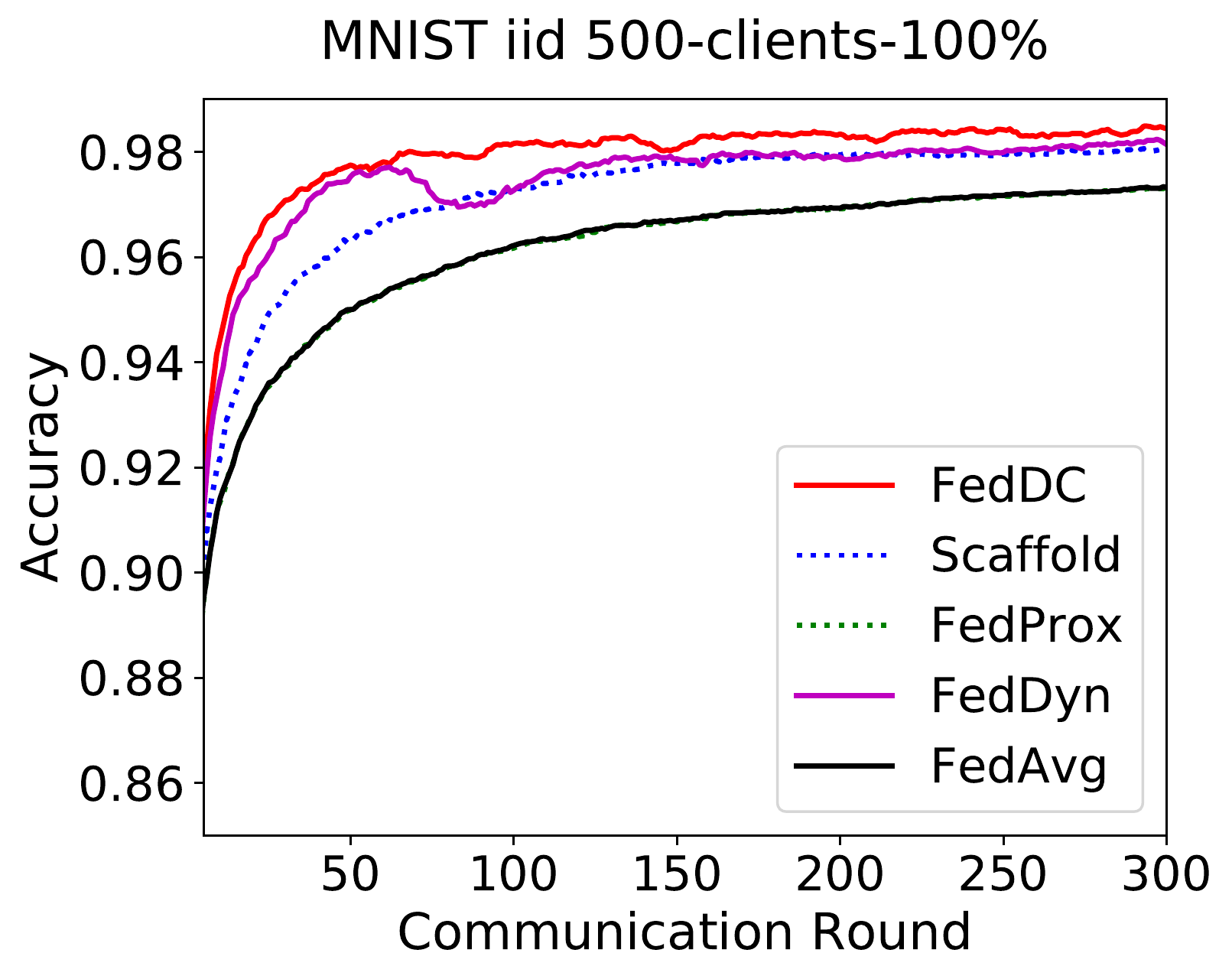}	\caption{}
	\end{subfigure}
	\begin{subfigure}{0.3\linewidth}
		\includegraphics[width=1.0\linewidth]{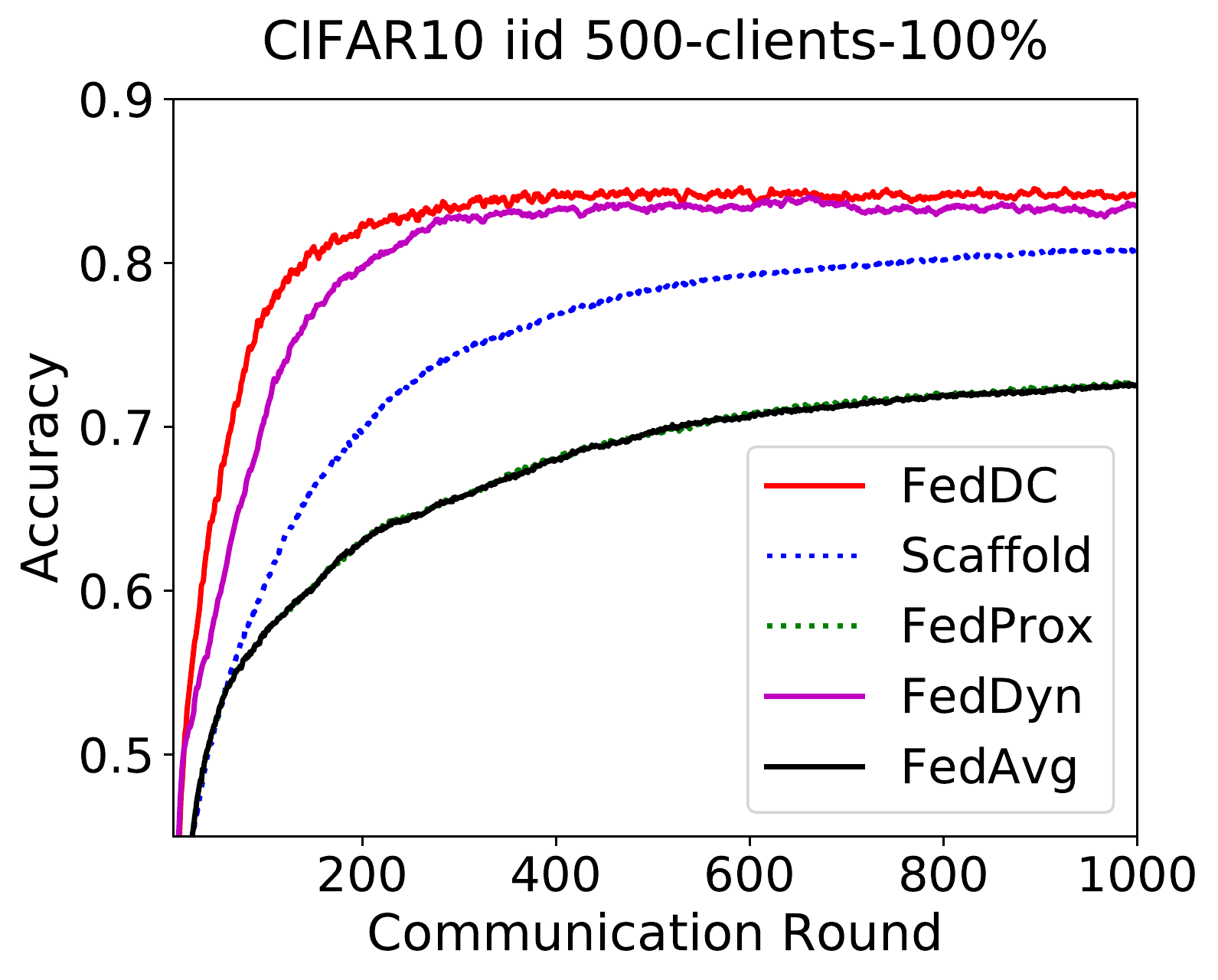}	\caption{}
	\end{subfigure}
	\begin{subfigure}{0.3\linewidth}
		\includegraphics[width=1.0\linewidth]{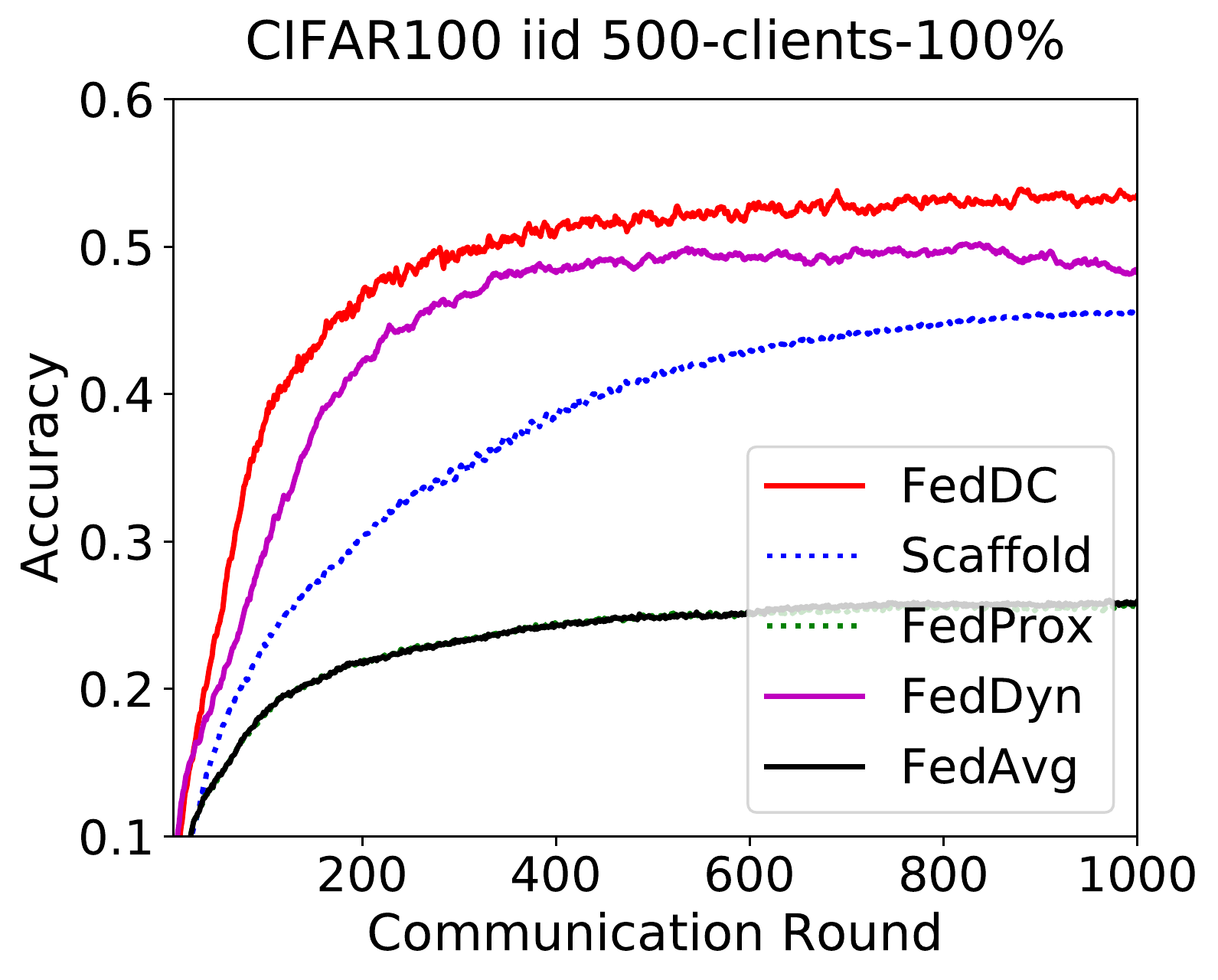}	\caption{}
	\end{subfigure}
	\caption{Convergence plots for massive clients ($500$) with $100\%$ client participating settings in the iid datasets of MNIST, CIFAR10 and CIFAR100.}\label{figure_client_number}
\end{figure}

\textbf{Impact of clients size and client sampling.} We first discuss the impact of different client settings for model convergence. 
Because part of the data cannot accurately describe the global data distribution in each round, part client participating introduce more randomness to the model than full client participating.
The total number of data points is fixed so that more clients means fewer samples in each client. The reduction in the amount of client local data would trigger more randomness in local optimization that makes it more challenging to track the parameter drift.
Figure \ref{figure_client_number} shows the results with massive clients.
In the massive clients setting, all methods are slower to reach a reasonable performance because each client owns fewer data. The FedDC spend $43$ rounds to reach $80\%$ accuracy in $100$ clients $100\%$ participating on iid CIFAR10, but that is $143$ rounds in the massive setting with $500$ clients. The results demonstrate that FedDC has a stronger ability to integrate information from massive clients to save communication and enhance model performance compared with baselines. FedDC brings communication-saving, which results in faster convergence than the baselines.
In the convergence plots of CIFAR10, CIFAR100, MNIST and EMNIST-L, we compare the model performance of full participating and part participating. 
All methods spend more communication rounds to achieve acceptable performance in the experiments with $15\%$ client participating than full participation.
FedDC outperforms FedAvg and FedProx a lot in test accuracy, and there are also satisfactory improvements of FedDC over Scaffold and FedDyn.  
Nevertheless, compared with other methods, FedDC improves the convergence speed and model accuracy significantly.
By the way, in the experiments with the different number of clients, we find it is beneficial to model convergence if the hyper-parameter $\alpha$ appropriately increased when the number of clients increases. That indicates that FedDC requires stronger constraints on the penalized terms in heterogeneous settings with bigger randomness. 

In summary, FedDC can better handle data heterogeneity, so that FedDC converges faster and obtains better model performance in the experiments. In addition, from the results with different numbers of customers, different data distributions, and different levels of client participation, we conclude that FedDC is strong robustness in various heterogeneous scenarios.

\begin{table}[H]\small
	\centering
	\caption{The number of communication round in different methods to achieve a target accuracy on CIFAR10 and CIFAR100 while containing with $100$ clients which $100\%$ participating each round. The $SpeedUp$ denotes the communication-saving relative to FedAvg. }
	\label{tab_communication_saving_cifar_100}
	\begin{tabular}{|c|c|cccccc|}
		\hline
		\multirow{2}{*}{Method} &
		\multirow{2}{*}{Accuracy} &
		\multicolumn{2}{c|}{Non-iid (0.6-Dirichlet)} &
		\multicolumn{2}{c|}{Non-iid (0.3-Dirichlet)} &
		\multicolumn{2}{c|}{iid} \\ \cline{3-8} 
		&
		&
		\multicolumn{1}{c|}{Round} &
		\multicolumn{1}{c|}{SpeedUp} &
		\multicolumn{1}{c|}{Round} &
		\multicolumn{1}{c|}{SpeedUp} &
		\multicolumn{1}{c|}{Round} &
		SpeedUp \\ \hline
		\multicolumn{8}{|c|}{CIFAR10 $100$ clients $100\%$ participating} \\ \hline
		\multirow{3}{*}{FedAvg}   & 0.78 & 205 & -       & 346 & -     & 149 & -    \\ \cline{2-2}
		& 0.8  & \textgreater{}1000 & -        & \textgreater{}1000 & -    & 286 & -     \\ \cline{2-2}
		& 0.82 & \textgreater{}1000 & -        & \textgreater{}1000 & -     & 803 & -     \\ \hline
		\multirow{3}{*}{FedProx}  & 0.78 &    195 & 1.05$\times$         & 350    &  0.99$\times$     &   142  &  1.05$\times$     \\ \cline{2-2}
		& 0.8  &  474   &  \textgreater{}2.11$\times$        & \textgreater{}1000    &      1$\times$ &  277   & 1.03$\times$      \\ \cline{2-2}
		& 0.82 &   \textgreater{}1000  &     1$\times$     &   \textgreater{}1000  &    1$\times$   & \textgreater{}1000    &    1$\times$   \\ \hline
		\multirow{3}{*}{Scaffold} &
		0.78$\times$ &		123 &		1.67$\times$ &		148 &		2.34$\times$ &		89 &
		1.67$\times$ \\ \cline{2-2}
		& 0.8  & 165 & \textgreater{}6.06$\times$ & 218 & \textgreater{}4.59$\times$  & 120 & 2.38$\times$  \\ \cline{2-2}
		& 0.82 & 283 & \textgreater{}1.71$\times$ & 387 & \textgreater{}2.58$\times$  & 194 & 4.14$\times$  \\ \hline
		\multirow{3}{*}{FedDyn}   & 0.78 & 44  & 4.66$\times$     & 57  & 6.07$\times$  & 43  & 3.47$\times$  \\ \cline{2-2}
		& 0.8  & 60  & \textgreater{}16.67$\times$    & 75  & \textgreater{}17.54$\times$ & 55  & 5.2$\times$   \\ \cline{2-2}
		& 0.82 & 84  & \textgreater{}11.90$\times$  & 114 & \textgreater{}8.77$\times$  & 75  & 10.7$\times$  \\ \hline
		\multirow{3}{*}{FedDC}     & 0.78 & 43  & 4.77$\times$     & 53  & 6.53$\times$  & 35  & 4.25$\times$  \\ \cline{2-2}
		& 0.8  & 53  & \textgreater{}18.86$\times$  & 70  & \textgreater{}14.28$\times$ & 43  & 6.65$\times$  \\ \cline{2-2}
		& 0.82 & 70  & \textgreater{}14.28$\times$  & 114 & \textgreater{}8.77$\times$  & 56  & 14.34$\times$ \\ \hline\hline
		\multicolumn{8}{|c|}{CIFAR100 $100$ clients $100\%$ participating}               \\ \hline
		\multirow{3}{*}{FedAvg}   & 0.35 & 142 & -        & 112 & -   & 201 & - \\ \cline{2-2}
		& 0.4  & 476 & -      & 847 & -    & \textgreater{}1000 & -    \\ \cline{2-2}
		&
		0.5 &
		\textgreater{}1000 & - &\textgreater{}1000 &	- &		\textgreater{}1000 &		-\\ \hline
		\multirow{3}{*}{FedProx}  &  0.35    & 190    &  0.75$\times$        &  124   &    0.9$\times$   &     145&  1.39$\times$    \\ \cline{2-2}
		&     0.4 & 502    &   0.95$\times$     &  507   &  1.67$\times$     &  273   & \textgreater{}3.66$\times$      \\ \cline{2-2}
		&     0.5 & \textgreater{}1000    &   1$\times$       &   \textgreater{}1000  &  1$\times$     & \textgreater{}1000    &   1$\times$    \\ \hline
		\multirow{3}{*}{Scaffold} & 0.35 & 64  & 2.22$\times$     & 67  & 1.67$\times$  & 58  & 3.47$\times$  \\ \cline{2-2}
		& 0.4  & 91  & 5.23$\times$     & 94  & 9.01$\times$  & 84  & \textgreater{}11.9$\times$  \\ \cline{2-2}
		& 0.5  & 424 & \textgreater{}2.35$\times$     & 501 & \textgreater{}2$\times$     & 305 & \textgreater{}3.28$\times$  \\ \hline
		\multirow{3}{*}{FedDyn}   & 0.35 & 38  & 3.74$\times$     & 38  & 2.95$\times$  & 45  & 4.47$\times$  \\ \cline{2-2}
		& 0.4  & 51  & 9.33$\times$     & 53  & 15.98$\times$ & 56  & \textgreater{}17.85$\times$ \\ \cline{2-2}
		& 0.5  & 154 & \textgreater{}6.49$\times$     & 182 & \textgreater{}5.95$\times$  & 169 & \textgreater{}5.92$\times$  \\ \hline
		\multirow{3}{*}{FedDC}     & 0.35 & 30  & 4.73$\times$     & 33  & 3.39$\times$  & 29  & 6.93$\times$  \\ \cline{2-2}
		& 0.4  & 39  &12.2$\times$     & 41  & 20.65$\times$ & 37  & \textgreater{}27.03$\times$ \\ \cline{2-2}
		& 0.5  & 70  & \textgreater{}14.28$\times$    & 81  & \textgreater{}12.35$\times$ & 70  & \textgreater{}14.28$\times$ \\ \hline
	\end{tabular}
\end{table}

\begin{table}[!t]\small
	\centering
	\caption{The number of communication round in different methods to achieve a target accuracy while containing with $100$ clients which $15\%$ participating each round. The $SpeedUp$ denotes the communication-saving relative to FedAvg.}
	\label{tab_communication_saving_cifar_15}
	\begin{tabular}{|c|c|cccccc|}
		\hline
		\multirow{2}{*}{Method} &
		\multirow{2}{*}{Accuracy} &
		\multicolumn{2}{c|}{Non-iid (0.6-Dirichlet)} &
		\multicolumn{2}{c|}{Non-iid (0.3-Dirichlet)} &
		\multicolumn{2}{c|}{iid} \\ \cline{3-8} 
		&
		&
		\multicolumn{1}{c|}{Round} &
		\multicolumn{1}{c|}{SpeedUp} &
		\multicolumn{1}{c|}{Round} &
		\multicolumn{1}{c|}{SpeedUp} &
		\multicolumn{1}{c|}{Round} &
		SpeedUp \\ \hline
		\multicolumn{8}{|c|}{CIFAR10 $100$ clients $15\%$ participating} \\ \hline
		\multirow{3}{*}{FedAvg}   & 0.78 & 259 & -       & 491 & -     & 177 & -     \\ \cline{2-2}
		& 0.8  &616 & -        & \textgreater{}1000 & -     & \textgreater{}1000 & -     \\ \cline{2-2}
		& 0.82 & \textgreater{}1000 & -        & \textgreater{}1000 & -    &\textgreater{}1000 & -     \\ \hline
		\multirow{3}{*}{FedProx}  & 0.78   &  228        &1.13$\times$     & 485      &  1.1$\times$   &153& 1.15$\times$      \\ \cline{2-2}
		& 0.8  &  459   & 1.34$\times$         &  \textgreater{}1000   &   1$\times$  &  307   &    \textgreater{}3.28   \\ \cline{2-2}
		& 0.82 &  \textgreater{}1000   &     1$\times$     &   \textgreater{}1000  &    1$\times$   &   \textgreater{}1000  &     1$\times$ \\ \hline
		\multirow{3}{*}{Scaffold} &
		0.78&
		132 &
		1.96$\times$ &
		169 &
		2.91$\times$ &
		94 &
		1.88$\times$ \\ \cline{2-2}
		& 0.8  & 200 & 3.08$\times$ & 263 & \textgreater{}3.80$\times$  & 126 & \textgreater{}7.93$\times$  \\ \cline{2-2}
		& 0.82 & 332 & \textgreater{}3.01$\times$ & 600 & \textgreater{}1.67$\times$  & 204 & \textgreater{}4.9$\times$  \\ \hline
		\multirow{3}{*}{FedDyn}   & 0.78 & 118  & 2.19$\times$     & 146  & 3.39$\times$  & 110  & 1.61$\times$  \\ \cline{2-2}
		& 0.8  & 193  & 3.19$\times$    & 195  & \textgreater{}5.12$\times$ & 145  & \textgreater{}6.9$\times$   \\ \cline{2-2}
		& 0.82 & 254 & \textgreater{}3.93$\times$  & 512 & \textgreater{}1.95$\times$  & 231  & \textgreater{}4.33$\times$  \\ \hline
		\multirow{3}{*}{FedDC}     & 0.78 & 101  & 2.56$\times$     & 105  & 4.68$\times$  & 88  & 2.01$\times$  \\ \cline{2-2}
		& 0.8  & 141  & 4.37$\times$  & 143  & \textgreater{}6.99$\times$ & 108  & \textgreater{}9.26$\times$  \\ \cline{2-2}
		& 0.82 & 211  & \textgreater{}4.74$\times$  & 242 & \textgreater{}4.13$\times$  & 162  & \textgreater{}6.17$\times$ \\ \hline\hline
		\multicolumn{8}{|c|}{CIFAR100 $100$ clients $15\%$ participating}               \\ \hline
		\multirow{3}{*}{FedAvg}   & 0.35 & 170 & -       & 144 & -    & 260 & -     \\ \cline{2-2}
		& 0.4  & 615 & -        & 520 & -     & 724 & -    \\ \cline{2-2}
		&
		0.5 &
		\textgreater{}1000 &- &\textgreater{}1000 &	- &		\textgreater{}1000 &		- \\ \hline
		\multirow{3}{*}{FedProx}  &  0.35    & 227    &  0.75$\times$        &   148  &  0.97$\times$     &  187   &   1.39$\times$    \\ \cline{2-2}
		&    0.4  &    980 &    0.63$\times$      &  503   &    1.03$\times$   & 650    &    1.11$\times$   \\ \cline{2-2}
		&     0.5 &   \textgreater{}1000  &    1$\times$      &  \textgreater{}1000   &    1$\times$   &   \textgreater{}1000  &     1$\times$  \\ \hline
		\multirow{3}{*}{Scaffold} & 0.35 & 68  & 2.5$\times$     & 72  & 2$\times$  & 68 & 3.82$\times$  \\ \cline{2-2}
		& 0.4  & 106  & 5.8$\times$     & 114  & 3.56$\times$  & 113  & 6.41$\times$  \\ \cline{2-2}
		& 0.5  & \textgreater{}1000 & 1$\times$     & \textgreater{}1000 & 1$\times$    & \textgreater{}1000 & 1$\times$ \\ \hline
		\multirow{3}{*}{FedDyn}   & 0.35 & 98  & 1.73$\times$     & 78  & 1.46$\times$  & 106  & 2.45$\times$  \\ \cline{2-2}
		& 0.4  & 149  & 4.42$\times$     & 148  & 3.51$\times$ & 143  & 5.06$\times$ \\ \cline{2-2}
		& 0.5  & 574 & \textgreater{}1.74$\times$     & 710 & \textgreater{}1.41$\times$  & 619 & \textgreater{}1.62$\times$  \\ \hline
		\multirow{3}{*}{FedDC}     & 0.35 & 78  & 2.18$\times$     & 74  & 1.54$\times$  & 74  & 3.51$\times$  \\ \cline{2-2}
		& 0.4  & 102  & 6.03$\times$     & 103  & 5.05$\times$ & 100  & 7.04$\times$ \\ \cline{2-2}
		& 0.5  & 249  & \textgreater{}4.02$\times$    & 278  & \textgreater{}3.6$\times$ & 206  & \textgreater{}4.85$\times$ \\ \hline
	\end{tabular}
\end{table}

\begin{table}[!t]\small
	\centering
	\caption{The number of communication round in different methods to achieve a target accuracy on MNIST and EMNIST-L while containing with $100$ clients which $100\%$ participating each round. The $SpeedUp$ denotes the communication-saving relative to FedAvg. }
	\label{tab_communication_saving_mnist_100}
	\begin{tabular}{|c|c|cccccc|}
		\hline
		\multirow{2}{*}{Method} &
		\multirow{2}{*}{Accuracy} &
		\multicolumn{2}{c|}{Non-iid (0.6-Dirichlet)} &
		\multicolumn{2}{c|}{Non-iid (0.3-Dirichlet)} &
		\multicolumn{2}{c|}{iid} \\ \cline{3-8} 
		&
		&
		\multicolumn{1}{c|}{Round} &
		\multicolumn{1}{c|}{SpeedUp} &
		\multicolumn{1}{c|}{Round} &
		\multicolumn{1}{c|}{SpeedUp} &
		\multicolumn{1}{c|}{Round} &
		SpeedUp \\ \hline
		\multicolumn{8}{|c|}{MNIST $100$ clients $100\%$ participating} \\ \hline
		\multirow{2}{*}{FedAvg}   & 0.96        & 25                &-      & 28                &-       & 16                & -           \\ \cline{2-2}
		& 0.98  & 258               & -     & 492               &-          & 142               & -            \\ \hline
		\multirow{2}{*}{FedProx}   & 0.96      & 24                & 1.04$\times$       & 27                & 1.04$\times$& 16                & 1$\times$   \\ \cline{2-2}
		& 0.98  & 263               & 0.98$\times$  & 480               & 1.03$\times$        & 136               & 1.04$\times$ \\ \hline
		\multirow{2}{*}{Scaffold} 
		& 0.96   & 11                & 2.27$\times$ & 14                & 2$\times$            & 9                 & 1.78$\times$\\ \cline{2-2}
		& 0.98  & 58                & 4.45$\times$ & 58                & 8.48$\times$  & 53                & 2.68$\times$\\ \hline
		\multirow{2}{*}{FedDyn}  & 0.96& 8                 & 3.13$\times$   & 9                 & 3.11$\times$  & 7                 & 2.29$\times$   \\ \cline{2-2}
		& 0.98  & 46                & 5.61$\times$& 51                & 9.65$\times$  & 27                & 5.26$\times$\\ \hline
		\multirow{2}{*}{FedDC}     & 0.96 & 8                 & 3.13$\times$ & 10                & 2.8$\times$         & 7                 & 2.29$\times$ \\ \cline{2-2}
		& 0.98   & 35                & 7.37$\times$& 37                & 13.3$\times$ & 26                & 5.46$\times$\\ \hline\hline
		\multicolumn{8}{|c|}{EMNIST-L $100$ clients $100\%$ participating}               \\ \hline
		\multirow{2}{*}{FedAvg}    & 0.94  & 142               &-  & 192               & -          & 107               & -            \\ \cline{2-2}
		& 0.95 & \textgreater{}300 & -           & \textgreater{}300 & -           & \textgreater{}300 &- \\ \hline
		\multirow{2}{*}{FedProx}  & 0.94& 135               & 1.05$\times$  & 198               & 0.97$\times$    & 92                & 1.16$\times$ \\ \cline{2-2}
		& 0.95 & \textgreater{}300 & 1$\times$           & \textgreater{}300 & 1$\times$           & \textgreater{}300 & 1$\times$      \\ \hline
		\multirow{3}{*}{Scaffold}  & 0.94 & 43                & 3.30$\times$& 52                & 3.69$\times$  & 30                & 3.57$\times$ \\ \cline{2-2}
		& 0.95  & 75                & \textgreater{}4$\times$    & 150               & \textgreater{}2$\times$              & 66                &\textgreater{} 4.55$\times$   \\ \hline
		\multirow{2}{*}{FedDyn}    & 0.94 & 30                & 4.73$\times$  & 52                & 3.69$\times$  & 27                & 3.96$\times$\\ \cline{2-2}
		& 0.95  & 137               & \textgreater{}2.19$\times$& 160               & \textgreater{}1.88$\times$          & 69                & \textgreater{}4.35$\times$\\ \hline
		\multirow{2}{*}{FedDC}      & 0.94  & 43                & 3.3$\times$ & 60                & 3.2$\times$          & 21                & 5.1$\times$ \\ \cline{2-2}
		& 0.95      & 78                & \textgreater{}3.85$\times$    & 134               & \textgreater{}2.24$\times$   & 50                & \textgreater{}6$\times$  \\ \hline
	\end{tabular}
\end{table}

\begin{table}[!t]\small
	\centering
	\caption{The number of communication round in different methods to achieve a target accuracy on MNIST and EMNIST-L while containing with $100$ clients which $15\%$ participating each round. The $SpeedUp$ denotes the communication-saving relative to FedAvg.  }
	\label{tab_communication_saving_mnist_15}
	\begin{tabular}{|c|c|cccccc|}
		\hline
		\multirow{2}{*}{Method} &
		\multirow{2}{*}{Accuracy} &
		\multicolumn{2}{c|}{Non-iid (0.6-Dirichlet)} &
		\multicolumn{2}{c|}{Non-iid (0.3-Dirichlet)} &
		\multicolumn{2}{c|}{iid} \\ \cline{3-8} 
		&
		&
		\multicolumn{1}{c|}{Round} &
		\multicolumn{1}{c|}{SpeedUp} &
		\multicolumn{1}{c|}{Round} &
		\multicolumn{1}{c|}{SpeedUp} &
		\multicolumn{1}{c|}{Round} &
		SpeedUp \\ \hline
		\multicolumn{8}{|c|}{MNIST $100$ clients $15\%$ participating} \\ \hline
		\multirow{2}{*}{FedAvg}  & 0.96      & 32                & -      & 35                & -         & 23                & -     \\ \cline{2-2}
		& 0.98     & 361               & -     & \textgreater{}600 & -         & 158               & -              \\ \hline
		\multirow{2}{*}{FedProx}  & 0.96      & 31                & 1.03$\times$     & 34                & 1.03$\times$  & 23                & 1$\times$ \\ \cline{2-2}
		& 0.98 & 383               & 0.94$\times$ & 418               & \textgreater{}1.44$\times$  & 149               & 1.06$\times$\\ \hline
		\multirow{2}{*}{Scaffold} 
		& 0.96    & 20                & 1.6$\times$  & 23                & 1.52$\times$     & 16                & 1.44$\times$   \\ \cline{2-2}
		& 0.98    & 62                & 5.82$\times$     & 72                &\textgreater{} 8.33$\times$  & 50                & 3.16$\times$  \\ \hline
		\multirow{2}{*}{FedDyn}   & 0.96 & 21                & 1.52$\times$ & 23                & 1.52$\times$    & 18                & 1.28$\times$ \\ \cline{2-2}
		& 0.98  & 122               & 2.96$\times$ & 153               & \textgreater{}3.92$\times$ & 71                & 2.23$\times$ \\ \hline
		\multirow{2}{*}{FedDC}      & 0.96   & 18                & 1.78$\times$     & 22                & 1.59$\times$  & 16                & 1.44$\times$ \\ \cline{2-2}
		& 0.98  & 60      & 6.02$\times$ & 62                &\textgreater{} 9.68$\times$           & 46                & 3.43$\times$ \\ \hline\hline
		\multicolumn{8}{|c|}{EMNIST-L $100$ clients $15\%$ participating}               \\ \hline
		\multirow{2}{*}{FedAvg}    & 0.94   & 153               & -       & 245               &-           & 108               & -          \\ \cline{2-2}
		& 0.95 & \textgreater{}300 & -           & \textgreater{}300 & -           & \textgreater{}300 & -     \\ \hline
		\multirow{2}{*}{FedProx}  & 0.94  & 145               & 1.06$\times$  & 240               & 1.02$\times$   & 105               & 1.03$\times$ \\ \cline{2-2}
		& 0.95 & \textgreater{}300 & 1$\times$           & \textgreater{}300 & 1$\times$           & \textgreater{}300 & 1$\times$        \\ \hline
		\multirow{3}{*}{Scaffold} & 0.94   & 44                & 3.48$\times$ & 68                & 3.6$\times$& 42                & 2.57$\times$\\ \cline{2-2}
		& 0.95 & 95                & \textgreater{}4.21$\times$  & \textgreater{}300 & 1$\times$           & 87                & \textgreater{}3.45$\times$  \\ \hline
		\multirow{2}{*}{FedDyn}     & 0.94& 73                & 2.1$\times$  & 81                & 3.06$\times$        & 61                & 1.61$\times$  \\ \cline{2-2}
		& 0.95    & 127               & \textgreater{}2.36$\times$ & \textgreater{}300 & 1$\times$          & 255               & \textgreater{}1.18$\times$\\ \hline
		\multirow{2}{*}{FedDC}      & 0.94  & 48                & 3.19$\times$  & 74                & 3.31$\times$ & 47                & 2.3$\times$    \\ \cline{2-2}
		& 0.95 & 92                & \textgreater{}3.26$\times$ & \textgreater{}300 & 1$\times$           & 81                & \textgreater{}3.7$\times$ \\ \hline
	\end{tabular}
\end{table}

\clearpage
\section{Appendix: Convergence Proof of FedDC}\label{appendix_proof}
First we present algorithm of the proposed FedDC in Algorithm \ref{algorithmFedDC}. In each training round $t$, the server first selects the active client set $C_t$ (where $C_t \subseteq [N]$) and boardcasts the global model parameters to them. Then each active client updates the local model parameter and the corresponding local drift variables on its own local datasets. Finally, the server aggregates the sum of the local models and local drift variables to update the global model. In the algorithm, $F$ is the objective loss function, $g_i = \Delta\theta_i$ and $g= \Delta \theta$ are auxiliary variables used in the third term (the gradient correction item $G_i(\theta_i;g_i,g)$) of the objective function $F$.

\begin{algorithm}[htbp]
	\small
	\caption{Algorithm of FedDC }	\label{algorithmFedDC}
	\KwIn{Random initial global model parameter $w$, set training round $T$, the number of clients $N$, initial local drift variables as all zero matrix, the learning rate $\eta$, the number of local training batches $K$.}
	\KwOut{The trained global model $w$.}
	\For{ $t = 1,2,...,T$}{
		Sample the active client set $C_t\subseteq [N]$.\\ 
		\For{ each client $i \in{C_t}$ in parallel}{
			Set the local model parameter $\theta_i=w,$\\
			\For{ $k = 1,2,...,K$}{ 
				\textbf{Update} the local model parameter: \\$\quad\quad \theta_i=\theta_i+\eta\frac{\partial F(\theta_i;h_i,D_i,w)}{\partial\theta_i}$,
			}
			Set local gradient drift $\Delta\theta_i=\theta_i-w,$\\
			\textbf{Update} the local drift: $ h_i = h_i + \Delta\theta_i,$
		}
		\textbf{Update} the global model: $w=\frac{1}{|C_t|}\sum\limits_{i\in C_t}(\theta_i+h_i),$\\ 
		Set global gradient drift $\Delta \theta=\frac{1}{|C_t|}\sum\limits_{i\in C_t}\Delta\theta_i$,\\
	}
	\textbf{Return} $w$.
\end{algorithm}

We newly define some symbols to facilitate the convergence analysis of FedDC step by step. Specifically, we use the superscript $t$ to represent the communication round, and the subscript $i$ to represent the client index. For example, $\theta_i^t$ represents the local model parameter of client $i$ in the $t$-th round. Considering only the variables to be optimized, the objective function $F$ can be rewritten as:
\begin{gather}
F_i(\theta_i)= \mathbb{E}_{(x,y)\in D_i}l(\theta_i,(x,y)) +\frac{\alpha}{2} ||\theta_i-(w^{t-1}-{h}_i^{t-1})||^2   \notag\\
+ \frac{1}{K\eta} \langle\theta_i,\Delta\theta_i^{t-1}-\Delta\theta^{t-1}\rangle,
\end{gather}
where $\Delta \theta_i^{t-1}=\Delta g_i^{t-1}$ and $\Delta\theta^{t-1}=\mathbb{E} \Delta g_i^{t-1}$,
In the $k$-th local training iteration of the $t$-th communication round, the client first optimizes local model with the gradient of local objective function which is represented as follows:
\begin{equation}\label{local_update}
\theta_i^{t,k}=\theta_i^{t,k-1}-\eta\nabla_{\theta}F_i(\theta_i^{t,k-1}).
\end{equation}
The update value of the corresponding local drift variable is represented as $\Delta h_i^t =-\eta\sum_{k=1}^K \nabla_\theta F_i(\theta_i^{t,k},h_i^{t,k})$.
After the local training process completes, the server updates the global parameters based on the updated $\theta_i^{t}$ and $h_i^{t}$ with the following approach:
\begin{equation}\label{update_global}
w^t=\mathbb{E}_{i\in [C_t]}(\theta_i^{t}+h_i^{t})=\mathbb{E}_{i\in [C_t]}(\theta_i^{t-1}+h_i^{t-1}+\Delta \theta_i^t+\Delta h_i^t ),
\end{equation}
where $\theta_i^t$ and $h_i^t$ are abbreviations for $\theta_i^{t,K}$ and $h_i^{t,K}$, respectively.

\subsection{Discussion of FedDC.} 
We first present the intuition and results of FedDC convergence analysis.
The parameters of clients' local model indirectly align with global parameters by adding the local drift variables. In federated learning, suppose the clients' local optimal points $\theta_i^*,\forall i\in[N]$ are arbitrarily different from each other due to the heterogeneous training data located on various clients. We show that in FedDC, when each client reaches the local optimal points, and the global model also converges to a stationary point at the same time.
Based on the Eq.~(\ref{update_global}), if the local model converges to local optima, then $\Delta \theta_i^t\rightarrow 0, \forall i\in[N]$, that implies  $w^{t}=w^{t-1}+\Delta w^{t+1}=w^{t-1}+2\Delta \theta^t=w^{t-1}$. That indicates the global model also converges when clients' local models all converge to their local stationary points.
 
\textbf{Penalized Term.}
 The penalized term is mainly used to help the local drift variables to track the parameter gap. Due to the data heterogeneity among clients, it is impractical to assume that all local models converge to a consistent stationary point. In FedDC, the parameter drift specifically tracks the gap between local models and global models.
 The parameter deviation from local models to the global model is caused by the following two factors: 1) the update drift in the current round, and 2) the residual parameter deviation. 
 FedProx~\cite{li2020federated} and Scaffold~\cite{karimireddy2021scaffold} have proved that reducing the update drift is effective for speeding up the convergence time. However, the residual parameter drift has a cumulative effect between communication rounds, making it more critical to convergence and performance of the training process.
 We use auxiliary local drift variables to denote the parameter deviation of the client's local model in federated learning from the unbiased global model. In this way, we decouple the training of the global model from the clients' local models. Each client updates its local drift variables under the limitation of the penalized term $\frac{\alpha}{2}||\theta_i-(w-h_i)||^2$ which ensure the effectiveness of the local drift variables. 
 
 \textbf{Gradient correction.}
 To briefly and clearly illustrate the effectiveness of the gradient correction term, we first disregard the drift variable and the penalized term. We assume that all clients are active. Under this condition, we suppose $L_i(\theta_i^t)=\mathbb{E}_{(x,y)\in D_i}l(\theta_i,(x,y))$ and $L(\theta)=\mathbb{E}_{(x,y)\in D}l(\theta,(x,y))$. Client $i$'s corrected gradient which uses the gradient correction term satisfies $g_i^{t,k}=\nabla_\theta L_i(\theta_i^{t,k})+\frac{1}{K\eta}(\Delta \theta_i^{t-1}-\Delta\theta^{t-1})\approx \nabla_\theta L_i(\theta_i^{t,k}) +(\nabla_\theta L(\theta_i^{t-1,k})-\nabla_\theta L_i(\theta_i^{t-1,k}))$ to optimize its model instead of $\nabla_\theta L_i(\theta_i^{t,k})$.
 The gradient variance is $\frac{1}{N}\sum_{i=1}^N ||g_i^{t,k}- g^{t,k}||^2$, where $g^{t,k}=\frac{1}{N}\sum_{i=1}^N g_i^{t,k} \approx \nabla F(\theta_i^t)$. Thus, the different degrees of local gradient can be expressed as 
 \begin{equation}
 \small
 \begin{split}
 \frac{1}{N}\sum_{i=1}^N ||g_i^{t,k}- g^{t,k}||^2 & \approx \frac{1}{N} \sum_{i=1}^N ||\nabla_\theta L_i(\theta_i^{t,k}) +(\nabla_\theta L(\theta_i^{t-1,k})-\nabla_\theta L_i(\theta_i^{t-1,k}))-\nabla_\theta L(\theta_i^{t,k})||^2 \\
 &\leq \frac{2}{N} \sum_{i=1}^N [||\nabla_\theta L_i(\theta_i^{t,k}) -\nabla_\theta L_i(\theta_i^{t-1,k})||^2+||\nabla_\theta L(\theta_i^{t-1,k})-\nabla_\theta L(\theta_i^{t,k})||^2].
 \label{lammaf15}
 \end{split}
 \end{equation}
 The variance of the local gradients is bounded by the above inequality, which is independent of the dissimilarity of their local objective functions. With the smoothness assumption of $L$ and $L_i, (\forall i\in [N])$, their gradients would not change a lot. We deduce that the gradient of each client is strictly bounded. Thus, the gradient correction term is effective to reduce the gradient drift.

 \textbf{Convergence results.}
 We show the convergence theoretical analysis for FedDC in convex and non-convex functions. With the local drift variable to correct the local parameter, we denote $L(\cdot)$ as the global empirical loss objective and we have $L(w)=F(w)$.
 We suppose the objective function is $\beta$-Lipschitz continuous gradient and $B$-local dissimilarity bounded under the $\gamma$-inexact solution assumption \cite{MLSYS2020_38af8613}, that implies the following expected objective decent in each round:
 \begin{equation}{\label{lammacon1}}
 \begin{split}
 &\mathbb{E}_{C_t} L(w^{t}) \leq L(w^{t-1})-(\frac{2-2\gamma B}{\alpha}-\frac{2\beta B(1+\gamma)}{\alpha\bar\alpha}-2\beta\frac{B^2(1+\gamma)^2}{\hat\alpha^2})||\nabla L(w^{t-1})||^2,
 \end{split}
 \end{equation}
 where $\bar\alpha=\alpha-\beta_d>0$ is a constant, $C_t$ is the selected active client set in round $t$. We can use the objective function decrease to note the convergence on convex and non-convex $L(\cdot)$.
 In non-convex case, assuming $\Gamma=L(w^0)-L(w^*)$, $p=(\frac{1}{\alpha}-(\frac{\gamma }{\alpha}-\frac{(1+\gamma)\sqrt{2}}{\bar\alpha \sqrt{C}}-
 \frac{\beta (1+\gamma)}{\alpha\bar\alpha})\sqrt{1+\frac{\sigma^2}{\epsilon}}
 -(\frac{\beta(1+\gamma)^2}{2\bar\alpha^2}-\frac{\beta (1+\gamma)^2(2*\sqrt{2}+2)}{\bar\alpha^2 C}))(1+\frac{\sigma^2}{\epsilon})>0$, the relation in Eq.~(\ref{lammacon1}) holds for FedDC, we get 
 $\mathbb{E}_{C_t} L(w^t)\leq L(w^{t-1})-2p||\nabla L(w^{t-1})||^2$. Giving a $\epsilon>0$, we prove that $\sum_{t=1}^T||\nabla L(w^t)||^2\leq \epsilon$ when the number of communication round satisfies $T=O(\frac{\Gamma}{p\epsilon})$. If the objective functions are convex, setting $\beta_d=0,\bar\alpha=\alpha$, if $\gamma=0$, $B\leq \sqrt{C}$ and $1<<B \leq 0.5\sqrt{C}$, we have 
 \begin{equation}{\label{lammacon2}} 
 \mathbb{E}_{C_t} L(w^t)\leq L(w^{t-1})-\frac{2}{\alpha^2}[
 \alpha(1-\frac{\sqrt{2}B}{\sqrt{C}})-(B+(\frac{(2\sqrt{2}+2)}{C}+\frac{1}{2})\beta B^2)]||\nabla L(w^{t-1})||^2. 
 \end{equation}
 Let $\alpha = 6\beta B^2$, to achieve the convergence state where $\sum_{t=1}^T||\nabla L(w^t)||^2\leq \epsilon$, FedDC costs $T=O(\frac{\beta B^2\Gamma}{\epsilon})$ communication rounds. 
 The detailed proof is given in following.

\subsection{Detailed convergence Proof of FedDC.} 

\textbf{A1: $\beta$-smoothness function.} $f$ is $\beta$-smoothness that satisfies 
\begin{equation}{\label{lammappa1}}
||\nabla f(\theta_1)-\nabla f(\theta_2)||\leq \beta ||\theta_1-\theta_2||, \quad for \quad \forall \theta_1,\theta_2,
\end{equation}  
that also implies a quadratic upper bound for $f$,
\begin{equation}{\label{lammappa1-1}}
f(\theta_2)\leq f(\theta_1)+\langle  \nabla f(\theta_1),\theta_2-\theta_1\rangle+\frac{\beta}{2}||\theta_2-\theta_1||^2.
\end{equation}  

\textbf{A2: $\mu$-convex function.} $f$ is a $\mu$-convex function for $\mu>0$ that satisfies 
\begin{equation}{\label{lammappa2}}
\langle\nabla f(\theta_1),\theta_2-\theta_1\rangle \leq f(\theta_2)-f(\theta_1)-\frac{\mu}{2} ||\theta_1-\theta_2||^2, \quad for \quad \forall \theta_1,\theta_2.
\end{equation}  

\textbf{D1: $B$-local dissimilarity bounded.}
If the local empirical loss $L_i$ is $B$-local dissimilarity where $\mathbb{E}||\nabla L_i(\theta)||^2\leq ||\nabla L(w)||^2B^2$, and we define $B(\theta)=\sqrt{\frac{\mathbb{E}||\nabla L_i(\theta)||^2}{||\nabla L(w)||^2}}$.

\textbf{A3: $\gamma$-inexact solution.}
We define function $F_i(\theta_i,\hat{\theta_i})$ as $F_i(\theta_i,\hat{\theta_i})=L_i(\theta_i)+\frac{\alpha}{2}||\theta_i-\hat\theta_i||^2, \quad where \quad \alpha\in[0,1] \quad and \quad \hat\theta_i=w-h_i $,
we get the gradient of $F_i$: $\nabla F_i(\theta_i,\hat{\theta_i}) = \nabla L_i(\theta_i) + \alpha (\theta_i-\hat\theta_i)$.
If $\theta_i^*$ is a $\gamma$-inexact point of $\min F_i(\theta_i,\hat\theta_i)$, it satisfies
$||\nabla F_i(\theta_i^*,\hat{\theta_i})||
\leq \gamma||\nabla F_i(\hat{\theta_i},\hat{\theta_i})||$.

\textbf{A4. Bounded dissimilarity assumption for $L$.} There exists a $B_\epsilon$ while $\epsilon>0$, for any $w$, that satisfies $||\nabla L(w)||^2>\epsilon$, and $B(w)>B_\epsilon$.

Our convergence proof for FedDC use a similar method as that in FedProx \cite{li2020federated}. 
In FedDC, the parameter of global model consists of the average of local model parameters and the average of local drift variables.
\begin{equation}{\label{lammapp2}}
w=\mathbb{E}_i(\theta_i+h_i)=\mathbb{E}_i\theta_i+\mathbb{E}_i h_i
\end{equation}
We define a virtual variable $\hat\theta_i^t$ as the corrected local parameter in $t$-th round, that satisfies
\begin{equation}{\label{lammapp3}}
\hat\theta_i^t=w^t-h_i^t.
\end{equation}
We get $\mathbb{E}\hat\theta_i^t=w^t-\mathbb{E}h_i^t=\mathbb{E}_i\theta_i^t$ from the definition of \ref{lammapp2} and \ref{lammapp3}, where $\hat\theta_i^t$ is independent with the active client set $C_t$. 
In FedDC, we define $ L_i=\mathbb{E}_{(x,y)\in D_i}l(\theta_i,(x,y)) +\frac{1}{K\eta} \langle\theta_i,\Delta\theta_i^{t-1}-\Delta\theta^{t-1}\rangle$, thus, the local objective function of $i$-th client is
\begin{equation}{\label{lammapp4}}
F_i(\theta_i)=L_i(\theta_i)+\frac{\alpha}{2} ||\theta_i-(w^{t-1}-h_i^{t-1})||^2=L_i(\theta_i)+\frac{\alpha}{2}||\theta_i-\hat\theta_i^{t-1}||^2.
\end{equation}
The gradient of $F_i$ in round ${t+1}$ is
\begin{equation}{\label{lammapp5}}
\nabla F_i(\theta_i^{t+1})=\nabla L_i(\theta_i^{t+1})+ \alpha(\theta_i^{t+1}-\hat\theta_i^t).
\end{equation}
In addition, let $L(\theta)=\mathbb{E}_{(x,y)\in D}l(\theta_i,(x,y))$, from the denifition of $L_i$, we have $L(\theta)=\mathbb{E}_{(x,y)\in D}l(\theta_i,(x,y))=\mathbb{E}L_i(\theta)$.
We define $\bar\theta^t=\mathbb{E}\theta_i^t=\mathbb{E}\hat\theta_i^t$, the expectation of Eq. \ref{lammapp5} satisfies
\begin{equation}{\label{lammapp6}}
\mathbb{E}_i\nabla F_i(\theta_i^{t+1})=\mathbb{E}_i\nabla L_i(\theta_i^{t+1})+ \alpha \mathbb{E}_i(\theta_i^{t+1}-\hat\theta_i^t)=\mathbb{E}_i\nabla L_i(\theta_i^{t+1})+ \alpha \mathbb{E} (\bar\theta^{t+1}-\bar\theta^t),
\end{equation}
then we get 
\begin{equation}{\label{lammapp7}}
\bar\theta^{t+1}-\bar\theta^t=\frac{1}{\alpha}(\mathbb{E}_i\nabla F_i(\theta_i^{t+1})-\mathbb{E}_i\nabla L_i(\theta_i^{t+1})).
\end{equation}
In addition, form the process of FedDC, we get
\begin{equation}{\label{lammapp8}}
h_i^{t+1}=h_i^t+\theta_i^{t+1}-\theta_i^t \quad \rightarrow   \quad h_i^{t+1}-h_i^t=\theta_i^{t+1}-\theta_i^t,
\end{equation}
so that the  difference of the global parameters in $t+1$-th round and $t$-th round is  
\begin{equation}{\label{lammapp9}}
w^{t+1}-w^{t}=\mathbb{E}[ (h_i^{t+1}-h_i^t)+(\theta_i^{t+1}-\theta_i^t)]=2(\bar\theta^{t+1}-\bar\theta^t).
\end{equation}

\textbf{Theorem 1: Convergence of FedDC in non-convex case.} For non-convex and $\beta$-Lipschitz smooth function $L_i,\forall i\in[N]$, there exists a $\beta_d>0$, where $\bar\alpha=\alpha-\beta_d>0$ and $\nabla^2 L_i\geq -\beta_dI$. We assume the local empirical loss $L_i$ is non-convex and $B$-dissimilarity, in which $B(\theta^t)\leq B$. The global objective of FedDC decreases as follows: 
\begin{equation}{\label{lammap}}
\begin{split}
&\mathbb{E}_{C_t} L(w^t)\leq L(w^{t-1})-2p||\nabla L(w^{t-1})||^2,
\end{split}
\end{equation}
where $p=(\frac{\gamma}{\alpha}-\frac{B(1+\gamma)\sqrt{2}}{\bar\alpha \sqrt{N}}-\frac{\beta B(1+\gamma)}{\alpha\bar\alpha}-\frac{\beta(1+\gamma)^2B^2}{2\bar\alpha^2}-\frac{\beta B^2(1+\gamma)^2(2\sqrt{2C}+2)}{\bar\alpha^2 N})>0$, $C_t$ is the active client set in round $t$ which contains $C$ clients.

\textbf{Proof for Theorem 1.}
In the proof, we follow the techniques of \cite{li2020federated}, assume the local empirical loss $L_i$ is $\gamma$-inexactness solver.
We define $e_i^{t}$ as
\begin{equation}{\label{lammapp13}}
\nabla L_i(\theta_i^t) + \alpha (\theta_i^t-\hat\theta_i^{t-1}) -e_i^t=0.
\end{equation}
In addition, we have $\nabla F_i(\hat\theta_i^{t-1},\hat\theta_i^{t-1}) = \nabla L_i(\hat\theta_i^{t-1}) $, so with the $B$-local dissimilarity bounded assumption we can get:
$||\nabla F_i(\theta_i^t,\hat\theta_i^{t-1})||\leq ||\nabla F_i(\theta_i^*,\hat\theta_i^{t-1})||\leq \gamma||\nabla F_i(\hat\theta_i^{t-1},\hat\theta_i^{t-1})||$, that implies
\begin{equation}{\label{lammapp14}}
||e_i^t||	\leq	\gamma||\nabla L_i(\hat\theta_i^{t-1})||.
\end{equation}

As $\bar\theta^t=\mathbb{E}_i{\theta_i^t}=\mathbb{E}_i{\hat\theta_i^t}$, so that we get the following equation
\begin{equation}{\label{lammapp15}}
\bar\theta^t-\bar\theta^{t-1}=\mathbb{E}_i[\theta_i^t-\hat\theta_i^{t-1}]=\frac{1}{\alpha}\mathbb{E}_i(-\nabla L_i(\theta_i^{t})+e_i^{t}).
\end{equation}
Let $\bar\alpha=\alpha-L_d>0$ and $\ddot\theta_i^{t}=\arg\min_\theta F_i(\theta,\hat\theta_i^{t-1})$. Due to that $F_i$ is $\bar\alpha$ strong convex function, we get
\begin{equation}{\label{lammapp16}}
||\ddot\theta_i^t-\theta_i^t|| \leq \frac{\gamma}{\bar\alpha}||\nabla L_i(\hat\theta_i^{t-1})||.
\end{equation}
With the strong convex nature of $F_i$ again, we get
\begin{equation}{\label{lammap16-1}}
||\ddot\theta_i^t-\hat\theta_i^{t-1}|| \leq \frac{1}{\bar\alpha}||\nabla L_i(\hat\theta_i^{t-1})||.
\end{equation}
Using triangle inequality for \ref{lammapp16} and \ref{lammap16-1}, we get:
\begin{equation}{\label{lammapp17}}
||\theta_i^t-\hat\theta_i^{t-1}|| \leq \frac{1+\gamma}{\bar\alpha}||\nabla L_i(\hat\theta_i^{t-1})||.
\end{equation}
With the bounded dissimilarity assumption and $\mathbb{E}\hat\theta_i^{t-1}=\mathbb{E}\hat\theta^{t-1}=\bar\theta_i^{t-1}$, we get
\begin{equation}{\label{lammapp18}}
\begin{split}
&||\bar\theta^t-\bar\theta^{t-1}||
\leq \mathbb{E}_i||\theta_i^t-\hat\theta_i^{t-1}|| 
\leq \frac{1+\gamma}{\bar\alpha}\mathbb{E}_i||\nabla L_i(\hat\theta_i^{t-1})|| \\
&\leq \frac{1+\gamma}{\bar\alpha}\sqrt{\mathbb{E}_i||\nabla L_i(\hat\theta_i^{t-1})||^2}
\leq B\frac{1+\gamma}{\bar\alpha} ||\nabla L(\bar w^{t-1})||,
\end{split}
\end{equation}
where the last inequality is due to the bounded dissimilarity assumption and $F(w)=L(w)$, $\nabla \nabla_w F(w)=\nabla_\theta L_i(\theta_i)$.

We define $M_{t}$ as $\bar\theta^t-\hat\theta^{t-1}=-\frac{1}{\alpha}(\nabla L(w^{t-1})+M_t))$. Taking Eq. \ref{lammapp15} into it, we get $M_t=\mathbb{E}_i[(\nabla L_i(\theta_i^{t})-\nabla L_i (\hat\theta_i^{t-1})-e_i^t]$. $M_t$ is bounded with
\begin{equation}{\label{lammapp19}}
\begin{split}
||M_t||
&\leq \mathbb{E}_i[(\beta ||\theta^{t}-\hat\theta^{t-1}||+||e_i^t||]
\leq (\frac{\beta(1+\gamma)}{\bar \alpha}+\gamma)\mathbb{E}_i||\nabla L_i(\hat\theta^{t-1})||\\
&\leq (\frac{\beta(1+\gamma)}{\bar \alpha}+\gamma)B||\nabla L(w^{t-1}) ||,
\end{split}
\end{equation}
The last is due to $\nabla_w L(w)=\nabla_w F(w)=\nabla_\theta L_i(\theta)$ and the bounded dissimilarity assumption. 

Because $h_i^t=h_i^{t-1}+\Delta \theta_i^t$, we get $\mathbb{E}(\theta_i^t-\theta_i^{t-1})=\mathbb{E}(h_i^t-h_i^{t-1})=\mathbb{E}(\bar\theta^t-\ddot\theta^{t-1})$, and $w^t-w^{t-1}=\mathbb{E}(\theta_i^t+h_i^t)-\mathbb{E}(\theta_i^{t-1}+h_i^{t-1})=2\mathbb{E}(\theta_i^t-\theta_i^{t-1})$.

With $\beta$-Lipschitz smoothness assumption of $L$ and Taylor expansion, we get
\begin{equation}{\label{lammapp20}}
\begin{split}
&L(w^{t})
\leq L(w^{t-1})+<\nabla L(w^{t-1}), w^t-w^{t-1}>+\frac{\beta}{2}|| w^t-w^{t-1}||^2 \\
&\leq_1 L(w^{t-1})+<\nabla L(w^{t-1}), 2\mathbb{E}_i(\theta_i^t-\theta_i^{t-1})>+\frac{\beta}{2}||2\mathbb{E}(\theta_i^t-\theta_i^{t-1})||^2\\
&\leq_2 L(w^{t-1})-\frac{2}{\alpha}||\nabla L(w^{t-1})||^2-\frac{2}{\alpha}<\nabla L(w^{t-1}),M_t >+\frac{2\beta B^2(1+\gamma)^2}{\hat\alpha^2}||\nabla L(w^{t-1})||^2\\
&\leq L(w^{t-1})-(\frac{2-2\gamma B}{\alpha}-\frac{2\beta B(1+\gamma)}{\alpha\bar\alpha}-2\beta\frac{B^2(1+\gamma)^2}{\hat\alpha^2})||\nabla L(w^{t-1})||^2,
\end{split}
\end{equation}
where $(\leq_1)$ is due to $w^t-w^{t-1}=2\mathbb{E}(\theta_i^t-\theta_i^{t-1})$, $(\leq_2)$ is due to the definition of $M$.
Set a proper $\alpha$ for the above inequality, $L(w^t)-L(w^{t-1})$ is decrease proportional to $||\nabla L(w^{t-1})||^2$. The above inequality demonstrates that if the hyper-parameter $\alpha$ of the penalized term is large enough, the works would be decreased.

\textbf{Proof for partial client participation settings.}
In practice, FedDC runs on sampled active clients each round. We assume there are $C$ clients are chosen randomly to the active set $C_t$ in round $t$.
With a local Lipschitz continuity assumption for $L$, if $\beta_l$ is the continuity constant, we get
\begin{equation}{\label{lammapp21}}
L(w_2)\leq L(w_1)+\beta_l||w_1-w_2||, \quad for \quad \forall w_1,w_2,
\end{equation}
besides, we assume $\theta^t=\mathbb{E}_{C_t} \theta_i^t$ and $\bar\theta^t=\mathbb{E}_i \theta_i^t$, the following satisfies that
\begin{equation}{\label{lammap15}}
\begin{split}
\beta_l &\leq ||\nabla L(w^{t-1})||+\beta \max(||\bar w^t-w^{t-1}||,||w^t-w^{t-1}||)\\
&\leq ||\nabla L(w^{t-1})||+\beta (||\bar w^t-w^{t-1}||+||w^t-w^{t-1}||).
\end{split}
\end{equation}
So that in the partial client participating settings we need to bound
\begin{equation}{\label{lammappp23}}
\mathbb{E}_{C_t}L(w^t)\leq L(\bar w^t)+\mathbb{E}_{C_t}\beta_l||w^t-\bar w^t||
\leq L(\bar w^t)+\mathbb{E}_{C_t}\beta_l||w^t-\bar w^t||,
\end{equation}
where the expectation is calculated on the active client set $C_t$.
\begin{equation}{\label{lammapp24}}
\begin{split}
&\mathbb{E}_{C_t}\beta_l||w^t-\bar w^t|| 
\leq \mathbb{E}_{C_t}[||\nabla L(w^{t-1})||+\beta (||\bar w^t-w^{t-1}||+||w^t-w^{t-1}||)]*||w^t-\bar w^t|| \\
&\leq[||\nabla L(w^{t-1})||+\beta ||\bar w^t-w^{t-1}||]*\mathbb{E}_{C_t}||w^t-\bar w^t||+\beta\mathbb{E}_{C_t}||w^t-\bar w^t||*||w^t-w^{t-1}|| \\
&\leq (||\nabla L(w^{t-1})||+2\beta (||\bar w^t-w^{t-1}||)\mathbb{E}_{C_t}||w^t-\bar w^t||
+\mathbb{E}_{C_t}||w^t-\bar w^t||^2.
\end{split}
\end{equation}

Taking $\mathbb{E}(\theta_i^t-\theta_i^{t-1})
\leq B\frac{1+\gamma}{\bar\alpha} ||\nabla L(\hat\theta^{t-1})||$ from \ref{lammapp19}, we have
\begin{equation}{\label{lammapp25}}
\mathbb{E}_{C_t}||\theta^t-\bar\theta^{t}||\leq \sqrt{\mathbb{E}_{C_t}||\theta^t-\bar\theta^{t}||^2},
\end{equation}
and
\begin{equation}{\label{lammapp26}}
\begin{split}
&\mathbb{E}_{C_t}||\theta^t-\bar\theta^{t}||^2 \leq\frac{1}{C}\mathbb{E}_{i\in C_t}(||\theta_i^t-\bar\theta^{t}||^2)\\
&\leq \frac{2}{C}\mathbb{E}_{i\in C_t}||\theta_i^t-\theta^{t-1}||^2\\
&\leq \frac{2}{C}\frac{(1+\gamma)^2}{\bar\alpha^2}\mathbb{E}_{i\in C_t}||\nabla L_i(\theta^{t-1})||^2, 
\end{split}
\end{equation}
where the last inequality is due to bounded dissimilarity assumption. Further, with $h_i$ fixed, we get $\mathbb{E}_{C_t}||w^t-\bar w^{t}||^2=\mathbb{E}_{C_t}||\theta^t-\bar\theta^{t}||^2\leq \frac{2}{C}\frac{(1+\gamma)^2}{\bar\alpha^2}\mathbb{E}_{i\in C_t}||\nabla L_i(\theta^{t-1})||^2 \leq \frac{2B^2(1+\gamma)^2}{C\bar\alpha^2}||\nabla L(w^{t-1})||^2$. Replace the bound in \ref{lammapp24}, the inequality becomes
\begin{equation}{\label{lammapp27}}
\mathbb{E}_{C_t}\beta_l||w^t-\bar w^t|| \leq (\frac{B\sqrt{2}(1+\gamma)}{\bar\alpha \sqrt{C}}+\frac{\beta B^2(1+\gamma)^2}{\bar\alpha^2 C}(2*\sqrt{2}+2))||\nabla L(w^{t-1})||^2.
\end{equation}

We combine \ref{lammapp20},\ref{lammappp23},\ref{lammapp27} to get 
\begin{equation}{\label{lammap21}}
\begin{split}
&\mathbb{E}_{C_t} L(w^t)\leq L(w^{t-1})-2(\frac{1-\gamma B}{\alpha}-\frac{B(1+\gamma)\sqrt{2}}{\bar\alpha \sqrt{C}}-\\
&\frac{\beta B(1+\gamma)}{\alpha\bar\alpha}-\frac{\beta(1+\gamma)^2B^2}{2\bar\alpha^2}-\frac{\beta B^2(1+\gamma)^2(2*\sqrt{2}+2)}{\bar\alpha^2 C})||\nabla L(w^{t-1})||^2.
\end{split}
\end{equation}

\subsection{Bounded Gradients}
We get prove the following corollary. We hold the bounded dissimilarity assumption for any $L_i$. The bounded variance of gradients is
\begin{equation}{\label{lammapp28}}
\mathbb{E}||\nabla L_i(\theta)-\nabla L(w)||^2\leq \sigma^2, \quad \forall \epsilon>0,
\end{equation}
Then we get $B_\epsilon\leq \sqrt{1+\frac{\sigma^2}{\epsilon}}$.
We can restate the convergence result in Theorem 1 based on this Corollary and the the bounded variance assumption.

\textbf{Proof for bounded Gradients.}
We get the following inequalities,
\begin{equation}{\label{lammapp29}}
\begin{split}
&\mathbb{E}||\nabla L_i(\theta)-\nabla L(w)||^2=\mathbb{E}||\nabla L_i(\theta)||-||\nabla L(w)||^2
\leq \mathbb{E}||\nabla L_i(\theta)-\nabla L(w)||^2\leq \sigma^2\\
&\mathbb{E}||\nabla L_i(\theta)||^2\leq \sigma^2+||\nabla L(w)||^2\\
&B_\epsilon=(\frac{(\mathbb{E}_i ||\nabla L_i(\theta)||^2)}{||\nabla L(w)||^2})^\frac{1}{2}\leq(1+\frac{\sigma^2}{\epsilon})^{\frac{1}{2}}.
\end{split}
\end{equation}

\subsection{Convergence of FedDC in non-convex case}
\textbf{Assumption:B-local dissimilarity}
If $L_i$ is non-convex, $\beta$-Lipschitz smooth function, and $B$-local dissimilarity bounded. There existing $\beta_d$ makes $\nabla^2 L_i\geq -\beta_d-I$ and $\bar\alpha=\alpha-\beta_d>0$. $B(\theta)\leq B$. We can select $\alpha,C,\gamma$ which satisfies that:
\begin{equation}{\label{lammapp30}}
p=(\frac{1}{\alpha}-(\frac{\gamma }{\alpha}-\frac{(1+\gamma)\sqrt{2}}{\bar\alpha \sqrt{C}}-
\frac{\beta (1+\gamma)}{\alpha\bar\alpha})\sqrt{1+\frac{\sigma^2}{\epsilon}}\\
-(\frac{\beta(1+\gamma)^2}{2\bar\alpha^2}-\frac{\beta (1+\gamma)^2(2*\sqrt{2}+2)}{\bar\alpha^2 C}))(1+\frac{\sigma^2}{\epsilon})>0.
\end{equation}
In each round of FedDC, the global objective decreases as 
\begin{equation}{\label{lammap25}}
\begin{split}
\mathbb{E}_{C_t} L(w^t)&\leq L(w^{t-1})-2(\frac{1}{\alpha}-(\frac{\gamma }{\alpha}-\frac{(1+\gamma)\sqrt{2}}{\bar\alpha \sqrt{C}}-
\frac{\beta (1+\gamma)}{\alpha\bar\alpha})\sqrt{1+\frac{\sigma^2}{\epsilon}}\\
&-(\frac{\beta(1+\gamma)^2}{2\bar\alpha^2}-\frac{\beta (1+\gamma)^2(2*\sqrt{2}+2)}{\bar\alpha^2 C}))(1+\frac{\sigma^2}{\epsilon})*||\nabla L(w^{t-1})||^2\\
&\leq L(w^{t-1})-2p||\nabla L(w^{t-1})||^2.
\end{split}
\end{equation}

\subsection{Convergence of FedDC in convex case}
We suppose $\beta_d=0,\bar\alpha=\alpha$ in the convex case, if $\gamma=0$, $B\leq \sqrt{C}$, we can find that $||\nabla_L(w^t)||$ is proportional decreased. Assuming $1<<B \leq 0.5\sqrt{C}$, we get 
\begin{equation}{\label{lammapp31}}
\begin{split}
&\mathbb{E}_{C_t} L(w^t)\leq L(w^{t-1})-2(
\frac{1-\frac{\sqrt{2}B}{\sqrt{C}}}{\alpha}-\frac{B+(\frac{(2\sqrt{2}+2)}{C}+\frac{1}{2})\beta B^2}{\alpha^2})||\nabla L(w^{t-1})||^2,
\end{split}
\end{equation}
and
\begin{equation}{\label{lammapp32}}
\begin{split}
&\mathbb{E}_{C_t} L(w^t)\leq L(w^{t-1})-(\frac{1}{\alpha}-\frac{3\beta B^2}{\alpha^2})||\nabla L(w^{t-1})||^2.
\end{split}
\end{equation}
Setting $\alpha = 6\beta B^2$, we get  
\begin{equation}{\label{lammapp33}}
\begin{split}
&\mathbb{E}_{C_t} L(w^t)\leq L(w^{t-1})-\frac{1}{12\beta B^2}||\nabla L(w^{t-1})||^2.
\end{split}
\end{equation}
We can use the decrease in the global objective according to above inequality to characterize the FedDC's convergence rate. To achieve a threshold $\epsilon$ where $\sum_{t=1}^T||\nabla L(w^t)||^2\leq \epsilon$, if the model achieve the optimal point at $T$ round, we denoted $\Gamma=L(w^0)-L(w^*)$. From \ref{lammapp33} and the above definition, we get:
\begin{equation}{\label{lammapp34}}
\begin{split}
&\mathbb{E}_{S_T} L(w^T)- L(w^{0})=\mathbb{E}_{C_t} L(w^*)- L(w^{0})\leq -\sum_{t=1}^T\frac{1}{12\beta B^2}||\nabla L(w^{t-1})||^2\\
&\rightarrow 
\sum_{t=1}^T||\nabla L(w^{t-1})||^2 \leq 12\beta B^2 (L(w^0)-L(w^*)).
\end{split}
\end{equation}
Thus, FedDC spend $O(\frac{\beta B^2\Gamma}{\epsilon})$ to achieve convergence state where $\sum_{t=1}^T||\nabla L(w^{t-1})||^2\leq \epsilon$.


\end{document}